%% file: arxiv.tex
\documentclass[final]{arxiv/cvpr}
\usepackage{times}
\usepackage{epsfig}
\usepackage{graphicx}
\usepackage{amsmath}
\usepackage{amsthm}
\usepackage{amssymb}
\usepackage{multirow}
\usepackage{float}
\usepackage{makecell}
\usepackage{adjustbox}
\usepackage{stackengine}
\usepackage{array}
\usepackage{subcaption}
\usepackage{xr-hyper}
\usepackage{capt-of}

\usepackage{booktabs}
\usepackage[table, x11names]{xcolor}
\usepackage[normalem]{ulem}
\usepackage{hhline}
\usepackage[section]{placeins}
\usepackage{enumitem}
\usepackage{xr}
\usepackage{comment}

\usepackage[pagebackref=true,breaklinks=true,colorlinks,bookmarks=false]{hyperref}

\newtheorem{defn}{Definition}

\newtheorem{property}{Property}

\newcommand{\cutsectionup}{\vspace*{-0.15in}}

\newcommand{\cutsubsectiondown}{\vspace*{-0.07in}}
\newcommand{\cutparagraphup}{\vspace*{-0.1in}}

\begin{document}
\title{SetVAE: Learning Hierarchical Composition for\\Generative Modeling of Set-Structured Data}
\author{
\begin{tabular}[t]{cccc}
Jinwoo Kim\thanks{Equal contribution} & Jaehoon Yoo\footnotemark[1] & Juho Lee & Seunghoon Hong
\end{tabular}\\
KAIST\\
{\tt\small\{jinwoo-kim, wogns98, juholee, seunghoon.hong\}@kaist.ac.kr}
}
\maketitle
\begin{abstract}
   Generative modeling of set-structured data, such as point clouds, requires reasoning over local and global structures at various scales.
   However, adopting multi-scale frameworks for ordinary sequential data to a set-structured data is nontrivial as it should be invariant to the permutation of its elements.
   In this paper, we propose SetVAE, a hierarchical variational autoencoder for sets.
   Motivated by recent progress in set encoding, we build SetVAE upon attentive modules that first partition the set and project the partition back to the original cardinality.
   Exploiting this module, our hierarchical VAE learns latent variables at multiple scales, capturing coarse-to-fine dependency of the set elements while achieving permutation invariance.
   We evaluate our model on point cloud generation task and achieve competitive performance to the prior arts with substantially smaller model capacity.
   We qualitatively demonstrate that our model generalizes to unseen set sizes and learns interesting subset relations without supervision.
   Our implementation is available at \url{https://github.com/jw9730/setvae}.
\end{abstract}
\vspace{-0.2in}
\input{arxiv/introduction}
\input{arxiv/preliminaries}
\input{arxiv/method}
\input{arxiv/related_work}
\input{arxiv/experiment}
\input{arxiv/conclusion}
\cutparagraphup
\paragraph{Acknowledgement} 
This work was supported in part by Institute of Information \& communications Technology Planning \&  Evaluation (IITP) grant (2020-0-00153, 2016-0-00464, 2019-0-00075), Samsung Electronics, HPC support by MSIT \& NIPA.
{\small
\bibliographystyle{arxiv/ieee_fullname}
\bibliography{arxiv/egbib}
}
\clearpage
\clearpage
\appendix
\input{arxiv/supp}
\end{document}


\title{Learning Hierarchical Composition for\\ Generative Modeling of Exchangeable Data\\ {\emph {\large Supplementary Document}}}

\author{First Author\\
Institution1\\
Institution1 address\\
{\tt\small firstauthor@i1.org}
\and
Second Author\\
Institution2\\
First line of institution2 address\\
{\tt\small secondauthor@i2.org}
}


\maketitle
This document provides comprehensive descriptions and results of our method that could not be accommodated in the main paper due to space restriction.

\section{Implementation details}
In this section, we discuss detailed derivations and descriptions of SetVAE presented in Section~\ref{sec:method} and Section~\ref{sec:architecture}.

\subsection{KL Divergence of Initial Set Distribution}
\label{appendix:initial_kl}
This section provides a proof that the KL divergence between the approximate posterior and the prior over the initial set in Eq.~\eqref{eqn:elbo_vanillaSetVAE} and \eqref{eqn:elbo_hierarchicalSetVAE} is a constant. 

Following the definition in Eq.~\eqref{eqn:prior_initialset} and Eq.~\eqref{eqn:posterior_vanillaSetVAE}, we decompose the prior as $p(\mathbf{z}^{(0)}) = p(n)p(\mathbf{z}^{(0)}|n)$ and the approximate posterior as $q(\mathbf{z}^{(0)}|\mathbf{x}) = \delta(n)q(\mathbf{z}^{(0)}|n, \mathbf{x})$, where $\delta(n)$ is defined as a delta function centered at $n=|\mathbf{x}|$.
Here, the conditionals are given by
\begin{align}
     p(\mathbf{z}^{(0)}|n) &= \prod_{i=1}^n{p(\mathbf{z}_i^{(0)})}, \\
     q(\mathbf{z}^{(0)}|n, \mathbf{x}) &= \prod_{i=1}^n{q(\mathbf{z}_i^{(0)}|\mathbf{x})}.
\end{align}
As described in the main text, we set the elementwise distributions identical, $p(\mathbf{z}_i^{(0)}) = q(\mathbf{z}_i^{(0)}|\mathbf{x})$.
This renders the conditionals equal,
\begin{equation}
    p(\mathbf{z}^{(0)}|n) = q(\mathbf{z}^{(0)}|n, \mathbf{x}).\label{eqn:initial_set_equal}
\end{equation}
Then, the KL divergence between the approximate posterior and the prior in Eq.~\eqref{eqn:elbo_vanillaSetVAE} is written as
\begin{align}
    \textnormal{KL}&(q(\mathbf{z}^{(0)}|\mathbf{x})||p(\mathbf{z}^{(0)})) \nonumber\\
    &= \textnormal{KL}(\delta(n)q(\mathbf{z}^{(0)}|n, \mathbf{x})||p(n)p(\mathbf{z}^{(0)}|n)) \\
    &= \textnormal{KL}(\delta(n)p(\mathbf{z}^{(0)}|n)||p(n)p(\mathbf{z}^{(0)}|n)), \label{eqn:kl_initial_set}
\end{align}
where the second equality comes from the Eq.~\eqref{eqn:initial_set_equal}.
From the definition of KL divergence, we can rewrite Eq.~\eqref{eqn:kl_initial_set} as
\begin{align}
    \textnormal{KL}&(q(\mathbf{z}^{(0)}|\mathbf{x})||p(\mathbf{z}^{(0)})) \nonumber\\
    &= \mathbb{E}_{\delta(n)p(\mathbf{z}^{(0)}|n)}{\left[\log{\frac{\delta(n)p(\mathbf{z}^{(0)}|n)}{p(n)p(\mathbf{z}^{(0)}|n)}}\right]} \\
    &= \mathbb{E}_{\delta(n)p(\mathbf{z}^{(0)}|n)}{\left[\log{\frac{\delta(n)}{p(n)}}\right]}, \label{eqn:kl_initial_set_reduced}
\end{align}
As the logarithm in Eq.~\eqref{eqn:kl_initial_set_reduced} does not depend on $\mathbf{z}^{(0)}$, we can take it out from the expectation over $p(\mathbf{z}^{(0)}|n)$ as follows:
\begin{align}
    \textnormal{KL}&(q(\mathbf{z}^{(0)}|\mathbf{x})||p(\mathbf{z}^{(0)})) \nonumber\\
    &= \mathbb{E}_{\delta(n)}{\left[\mathbb{E}_{p(\mathbf{z}^{(0)}|n)} {\left[\log{\frac{\delta(n)}{p(n)}}\right]}\right]} \\
    &= \mathbb{E}_{\delta(n)}{\left[\log{\frac{\delta(n)}{p(n)}}\right]}, \label{eqn:kl_initial_set_cardinality}
\end{align}
which can be rewritten as
\begin{align}
    \mathbb{E}_{\delta(n)}{\left[\log{\delta(n)} - \log{p(n)}\right]}.
\end{align}
The expectation over the delta function $\delta(n)$ is simply an evaluation at $n = |\mathbf{x}|$.
As $\delta$ is defined over a discrete random variable $n$, its probability mass at the center $|\mathbf{x}|$ equals 1.
Therefore, $\log{\delta(n)}$ at $n = |\mathbf{x}|$ reduces to $\log1 = 0$, and we obtain
\begin{align}
    \textnormal{KL}&(q(\mathbf{z}^{(0)}|\mathbf{x})||p(\mathbf{z}^{(0)})) = - \log p(|\mathbf{x}|). \label{eqn:kl_constant}
\end{align}
As discussed in the main text, we model $p(n)$ using the data cardinality's empirical distribution.
Thus, $p(|\mathbf{x}|)$ only depends on cardinality distribution of data, and $- \textnormal{KL}(q(\mathbf{z}^{(0)}|\mathbf{x})||p(\mathbf{z}^{(0)}))$ in Eq.~\eqref{eqn:elbo_vanillaSetVAE} can be omitted from optimization.

\subsection{Implementation of SetVAE}
\label{appendix:abl}
\begin{figure}[!t]
    \centering
    \begin{subfigure}[b]{0.395\linewidth}
        \centering
        \includegraphics[width=0.95\linewidth]{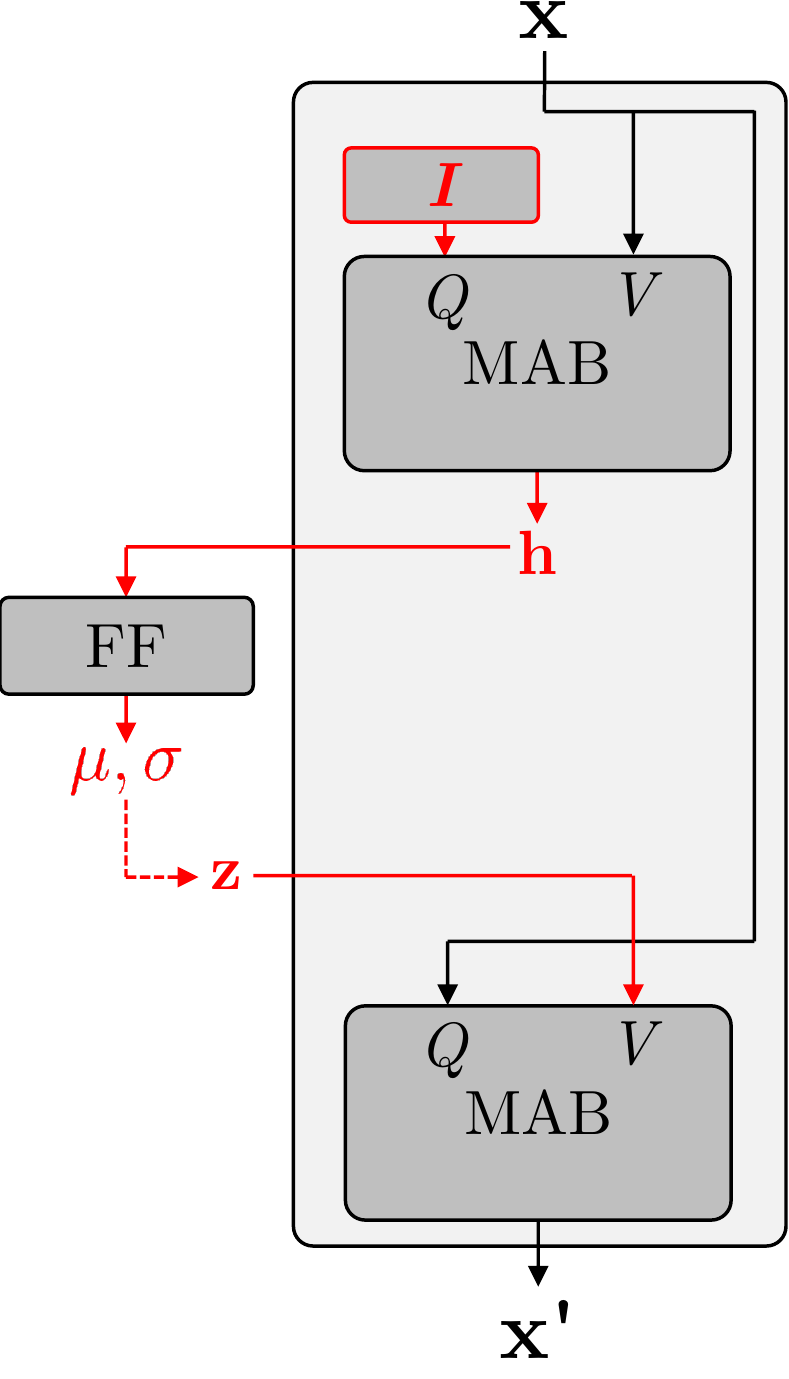}
        \caption{Generation}
        \label{fig:abl_prior}
    \end{subfigure}
    \begin{subfigure}[b]{0.59\linewidth}
        \centering
        \includegraphics[width=0.95\linewidth]{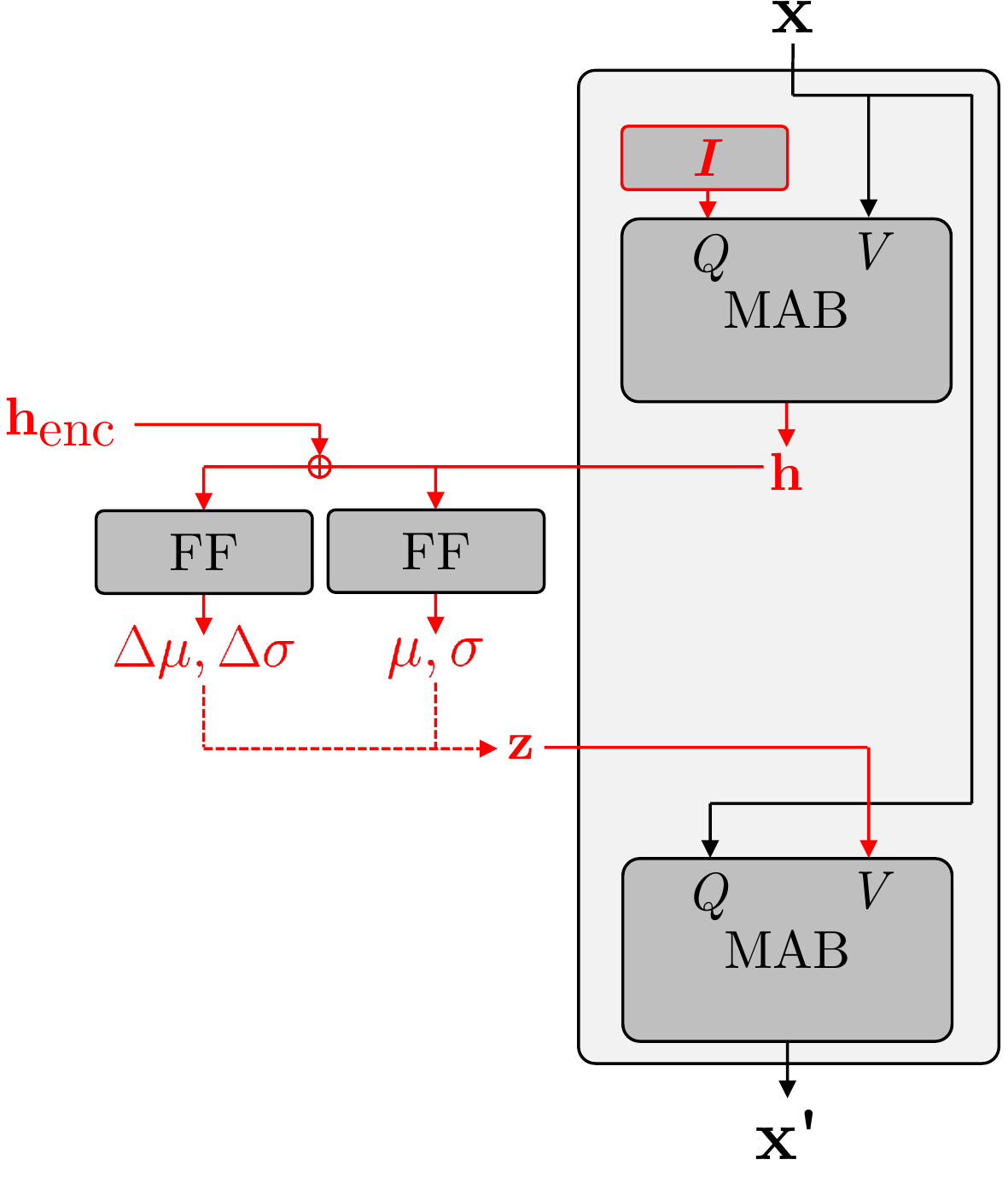}
        \caption{Inference}
        \label{fig:abl_posterior}
    \end{subfigure}
    \caption{The detailed structure of Attentive Bottleneck Layer during sampling (for generation) and inference (for reconstruction).}
\label{fig:abl_structure}
\end{figure}

\paragraph{Attentive Bottleneck Layer.}
In Figure~\ref{fig:abl_structure}, we provide the detailed structure of Attentive Bottleneck Layer (ABL) that composes the top-down generator of SetVAE (Section~\ref{sec:architecture}).
We share the common parameters in ABL for generation and inference, which is known to be effective in stabilizing the training of hierarchical VAE~\cite{kingma2017improving, vahdat2020nvae}. 

During generation (Figure~\ref{fig:abl_prior}), $\mathbf{z}$ is sampled from a Gaussian prior $\mathcal{N}(\mu, \sigma)$ (Eq.~\eqref{eqn:abp_z_prior}).
To predict $\mu$ and $\sigma$ from $\mathbf{h}$, we employ an elementwise fully-connected ($\textnormal{FC}$) layer, where its parameters are shared across each element of $\mathbf{h}$.
During inference, we sample the latent variables from the approximate posterior $\mathcal{N}(\mu+\Delta\mu, \sigma\cdot\Delta\sigma)$, where the correction factors $\Delta\mu, \Delta\sigma$ are predicted from the bottom-up encoding $\mathbf{h}_\textnormal{enc}$ by an additional $\textnormal{FC}$ layer.
Note that the $\textnormal{FC}$ layer for predicting $\mu, \sigma$ is shared for both generation and inference, and the $\textnormal{FC}$ layer that predicts $\Delta\mu, \Delta\sigma$ is the only component used exclusively during inference.

\paragraph{Slot Attention in ISAB and ABL.}
SetVAE discovers subset representation via projection attention in ISAB (Eq.~\eqref{eqn:isab_bottleneck}) and ABL (Eq.~\eqref{eqn:abl_bottleneck}).
However, with a plain attention mechanism, the projection attention may ignore some parts of input by simply not attending to them.
To prevent this, in both ISAB and ABL, we change the projection attention to Slot Attention \cite{locatello2020objectcentric}.

Specifically, plain projection attention\footnote{For simplicity, we explain with single-head attention instead of $\textnormal{MultiHead}$.} (Eq~\eqref{eqn:MAB_att}) treats input $\mathbf{x}\in\mathbb{R}^{n\times d}$ as key ($K$) and value ($V$), and uses a set of inducing points $\mathbf{I}\in\mathbb{R}^{m\times d}$ as query ($Q$).
First, it obtains the attention score matrix as follows:
\begin{equation}
    A = \frac{QK^\mathrm{T}}{\sqrt{d}} \in\mathbb{R}^{m\times n}.
\end{equation}
Here, each row index denotes an inducing point, and each column index denotes an input element.
Then, the value set is aggregated using $A$.
With $\textnormal{Softmax}(\cdot, d)$ denoting softmax normalization along $d$-th axis, the plain attention applies softmax to each row (key axis) of $A$, as follows:
\begin{align}
    \textnormal{Att}(Q, K, V) &= WV\in\mathbb{R}^{m\times d}, \\
    \textnormal{ where } W &= \textnormal{Softmax}(A, 2)\in\mathbb{R}^{m\times n}.
\end{align}
As a result, an input element can get zero attention if every query suppresses it.
To prevent this, Slot Attention applies softmax across each column (query axis) of $A$:
\begin{align}
    \textnormal{SlotAtt}(Q, K, V) &= WV, \\
    \textnormal{ where } W_{ij} = \frac{A'_{ij}}{\sum_{l=1}^n{A'_{il}}}&
    \textnormal{ for } A' = \textnormal{Softmax}(A, 1).
\end{align}
As attention coefficients for an input element sum up to 1 after softmax, slot attention guarantees that an input element is not ignored by every inducing points.

With adaptation of Slot Attention, we observe that inducing points often attend to distinct subsets of the input to produce $\mathbf{h}$, as illustrated in the Figure~\ref{fig:subset} and Figure~\ref{fig:composition} of the main text.
This is similar to the observation of \cite{locatello2020objectcentric} that the competition across queries encouraged segmented representations (slots) of objects from a multi-object image.
A difference is that unlike in \cite{locatello2020objectcentric} where the queries are \textit{i.i.d.} noise vectors, we design the query set as learnable parameter $\mathbf{I}$ without stochasticity.
Also, we do not introduce any refinement steps to the projected set $\mathbf{h}$ to avoid the complication of the model.

\section{Experiment Details}
This section discusses the detailed descriptions and additional results of experiments in Section~\ref{sec:experiment} in the main paper.

\subsection{ShapeNet Evaluation Metrics}
\label{appendix:metrics}
We provide descriptions of evaluation metrics used in the ShapeNet experiment (Section~\ref{sec:experiment} in the main paper).
We measure standard metrics including coverage (COV), minimum matching distance (MMD), and 1-nearest neighbor accuracy (1-NNA) \cite{achlioptas2018learning, yang2019pointflow}.
Following recent literature \cite{kim2020softflow}, we omit Jensen-Shannon Divergence (JSD) \cite{achlioptas2018learning} because it does not assess the fidelity of each point cloud.

Let $S_g$ be the set of generated point clouds and $S_r$ be the set of reference point clouds with $|S_r| = |S_g|$.

\emph{Coverage (COV)} measures the percentage of reference point clouds that is a nearest neighbor of at least one generated point cloud, computed as follows:
\begin{equation}
    \textnormal{COV}(P_g, P_r) = \frac{|\{\textnormal{argmin}_{\mathbf{y}\in S_r}D(\mathbf{x}, \mathbf{y})|\mathbf{x}\in S_g\}|}{|S_r|}.
\end{equation}

\emph{Minimum Matching Distance (MMD)} measures the average distance from each reference point cloud to its nearest neighbor in the generated point clouds:
\begin{equation}
    \textnormal{MMD}(P_g, P_r) = \frac{1}{|S_r|}\sum_{\mathbf{y}\in S_r}{\min_{\mathbf{x}\in S_g}{D(\mathbf{x}, \mathbf{y})}}.
\end{equation}

\emph{1-Nearest Neighbor Accuracy (1-NNA)} assesses whether two distributions are identical.
Let $S_{-\mathbf{x}}=S_r\cup S_g-\{\mathbf{x}\}$ and $N_{\mathbf{x}}$ be the nearest neighbor of $\mathbf{x}$ in $S_{-\mathbf{x}}$. With $\mathbf{1}(\cdot)$ an indicator function:
\begin{align}
    \textnormal{1-NNA}&(S_g, S_r) \nonumber\\
    &= \frac{\sum_{\mathbf{x}\in S_g}\mathbf{1}(N_{\mathbf{x}}\in S_g) + \sum_{\mathbf{y}\in S_r}\mathbf{1}(N_{\mathbf{y}}\in S_r)}{|S_g| + |S_r|}.
\end{align}

\subsection{Hierarchical Disentanglement}
This section describes an evaluation protocol used in Figure~\ref{fig:lda} in the main paper.
To investigate the latent representations learned at each level, we employed Linear Discriminant Analysis (LDA) as simple layer-wise classifiers.
The classifiers take the latent variable at each layer $\mathbf{z}^{l},~\forall l\in[1,L]$ as an input, and predict the identity and position of two digits (in $4\times4$ quantized grid) respectively in Set-MultiMNIST dataset.
To this end, we first train the SetVAE in the training set of Set-MultiMNIST.
Then we train the LDA classifiers using the validation dataset, where the input latent variables are sampled from the posterior distribution of SetVAE (Eq.~\eqref{eqn:posterior}). 
We report the training accuracy of the classifiers at each layer in Figure~\ref{fig:lda}.

\subsection{Ablation study}
In this section, we provide details of the ablation study presented in Table~\ref{table:ablation} of the main text.

\paragraph{Baseline}
As baselines, we use a SetVAE with unimodal Gaussian prior over the initial set elements, and a non-hierarchical, Vanilla SetVAE presented in Section~\ref{sec:method}.

To implement a SetVAE with unimodal prior, we only change the initial element distribution $p(\mathbf{z}_i^{(0)})$ from MoG (Eq.~\eqref{eqn:prior_mog}) to a multivariate Gaussian with a diagonal covariance matrix $\mathcal{N}(\mu^{(0)}, \sigma^{(0)})$ with learnable $\mu^{(0)}$ and $\sigma^{(0)}$.
This approach is adopted in several previous works in permutation-equivariant set generation \cite{yang2019pointflow, kosiorek2020conditional, locatello2020objectcentric}.

\begin{figure}[!t]
    \centering
    \begin{subfigure}[b]{0.395\linewidth}
        \centering
        \includegraphics[width=0.9\linewidth]{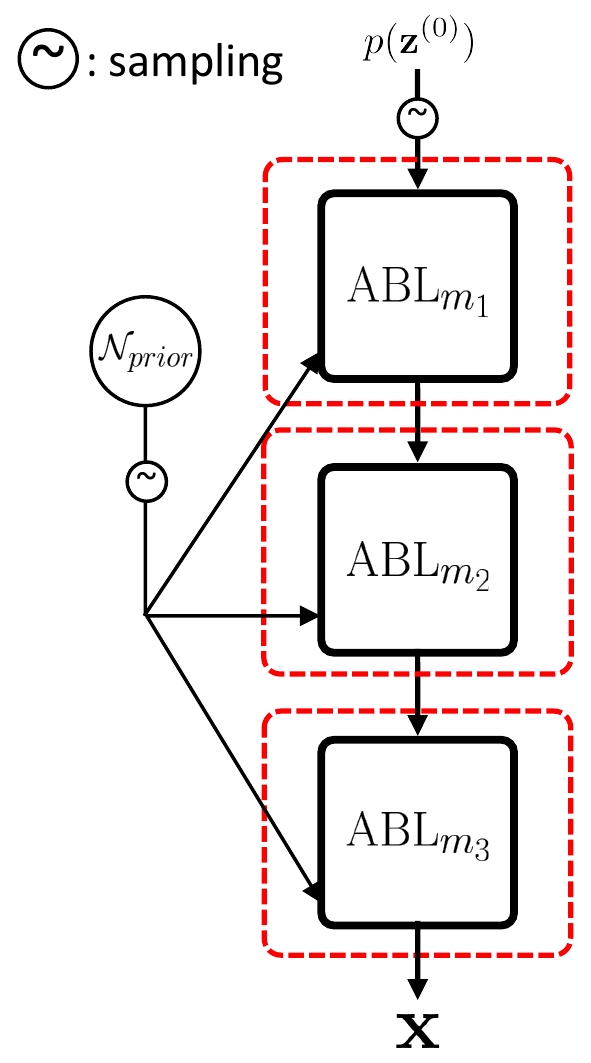}
        \caption{Generation}
        \label{fig:vanilla_prior}
    \end{subfigure}
    \begin{subfigure}[b]{0.59\linewidth}
        \centering
        \includegraphics[width=0.9\linewidth]{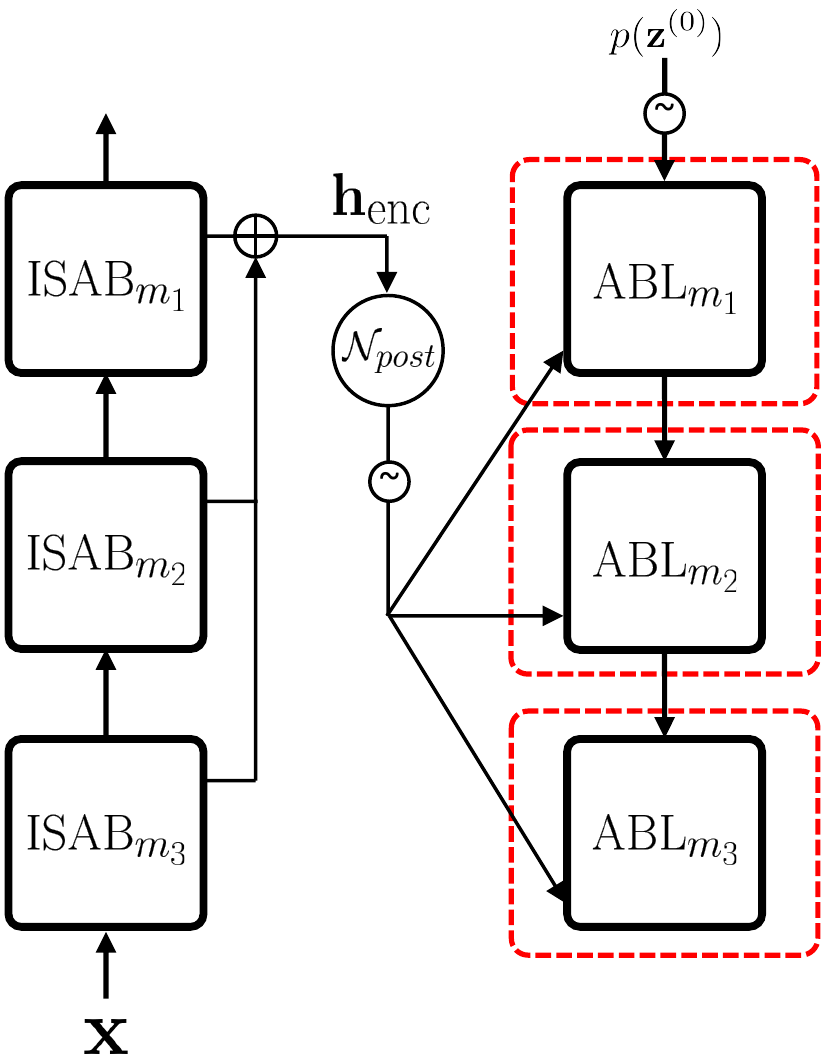}
        \caption{Inference}
        \label{fig:vanilla_posterior}
    \end{subfigure}
    \caption{Structure of Vanilla SetVAE without hierarchical priors and subset reasoning in generator.}
\label{fig:vanilla_setvae}
\end{figure}
To implement a Vanilla SetVAE, we employ a bottom-up encoder same to our full model and make the following changes to the top-down generator.
As illustrated in Figure~\ref{fig:vanilla_setvae}, we remove the subset relation in the generator by fixing the latent cardinality to 1 and employing a global prior $\mathcal{N}(\mu_1, \sigma_1)$ with $\mu_1, \sigma_1 \in \mathbb{R}^{1\times d}$ for all ABL.
As a permutation-invariant $\mathbf{h}_{\textnormal{enc}} \in \mathbb{R}^{1\times d}$,
we aggregate every elements of $\mathbf{h}$ from all levels of encoder network by average pooling.
During inference, $\mathbf{h}_{\textnormal{enc}}$ is provided to every ABL in the top-down generator.

\paragraph{Evaluation metric}
For the ablation study of SetVAE on the Set-MultiMNIST dataset, we measure the generation quality in image space by rendering each set instance to $64\times 64$ binary image based on the occurrence of a point in a pixel bin.
To measure the generation performance, we compute Frechet Inception Distance (FID) score \cite{heusel2017gans} using the output of the penultimate layer of a VGG11 network trained from scratch for MultiMNIST image classification into 100 labels (00-99).
Given the channel-wise mean $\mu_g$, $\mu_r$ and covariance matrix $\Sigma_g$, $\Sigma_r$ of generated and reference set of images respectively, we compute FID as follows:
\begin{equation}
    d^2 = {\| \mu_g - \mu_r \|}^2 + \textnormal{Tr} (\Sigma_g + \Sigma_r - 2 \sqrt{\Sigma_g \Sigma_r}).
\end{equation}
To train the VGG11 network, we replace the first conv layer to take single-channel inputs, and use the same MultiMNIST train set used for SetVAE.
We use SGD optimizer with Nesterov momentum, with learning rate 0.01, momentum 0.9, and L2 regularization weight 5e-3 to prevent overfitting.
We train the network for 10 epochs using batch size 128 so that the training top-1 accuracy exceeds 95\%.

\section{More Qualitative Results}
\paragraph{Ablation study}
This section provides additional results of the ablation study, which corresponds to the Table~\ref{table:ablation} of the main paper.
We compare the SetVAE with two baselines: SetVAE with a unimodal prior and the one using a single global latent variable (\emph{i.e.}, Vanilla SetVAE).

\begin{figure}[!t]
    \centering
    \includegraphics[width=0.8\linewidth]{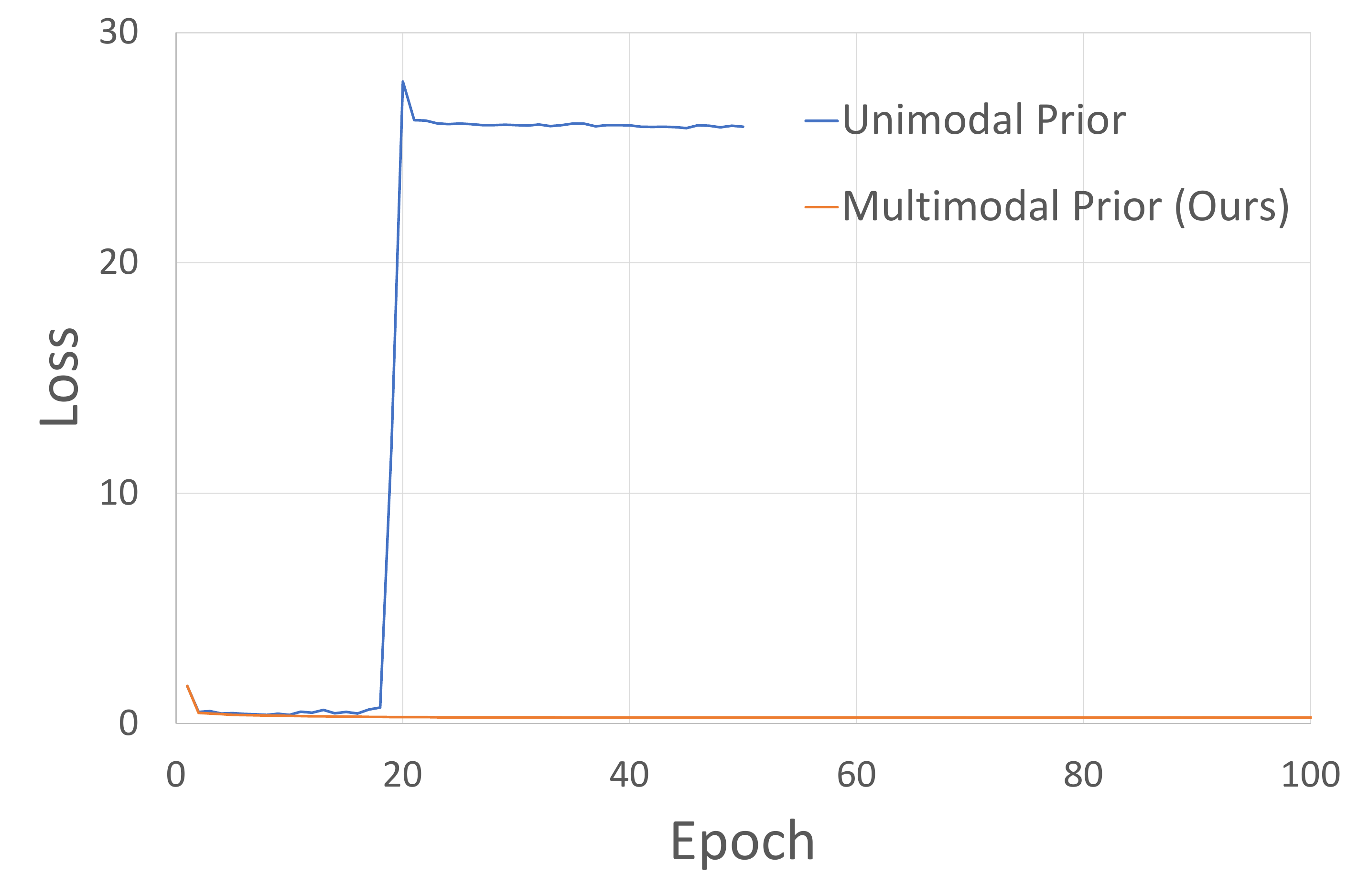}
    \caption{Training loss curves from SetVAE with multimodal and unimodal initial set trained on Set-MultiMNIST dataset.}
    \label{fig:ablation-curve}
\end{figure}

Figure~\ref{fig:ablation-curve} shows the training loss curves of SetVAE and the unimodal prior baseline on the Set-MultiMNIST dataset.
We observe that training of the unimodal baseline is unstable compared to SetVAE that uses a 4-component MoG.
We conjecture that a flexible initial set distribution provides a cue for the generator to learn stable subset representations.

\begin{figure}[!t]
    \centering
    \includegraphics[width=0.47\textwidth]{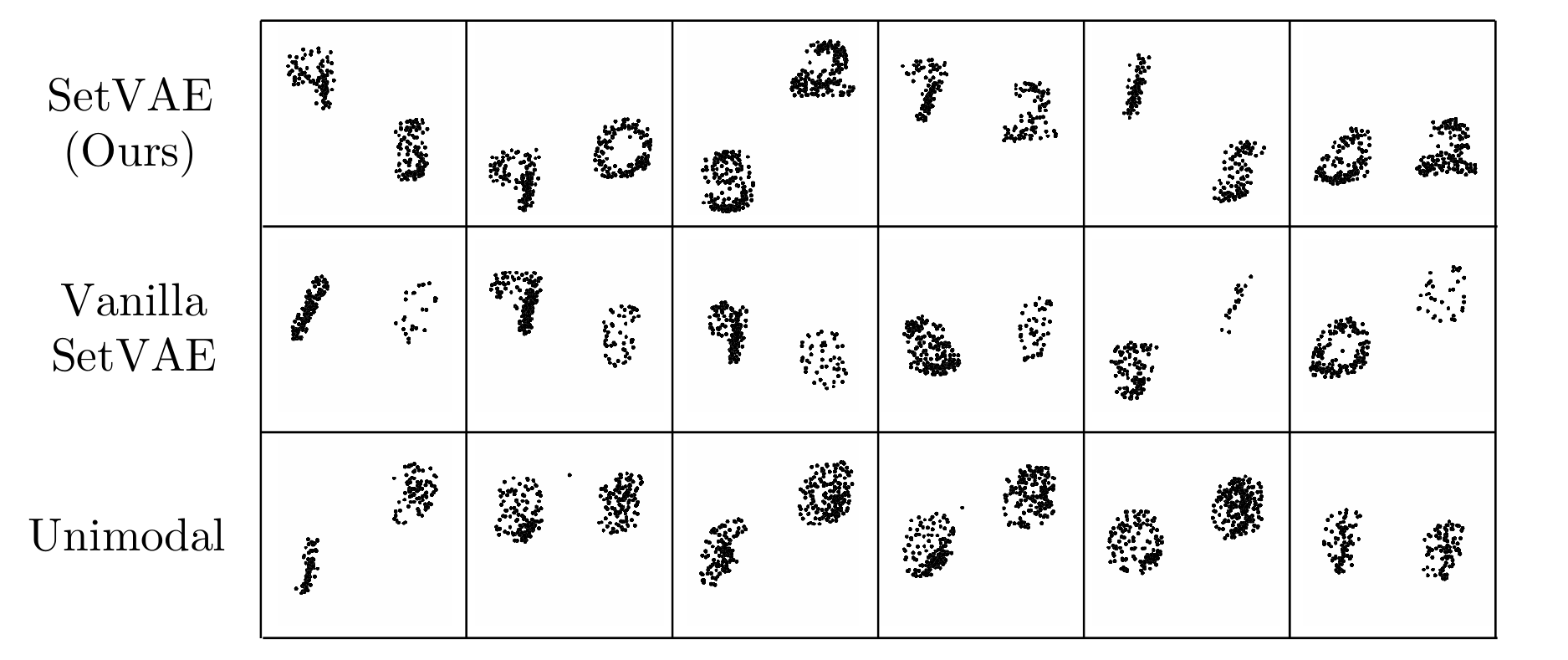}
    \caption{Samples from SetVAE and its ablated version trained on Set-MultiMNIST dataset.}
    \label{fig:ablation}
\end{figure}

In Figure~\ref{fig:ablation}, we present visualized samples from SetVAE and the two baselines.
As the training of unimodal SetVAE was unstable, we provide the results from a checkpoint before the training loss diverges (third row of Figure~\ref{fig:ablation}).
The Vanilla SetVAE without hierarchy (second row of Figure~\ref{fig:ablation}) focuses on modeling the left digit only and fails to assign a balanced number of points.
This failure implies that multi-level subset reasoning in the generative process is essential in faithfully modeling complex set data such as Set-MultiMNIST.

\begin{figure*}[!ht]
    \centering
    \includegraphics[width=0.9\textwidth]{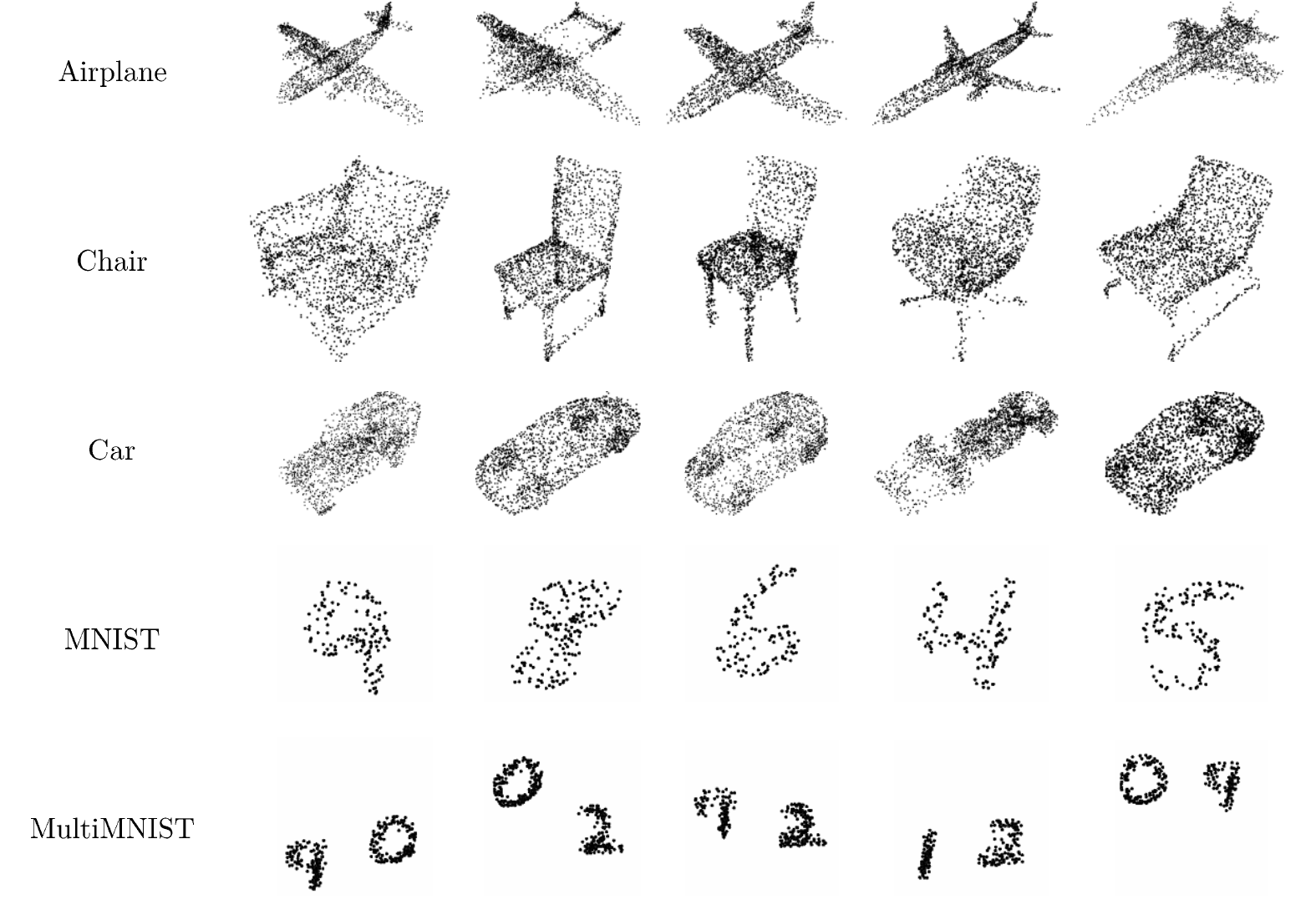}
    \caption{Additional examples of generated point clouds from SetVAE.}
    \label{fig:more_samples}
\end{figure*}

\begin{figure*}[!ht]
    \centering
    \includegraphics[width=1\textwidth]{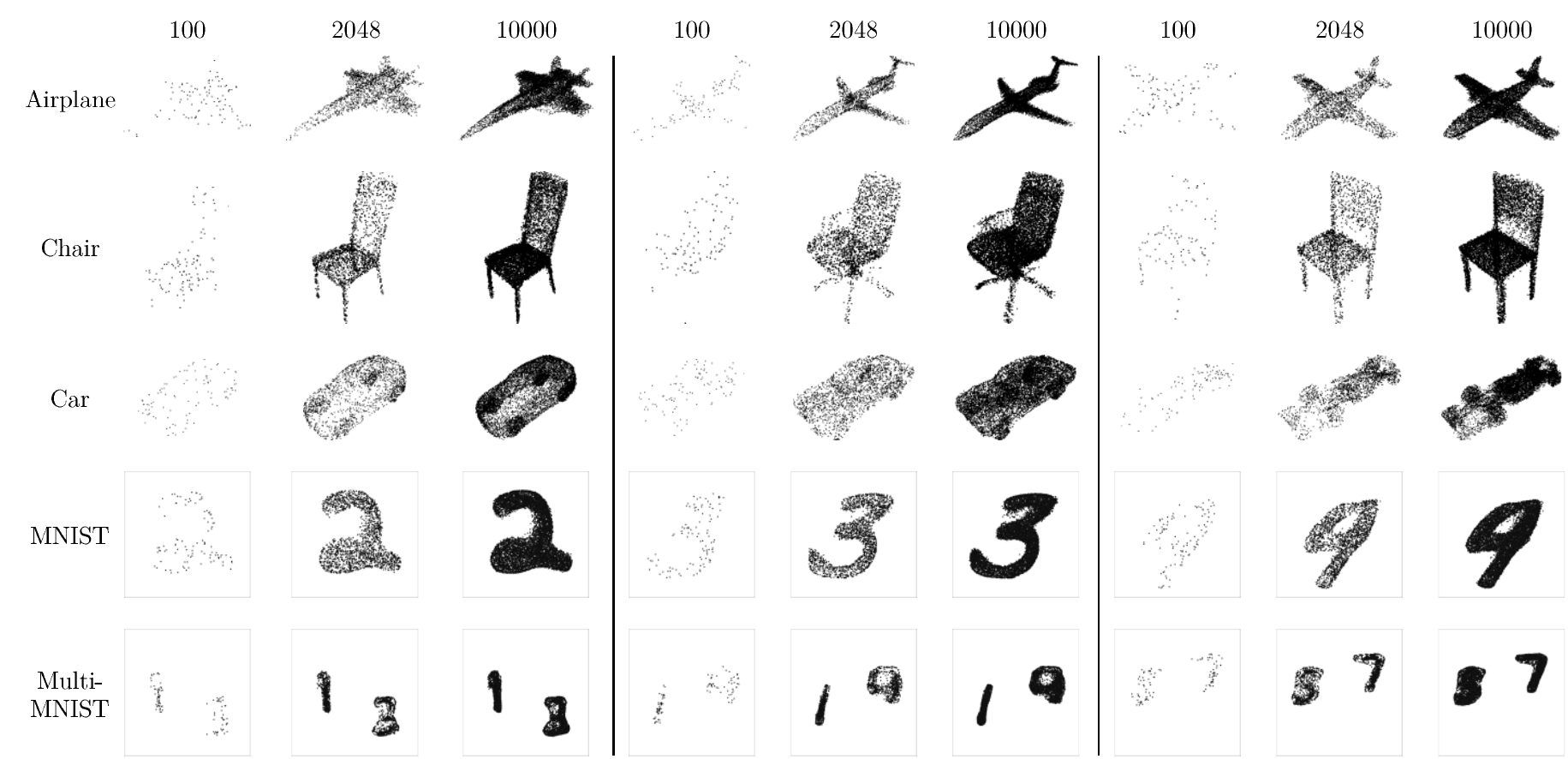}
    \caption{Additional examples demonstrating cardinality generalization of SetVAE.}
    \label{fig:more_disentanglement}
\end{figure*}

\paragraph{ShapeNet results}
Figure~\ref{fig:more_samples} presents the generated samples from SetVAE on ShapeNet, Set-MNIST and Set-MultiMNIST datasets, extending the results in Figure~\ref{fig:compare_pointflow} of the main text.
As illustrated in the figure, SetVAE generates point sets with high diversity while capturing thin and sharp details (\eg engines of an airplane and legs of a chair, \etc).
\paragraph{Cardinality disentanglement}
Figure~\ref{fig:more_disentanglement} presents the additional results of Figure~\ref{fig:cardinality} in the main paper, which illustrates samples generated by increasing the cardinality of the initial set $\mathbf{z}^{(0)}$ while fixing the hierarchical latent variables $\mathbf{z}^{(1:L)}$.
As illustrated in the figure, SetVAE is able to disentangle the cardinality of a set from the rest of its generative factors, and is able to generalize to unseen cardinality while preserving the disentanglement.

\begin{figure*}[!ht]
    \centering
    \includegraphics[width=1\textwidth]{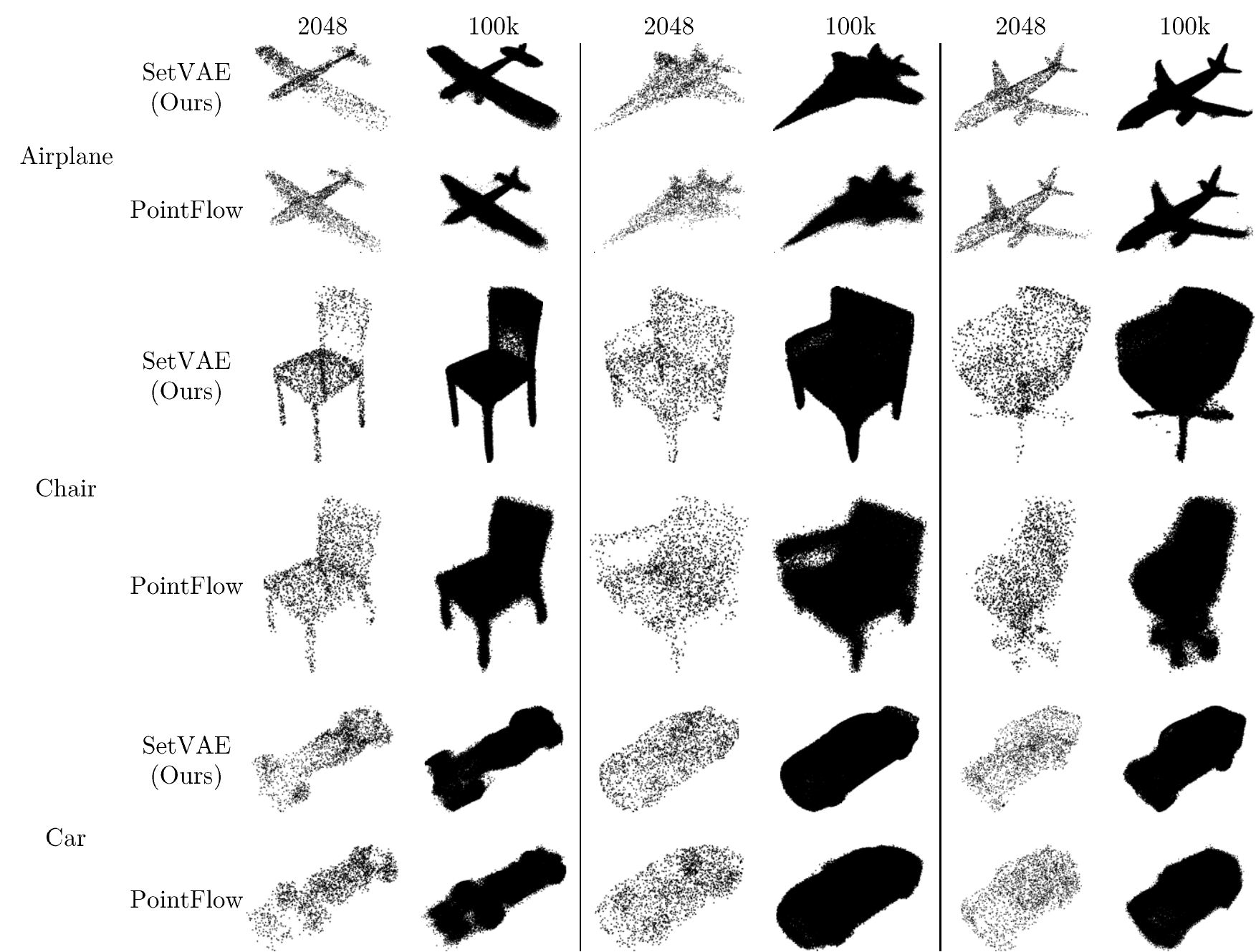}
    \caption{More examples in high-cardinality setting, compared with PointFlow.}
    \label{fig:more_mega_cardinality}
\end{figure*}

Notably, SetVAE can retain the disentanglement and generalize even to a high cardinality (100k) as well.
Figure~\ref{fig:more_mega_cardinality} presents the comparison to PointFlow with varying cardinality, which extends the results of the Figure~\ref{fig:mega-cardinality} in the main paper.
Unlike PointFlow that exhibits degradation and blurring of fine details, SetVAE retains the fine structure of the generated set even for extreme cardinality.

\begin{figure*}[!ht]
    \centering
    \includegraphics[height=0.9\textheight]{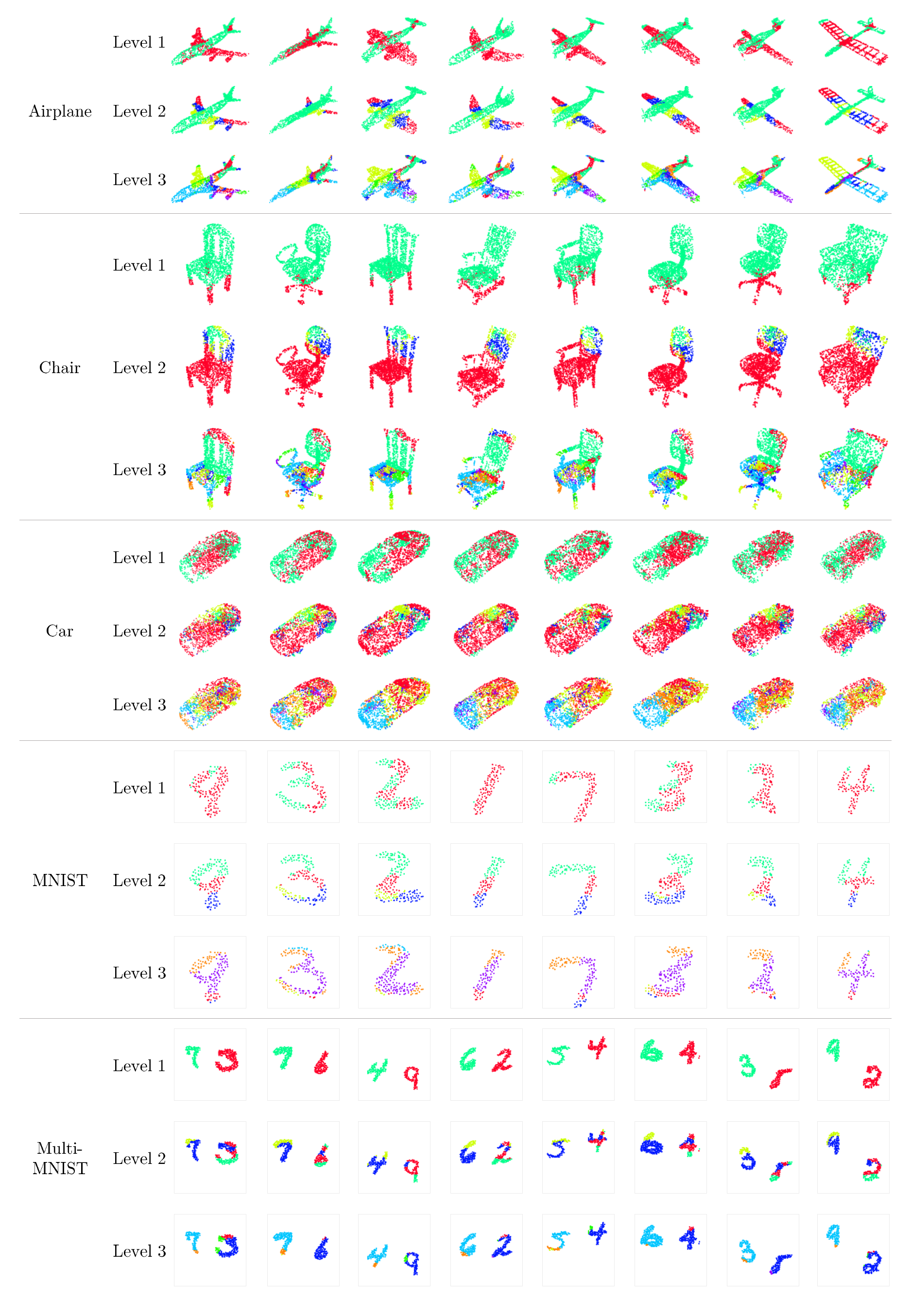}
    \caption{More examples of color-coded encoder attention.}
    \label{fig:more_encoder_attention}
\end{figure*}

\begin{figure*}[!ht]
    \centering
    \includegraphics[height=0.9\textheight]{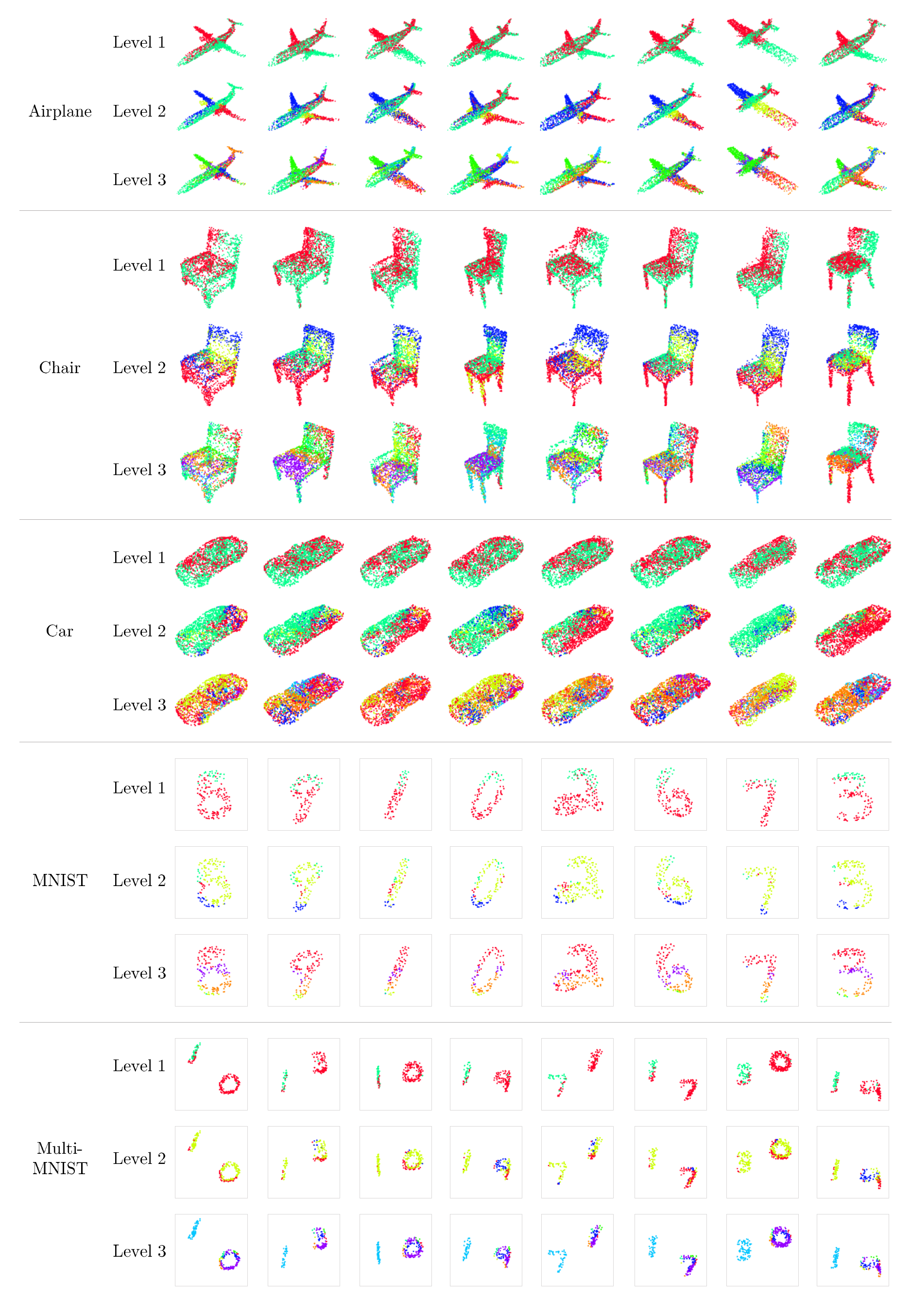}
    \caption{More examples of color-coded generator attention.}
    \label{fig:more_generator_attention}
\end{figure*}

\paragraph{Coarse-to-fine dependency}
In Figure~\ref{fig:more_encoder_attention} and Figure~\ref{fig:more_generator_attention}, we provide additional visualization of encoder and generator attention, extending the Figure~\ref{fig:subset} and Figure~\ref{fig:composition} in the main text.
We observe that SetVAE learns to attend to a subset of points consistently across examples.
Notably, these subsets often have a bilaterally symmetric structure or correspond to semantic parts.
For example, in the top level of the encoder (rows marked level 1 in Figure~\ref{fig:more_encoder_attention}), the subsets include wings of an airplane (colored red), legs of a chair (colored red), or hood \& wheels of a car (colored green).

Furthermore, SetVAE extends the subset modeling to multiple levels with a top-down increase in latent cardinality.
This allows SetVAE to encode or generate the structure of a set in various granularity, ranging from global structure to fine details.
Each column in Figure~\ref{fig:more_encoder_attention} and Figure~\ref{fig:more_generator_attention} illustrates the relations.
For example, in level 3 of Figure~\ref{fig:more_encoder_attention}, the bottom-up encoder partitions an airplane into fine-grained parts such as an engine, a tip of the wing, \etc.
Then, going bottom-up to level 1, the encoder composes them to symmetric pair of wings.
As for the top-down generator in Figure~\ref{fig:more_generator_attention}, it starts in level 1 by composing an airplane via the coarsely defined left and right sides.
Going top-down to level 3, the generator descends into fine-grained subsets like an engine and tail wing.

\clearpage


\section{Architecture and Hyperparameters}
\label{appendix:training}
Table~\ref{table:architecture} provides the network architecture and hyperparameters of SetVAE.
In the table, $\textnormal{FC}(d, f)$ denotes a fully-connected layer with output dimension $d$ and nonlinearity $f$.
$\textnormal{ISAB}_m(d, h)$ denotes an $\textnormal{ISAB}_m$ with $m$ inducing points, hidden dimension $d$, and $h$ heads (in Section~\ref{sec:settransformer}).
$\textnormal{MoG}_K(d)$ denotes a mixture of Gaussian (in Eq.~\eqref{eqn:prior_mog}) with $K$ components and dimension $d$.
$\textnormal{ABL}_m(d, d_z, h)$ denotes an $\textnormal{ABL}_m$ with $m$ inducing points, hidden dimension $d$, latent dimension $d_z$, and $h$ heads (in Section~\ref{sec:architecture}).
All $\textnormal{MAB}$s used in $\textnormal{ISAB}$ and $\textnormal{ABL}$ uses 2-layer MLP with $\textnormal{ReLU}$ activation as $\textnormal{FF}$ layers.

In Table~\ref{table:hyperparameters}, we provide detailed training hyperparameters.
For all experiments, we used Adam optimizer with first and second momentum parameters $0.9$ and $0.999$, respectively.

\begin{figure*}[!t]
\begin{minipage}{\linewidth}
\captionof{table}{Detailed network architectures used in our experiments.}
\vspace{-0.2cm}
\centering
    \begin{adjustbox}{width=0.95\textwidth}
        \label{table:architecture}
        \begin{tabular}{cc|cc|cc}
        \Xhline{2\arrayrulewidth}
        \\[-1em] \multicolumn{2}{c}{\textbf{ShapeNet}} & \multicolumn{2}{c}{\textbf{Set-MNIST}} & \multicolumn{2}{c}{\textbf{Set-MultiMNIST}}\\
        \\[-1em]\Xhline{2\arrayrulewidth}
        \\[-1em] \textbf{Encoder} & \textbf{Generator} & \textbf{Encoder} & \textbf{Generator} & \textbf{Encoder} & \textbf{Generator} \\
        \\[-1em]\Xhline{2\arrayrulewidth}
        \\[-1em]
        Input: $\textnormal{FC}(64, -)$ & Initial set: $\textnormal{MoG}_4(32)$ & Input: $\textnormal{FC}(64, -)$ & Initial set: $\textnormal{MoG}_4(32)$ & Input: $\textnormal{FC}(64, -)$ & Initial set: $\textnormal{MoG}_4(64)$ \\
        $\textnormal{ISAB}_{32}(64, 4)$ & $\textnormal{PlanarFlow}\times16$ & $\textnormal{ISAB}_{32}(64, 4)$ & $\textnormal{PlanarFlow}\times16$ & $\textnormal{ISAB}_{32}(64, 4)$ & $\textnormal{PlanarFlow}\times16$ \\
        $\textnormal{ISAB}_{16}(64, 4)$ & $\textnormal{ABL}_{2}(64, 16, 4)$ & $\textnormal{ISAB}_{16}(64, 4)$ & $\textnormal{ABL}_{2}(64, 16, 4)$ & $\textnormal{ISAB}_{16}(64, 4)$ & $\textnormal{ABL}_{2}(64, 16, 4)$ \\
        $\textnormal{ISAB}_{8}(64, 4)$ & $\textnormal{ABL}_{4}(64, 16, 4)$ & $\textnormal{ISAB}_{8}(64, 4)$ & $\textnormal{ABL}_{4}(64, 16, 4)$ & $\textnormal{ISAB}_{8}(64, 4)$ & $\textnormal{ABL}_{4}(64, 16, 4)$ \\
        $\textnormal{ISAB}_{4}(64, 4)$ & $\textnormal{ABL}_{8}(64, 16, 4)$ & $\textnormal{ISAB}_{4}(64, 4)$ & $\textnormal{ABL}_{8}(64, 16, 4)$ & $\textnormal{ISAB}_{4}(64, 4)$ & $\textnormal{ABL}_{8}(64, 16, 4)$ \\
        $\textnormal{ISAB}_{2}(64, 4)$ & $\textnormal{ABL}_{16}(64, 16, 4)$ & $\textnormal{ISAB}_{2}(64, 4)$ & $\textnormal{ABL}_{16}(64, 16, 4)$ & $\textnormal{ISAB}_{2}(64, 4)$ & $\textnormal{ABL}_{16}(64, 16, 4)$ \\
         & $\textnormal{ABL}_{32}(64, 16, 4)$ & & $\textnormal{ABL}_{32}(64, 16, 4)$ & & $\textnormal{ABL}_{32}(64, 16, 4)$ \\
        & Output: $\textnormal{FC}(3, -)$ & & Output: $\textnormal{FC}(2, \textnormal{tanh})$ & & Output: $\textnormal{FC}(2, \textnormal{tanh})$ \\
        & & & $(\textnormal{Output}+1)/2$ & & $(\textnormal{Output}+1)/2$ \\
        \\[-1em]\Xhline{2\arrayrulewidth}
        \end{tabular}
    \end{adjustbox}
    \vspace{1cm}
    \captionof{table}{Detailed training hyperparameters used in our experiments.}
    \vspace{-0.2cm}
    \begin{adjustbox}{width=0.95\textwidth}
    \label{table:hyperparameters}
    \footnotesize
    \begin{tabular}{cccc}
    \Xhline{2\arrayrulewidth}
    \\[-1em] & \textbf{ShapeNet} & \textbf{Set-MNIST} & \textbf{Set-MultiMNIST}\\
    \\[-1em] \Xhline{2\arrayrulewidth}
    \\[-1em]
    Training epochs & 4000 & 200 & 500 \\
    Learning rate & 1e-3, linearly decayed to zero from 2000epoch & 1e-3 & 1e-3 \\
    $\beta$ (Eq.~\eqref{eqn:beta_vae_loss}) & 1.0 & 0.01 & 0.01 \\
    \\[-1em]\Xhline{2\arrayrulewidth}
    \end{tabular}
    \end{adjustbox}
\end{minipage}
\end{figure*}

\clearpage
{\small
\bibliographystyle{cvpr_draft/ieee_fullname.bst}
\bibliography{cvpr_draft/egbib.bib}
}

%% file: arxiv/introduction.tex
\section{Introduction}
\label{sec:intro}
There have been increasing demands in machine learning for handling \emph{set-structured} data (\emph{i.e.}, a group of unordered instances).
Examples of set-structured data include object bounding boxes~\cite{li2019grains, carion2020endtoend}, point clouds~\cite{achlioptas2018learning, li2018point}, support sets in the meta-learning~\cite{finn2017modelagnostic}, \emph{etc.}
While initial research mainly focused on building neural network architectures to encode sets~\cite{zaheer2017deep,lee2019set}, generative models for sets have recently grown popular~\cite{zhang2020deep,kosiorek2020conditional, stelzner2020generative,yang2019pointflow, yang2020energybased}.

A generative model for set-structured data should verify the two essential requirements: (i) \textit{exchangeability}, meaning that a probability of a set instance is invariant to its elements' ordering, and (ii) handling \textit{variable cardinality}, meaning that a model should flexibly process sets with variable cardinalities.
These requirements pose a unique challenge in set generative modeling, as they prevent the adaptation of standard generative models for sequences or images~\cite{goodfellow2014generative, karras2019stylebased, oord2016pixel,oord2016conditional}. 
For instance, typical operations in these models, such as convolution or recurrent operations, exploit implicit ordering of elements (\emph{e.g.}, adjacency), thus breaking the exchangeability.
Several works circumvented this issue by imposing heuristic ordering~\cite{hong2018inferring, eslami2016attend, greff2019multi, ritchie2019fast}.
However, when applied to set-structured data, any ordering assumed by a model imposes an unnecessary inductive bias that might harm the generalization ability.

\begin{figure}[t]
    \begin{center}
        \includegraphics[width=0.47\textwidth]{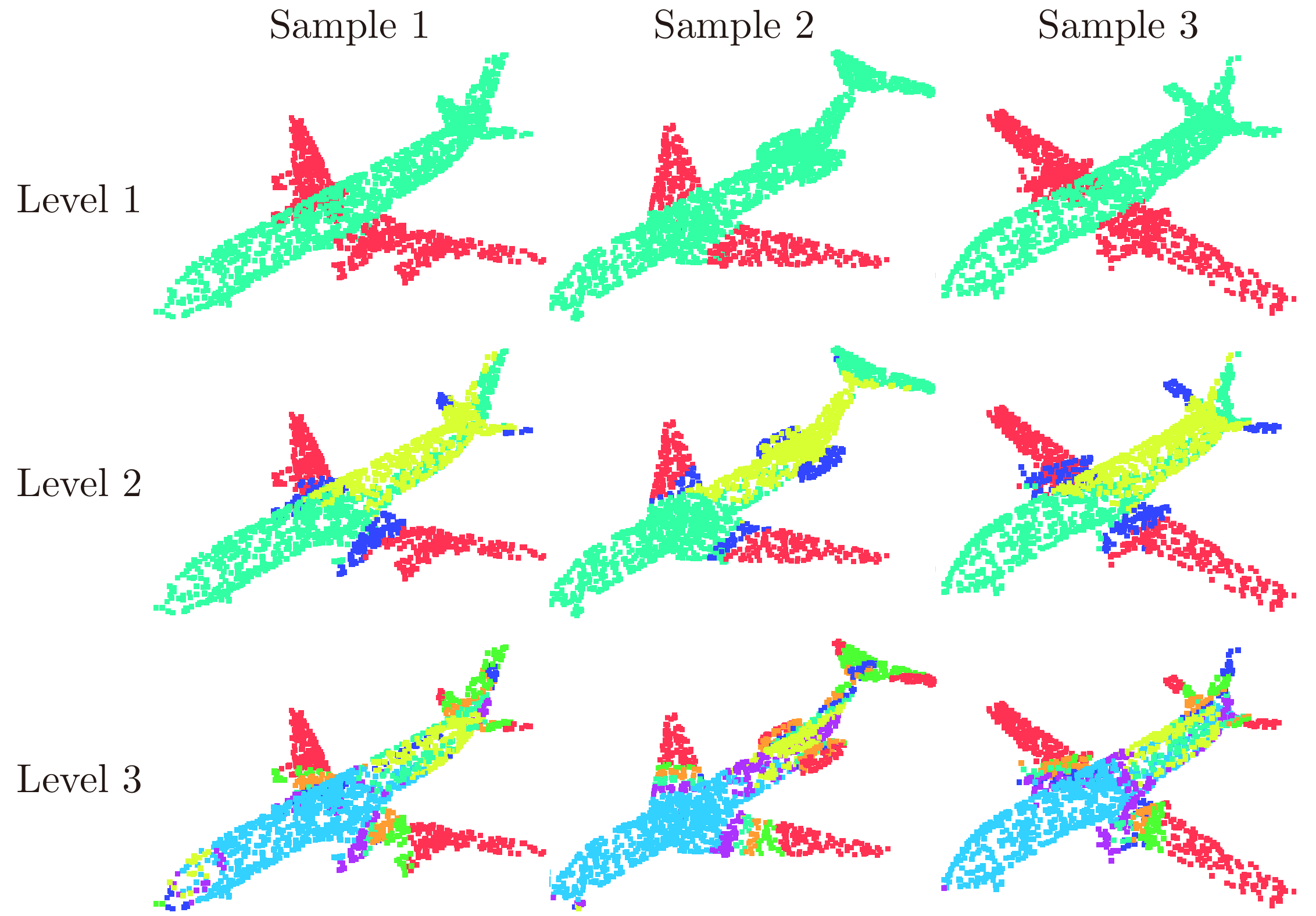}
    \end{center}
    \caption{Color-coded attention learned by SetVAE encoder for three data instances of ShapeNet Airplane \cite{chang2015shapenet}. Level 1 shows attention at the most coarse scale. Level 2 and 3 show attention at more fine-scales.
    }
\label{fig:eyecatcher}
\end{figure}

\begin{table*}[!ht]
\centering
\footnotesize
\caption{Summary of several set generative frameworks available to date. Our SetVAE jointly achieves desirable properties, with the advantages of the VAE framework combined with our novel contributions.
}
\begin{tabular}{l>{\centering}m{0.12\textwidth}>{\centering}m{0.1\textwidth}>{\centering}m{0.12\textwidth}>{\centering\arraybackslash}m{0.1\textwidth}}
\Xhline{2\arrayrulewidth}
\\[-1em] Model & Exchangeability & \shortstack{Variable\\cardinality} & \shortstack{Inter-element\\dependency} & \shortstack{Hierachical\\latent structure} \\
\\[-1em]\Xhline{2\arrayrulewidth}
\\[-1em]
l-GAN \cite{achlioptas2018learning} & $\times$ & $\times$ & $\bigcirc$ & $\times$ \\
PC-GAN \cite{li2018point} & $\bigcirc$ & $\bigcirc$ & $\times$ & $\times$ \\
PointFlow \cite{yang2019pointflow} & $\bigcirc$ & $\bigcirc$ & $\times$ & $\times$ \\
EBP \cite{yang2020energybased} & $\bigcirc$ & $\bigcirc$ & $\bigcirc$ & $\times$ \\
\\[-1em]\hline
\\[-1em] SetVAE (ours) & $\bigcirc$ & $\bigcirc$ & $\bigcirc$ & $\bigcirc$ \\
\\[-1em]\Xhline{2\arrayrulewidth}
\end{tabular}
\label{table:framework}
\end{table*}

There are several existing works satisfying these requirements.
Edwards et al.,~\cite{edwards2017neural} proposed a simple generative model encoding sets into latent variables, while other approaches build upon various generative models, such as generative adversarial networks~\cite{li2018point, stelzner2020generative}, flow-based models~\cite{yang2019pointflow,kim2020softflow}, and energy-based models~\cite{yang2020energybased}.
All these works define valid generative models for set-structured data, but with some limitations.
To achieve exchangeability, many approaches process set elements independently~\cite{li2018point, yang2019pointflow}, limiting the models in reflecting the interactions between the elements during generation.
Some approaches take the inter-element dependency into account~\cite{stelzner2020generative, yang2020energybased}, but have an upper bound on the number of elements~\cite{stelzner2020generative}, or less scalable due to heavy computations~\cite{yang2020energybased}.
More importantly, existing models are less effective in capturing subset structures in sets presumably because they represent a set with a single-level representation.
For sets containing multiple sub-objects or parts, it would be beneficial to allow a model to have structured latent representations such as one obtained via hierarchical latent variables.

In this paper, we propose SetVAE, a novel hierarchical variational autoencoder (VAE) for sets. SetVAE models interaction between set elements by adopting attention-based Set Transformers \cite{lee2019set} into the VAE framework, and extends it to a \textit{hierarchy} of latent variables \cite{sonderby2016ladder, vahdat2020nvae} to account for flexible subset structures.
By organizing latent variables at each level as a \textit{latent set} of fixed cardinality, SetVAE is able to learn hierarchical multi-scale features that decompose a set data in a coarse-to-fine manner (Figure \ref{fig:eyecatcher}) while achieving \textit{exchangeability} and handling \textit{variable cardinality}.
In addition, composing latent variables invariant to input's cardinality allows our model to generalize to arbitrary cardinality unseen during training.

The contributions of this paper are as follows:
\begin{itemize}
    \item We propose SetVAE, a novel hierarchical VAE for sets with \textit{exchangeability} and \textit{varying cardinality}.  To the best of our knowledge, SetVAE is the first VAE successfully applied for sets with arbitrary cardinality. SetVAE has a number of desirable properties compared to previous works, as summarized in Table \ref{table:framework}.
    \item Equipped with novel Attentive Bottleneck Layers (ABLs), SetVAE is able to model the coarse-to-fine dependency across the arbitrary number of set elements using a hierarchy of latent variables.
    \item We conduct quantitative and qualitative evaluations of SetVAE on generative modeling of point cloud in various datasets, and demonstrate better or competitive performance in generation quality with less number of parameters than the previous works.
\end{itemize}

%% file: arxiv/preliminaries.tex
\section{Preliminaries}
\label{sec:preliminary}
\subsection{Permutation-Equivariant Set Generation}

\label{sec:equivariant}
Denote a set as $\mathbf{x} = \{\mathbf{x}_i\}_{i=1}^n \in \mathcal{X}^n$, where $n$ is the cardinality of the set and $\mathcal{X}$ represents the domain of each element $\mathbf{x}_i \in\mathbb{R}^d$.
In this paper, we represent $\mathbf{x}$ as a matrix $\mathbf{x} = [\mathbf{x}_1, ..., \mathbf{x}_n]^\mathrm{T} \in\mathbb{R}^{n\times d}$.
Note that any operation on a set should be invariant to the elementwise permutation and satisfy the two constraints of \textit{permutation invariance} and \textit{permutation equivariance}.
\begin{defn}
A function $f:\mathcal{X}^n \rightarrow \mathcal{Y}$ is permutation invariant iff for any permutation $\pi(\cdot)$, $f(\pi(\mathbf{x}))= f(\mathbf{x})$.
\end{defn}
\begin{defn}
A function $f:\mathcal{X}^n \rightarrow \mathcal{Y}^n$ is permutation equivariant iff for any permutation $\pi(\cdot)$, $f(\pi(\mathbf{x}))=\pi( f(\mathbf{x}))$.
\end{defn}
In the context of generative modeling, the notion of permutation invariance translates into exchangeability, requiring a joint distribution of the elements invariant with respect to the permutation.
\begin{defn}
A distribution for a set of random variables $\mathbf{x} = \{\mathbf{x}_i\}_{i=1}^n$ is exchangeable if for any permutation $\pi$,
$p(\mathbf{x}) = p(\pi(\mathbf{x}))$.
\end{defn}

An easy way to achieve exchangeability is to assume each element to be \emph{i.i.d.} and process a set of initial elements $\mathbf{z}^{(0)} = \{\mathbf{z}_i^{(0)}\}_{i=1}^n$ independently sampled from $p(\mathbf{z}_i^{(0)})$ with an elementwise function $f_{\textnormal{elem}}$ to get the $\mathbf{x}$:
\begin{equation}
    \mathbf{x} = \{\mathbf{x}_i\}_{i=1}^n \textnormal{ where } \mathbf{x}_i = f_{\textnormal{elem}} (\mathbf{z}_i^{(0)})
\end{equation}

However, assuming elementwise independence poses a limit in modeling interactions between set elements.
An alternative direction is to process $\mathbf{z}^{(0)}$ with a permutation-equivariant function $f_{\textnormal{equiv}}$ to get the $\mathbf{x}$:
\begin{equation}
    \mathbf{x} = \{\mathbf{x}_i\}_{i=1}^n = f_{\textnormal{equiv}}(\{\mathbf{z}_i^{(0)}\}_{i=1}^n).
\end{equation}

We refer to this approach as the \textit{permutation-equivariant generative framework}.
As the likelihood of $\mathbf{x}$ does not depend on the order of its elements (because elements of $\mathbf{z}^{(0)}$ are i.i.d.), this approach achieves exchangeability.

\subsection{Permutation-Equivariant Set Encoding}
\label{sec:settransformer}
To design permutation-equivariant operations over a set, Set Transformer~\cite{lee2019set} provides attentive modules that model pairwise interaction between set elements while preserving invariance or equivariance.
This section introduces two essential modules in the Set Transformer. 
\begin{figure}[t!]
    \centering
    \begin{subfigure}[b]{0.23\textwidth}
        \centering
        \includegraphics[width=0.55\textwidth]{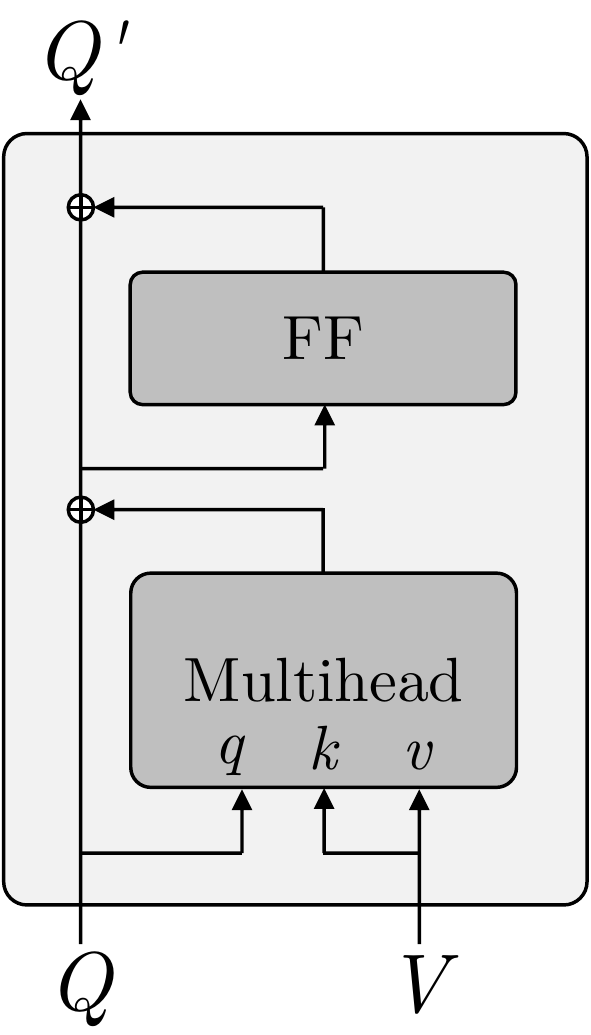}
        \caption{MAB}
        \label{fig:mab}
    \end{subfigure}
    \hfill
    \begin{subfigure}[b]{0.23\textwidth}
        \centering
        \includegraphics[width=0.55\textwidth]{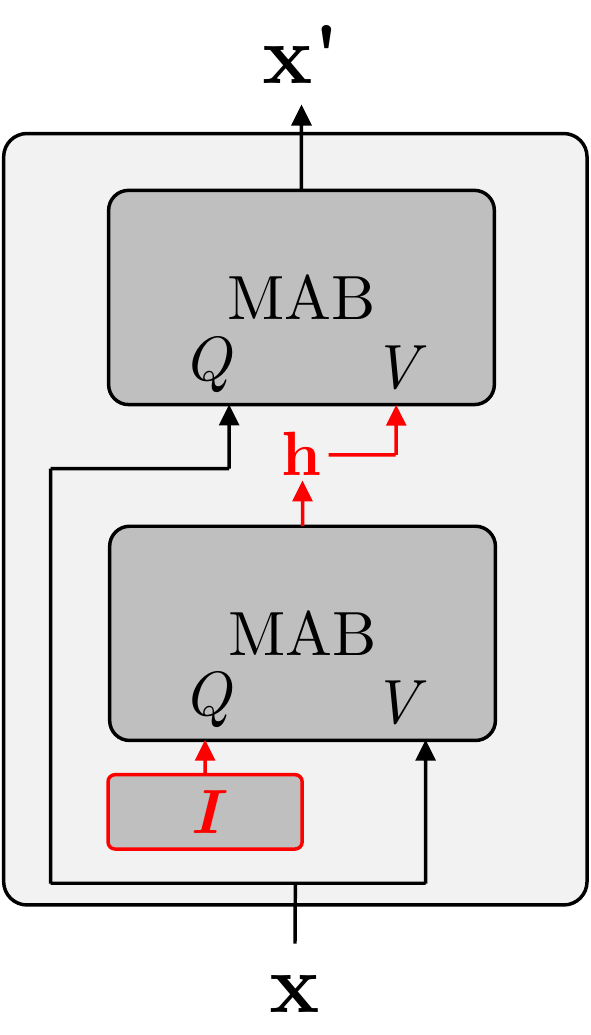}
        \caption{ISAB}
        \label{fig:isab}
    \end{subfigure}
    \vspace{-0.1in}
    \caption{Illustration of Multihead Attention Block (MAB) and Induced Set Attention Block (ISAB).}
\label{fig:st}
\vspace{-0.55cm}
\end{figure}

First, Multihead Attention Block (MAB) takes the query and value sets, $Q\in \mathbb{R}^{n_q\times d}$ and  $V\in \mathbb{R}^{n_v\times d}$, respectively, and performs the following transformation (Figure \ref{fig:mab}):
\begin{align}
    \textnormal{MAB}(Q, V) = \textnormal{LN}(\mathbf{a} + \textnormal{FF}(\mathbf{a})) \in \mathbb{R}^{n_q\times d},~~~~ \label{eqn:MAB}\\
    \textnormal{where }
    \mathbf{a} = \textnormal{LN}(Q + \textnormal{Multihead}(Q, V, V)) \in \mathbb{R}^{n_q\times d},\label{eqn:MAB_att}
\end{align}
where FF denotes elementwise feedforward layer, Multihead denotes multi-head attention~\cite{vaswani2017attention}, and LN denotes layer normalization \cite{lee2019set}.
Note that the output of Eq.~\eqref{eqn:MAB} is permutation equivariant to $Q$ and permutation invariant to $V$.

Based on MAB, Induced Set Attention Block (ISAB) processes the input set $\mathbf{x}\in\mathbb{R}^{n\times d}$ using a 
smaller set of inducing points $I\in \mathbb{R}^{m\times d}~(m<n)$ by (Figure \ref{fig:isab}):
\begin{align}
    \textnormal{ISAB}_m(\mathbf{x}) &= \textnormal{MAB}(\mathbf{x}, \mathbf{h}) \in \mathbb{R}^{n\times d}, \\
    \textnormal{where }\mathbf{h} &= \textnormal{MAB}(I, \mathbf{x}) \in \mathbb{R}^{m\times d}\label{eqn:isab_bottleneck}.
\end{align}
The ISAB first transforms the input set $\mathbf{x}$ into $\mathbf{h}$ by attending from $I$.
The resulting $\mathbf{h}$ is a permutation invariant projection of $\mathbf{x}$ to a lower cardinality $m$.
Then, $\mathbf{x}$ again attends to $\mathbf{h}$ to produce the output of $n$ elements.
As a result, ISAB is permutation equivariant to $\mathbf{x}$.

\begin{property}
In $\textnormal{ISAB}_m(\mathbf{x})$, $\mathbf{h}$ is permutation invariant to $\mathbf{x}$.
\end{property}
\begin{property}
\vspace{-0.2cm}
$\textnormal{ISAB}_m(\mathbf{x})$ is permutation equivariant to $\mathbf{x}$.
\end{property}

%% file: arxiv/method.tex
\section{Variational Autoencoders for Sets}
\label{sec:method}
The previous section suggests that there are two essential requirements for VAE for set-structured data: it should be able to model the likelihood of sets (i) in arbitrary cardinality and (ii) invariant to the permutation (\emph{i.e.,} exchangeable). 
This section introduces our SetVAE objective that satisfies the first requirement while achieving the second requirement is discussed in Section~\ref{sec:architecture}.

The objective of VAE \cite{kingma2014autoencoding} is to learn a generative model $p_\theta(\mathbf{x}, \mathbf{z}) = p_\theta(\mathbf{z}) p_\theta(\mathbf{x}|\mathbf{z})$ for data $\mathbf{x}$ and latent variables $\mathbf{z}$. 
Since the true posterior is unknown, we approximate it using the inference model $q_\phi(\mathbf{z}|\mathbf{x})$ and optimize the variational lower bound (ELBO) of the marginal likelihood $p(\mathbf{x})$:
\begin{equation}
    \mathcal{L}_{\text{VAE}} = \mathbb{E}_{q_\phi(\mathbf{z}|\mathbf{x})}{\left[\log p_\theta(\mathbf{x}|\mathbf{z})\right]}  - \text{KL}\left(q_\phi(\mathbf{z}|\mathbf{x})||p_\theta(\mathbf{z})\right).
    \label{eqn:original_vae}
\end{equation}

\cutparagraphup
\paragraph{Vanilla SetVAE}
When our data is a set $\mathbf{x}=\{\mathbf{x}_i\}_{i=1}^n$, Eq.~\eqref{eqn:original_vae} should be modified such that it can incorporate the set of arbitrary cardinality $n$\footnote{Without loss of generality, we use $n$ to denote the cardinality of a set but assume that the training data is composed of sets in various size.}.
To this end, we propose to decompose the latent variable $\mathbf{z}$ into the two independent variables as $\mathbf{z}=\{\mathbf{z}^{(0)}, \mathbf{z}^{(1)}\}$.
We define $\mathbf{z}^{(0)}=\{\mathbf{z}_i^{(0)}\}_{i=1}^n$ to be a set of \textit{initial elements}, whose cardinality is always the same as a data $\mathbf{x}$. 
Then we model the generative process as transforming $\mathbf{z}^{(0)}$ into a set $\mathbf{x}$ conditioned on the $\mathbf{z}^{(1)}$.

Given the independence assumption, the prior is factorized by $p(\mathbf{z})=p(\mathbf{z}^{(0)})p(\mathbf{z}^{(1)})$.
The prior on initial set $p(\mathbf{z}^{(0)})$ is further decomposed into the cardinality and element-wise distributions as:
\begin{equation}
    p(\mathbf{z}^{(0)}) = p(n)\prod_{i=1}^np(\mathbf{z}_{i}^{(0)}).
    \label{eqn:prior_initialset}
\end{equation}
We model $p(n)$ using the empirical distribution of the training data cardinality.
We find that the choice of the prior $p(\mathbf{z}_i^{(0)})$ is critical to the performance, and discuss its implementation in Section~\ref{sec:implementation}.

Similar to the prior, the approximate posterior is defined as $q(\mathbf{z}|\mathbf{x}) = q(\mathbf{z}^{(0)}|\mathbf{x})q(\mathbf{z}^{(1)}|\mathbf{x})$ and decomposed into:
\begin{equation}
    q(\mathbf{z}^{(0)}|\mathbf{x}) = q(n|\mathbf{x})\prod_{i=1}^nq(\mathbf{z}_{i}^{(0)}|\mathbf{x})
    \label{eqn:posterior_vanillaSetVAE}
\end{equation}
We define $q(n|\mathbf{x})=\delta(n)$ as a delta function with $n=|\mathbf{x}|$, and set $q(\mathbf{z}_{i}^{(0)}|\mathbf{x}) = p(\mathbf{z}_{i}^{(0)})$ similar to \cite{yang2019pointflow, kosiorek2020conditional, locatello2020objectcentric}.
The resulting ELBO can be written as
\begin{align}
    \mathcal{L}_{\text{SVAE}}&= \mathbb{E}_{q(\mathbf{z}|\mathbf{x})}{\left[ \log p(\mathbf{x}|\mathbf{z}) \right]} \nonumber\\
    &- \textnormal{KL}(q(\mathbf{z}^{(0)}|\mathbf{x})||p(\mathbf{z}^{(0)})) \nonumber\\
    &- \textnormal{KL}(q(\mathbf{z}^{(1)}|\mathbf{x})||p(\mathbf{z}^{(1)})).
    \label{eqn:elbo_vanillaSetVAE}
\end{align}
In the supplementary file, we show that the first KL divergence in Eq.~\eqref{eqn:elbo_vanillaSetVAE} is a constant and can be ignored in the optimization.
During inference, we sample $\mathbf{z}^{(0)}$ by first sampling the cardinality $n\sim p(n)$ then the $n$ initial elements independently from the prior $p(\mathbf{z}_{i}^{(0)})$.

\cutparagraphup
\paragraph{Hierarchical SetVAE}
To allow our model to learn a more expressive latent structure of the data, we can extend the vanilla SetVAE using hierarchical latent variables.

Specifically, we extend the plain latent variable $\mathbf{z}^{(1)}$ into $L$ disjoint groups $\{\mathbf{z}^{(1)}, ..., \mathbf{z}^{(L)}\}$, and introduce a top-down hierarchical dependency between $\mathbf{z}^{(l)}$ and $\{\mathbf{z}^{(0)}, ..., \mathbf{z}^{(l-1)}\}$ for every $l>1$. 
This leads to the modification in the prior and approximate posterior to
\begin{align}
    p(\mathbf{z}) &= p(\mathbf{z}^{(0)}) p(\mathbf{z}^{(1)}) \prod_{l>1} {p(\mathbf{z}^{(l)}|\mathbf{z}^{(<l)})}\label{eqn:prior}\\
    q(\mathbf{z}|\mathbf{x}) &= q(\mathbf{z}^{(0)}|\mathbf{x}) q(\mathbf{z}^{(1)}|\mathbf{x}) \prod_{l>1} {q(\mathbf{z}^{(l)}|\mathbf{z}^{(<l)}, \mathbf{x})}. \label{eqn:posterior}
\end{align}

Applying Eq.~\eqref{eqn:prior} and \eqref{eqn:posterior} to Eq.~\eqref{eqn:elbo_vanillaSetVAE}, we can derive the ELBO as
\begin{align}
    &\mathcal{L}_{\text{HSVAE}} = \mathbb{E}_{q(\mathbf{z}|\mathbf{x})}{\left[ \log p(\mathbf{x}|\mathbf{z}) \right]} \nonumber\\
    &~~- \textnormal{KL}(q(\mathbf{z}^{(0)}|\mathbf{x})||p(\mathbf{z}^{(0)})) - \textnormal{KL}(q(\mathbf{z}^{(1)}|\mathbf{x})||p(\mathbf{z}^{(1)})) \nonumber\\
    &~- \sum_{l=2}^L{\mathbb{E}_{q(\mathbf{z}^{(<l)}|\mathbf{x})}{\textnormal{KL}(q(\mathbf{z}^{(l)}|\mathbf{z}^{(<l)}, \mathbf{x})||p(\mathbf{z}^{(l)}|\mathbf{z}^{(<l)})}}.
    \label{eqn:elbo_hierarchicalSetVAE}
\end{align}

\cutparagraphup
\paragraph{Hierarchical prior and posterior}
To model the prior and approximate posterior in Eq.~\eqref{eqn:prior} and \eqref{eqn:posterior} with top-down latent dependency, we employ the bidirectional inference in \cite{sonderby2016ladder}.
We outline the formulations here and elaborate on the computations in Section~\ref{sec:architecture}. 

Each conditional $p(\mathbf{z}^{(l)}|\mathbf{z}^{(<l)})$ in the prior is modeled by the factorized Gaussian, whose parameters are dependent on the latent variables of the upper hierarchy $\mathbf{z}^{(<l)}$:
\begin{equation}
    p(\mathbf{z}^{(l)}|\mathbf{z}^{(<l)}) = \mathcal{N}\left(\mu_l(\mathbf{z}^{(<l)}), \sigma_l(\mathbf{z}^{(<l)})\right).
    \label{eqn:normal_prior}
\end{equation}

Similarly, each conditional in the approximate posterior $q(\mathbf{z}^{(l)}|\mathbf{z}^{(<l)}, \mathbf{x})$ is also modeled by the factorized Gaussian.
We use the residual parameterization in \cite{vahdat2020nvae} which predicts the parameters of the Gaussian using the displacement and scaling factors ($\Delta\mu$,$\Delta\sigma$) conditioned on $\mathbf{z}^{(<l)}$ and $\mathbf{x}$:
\begin{align}
    q(\mathbf{z}^{(l)}|\mathbf{z}^{(<l)}, \mathbf{x}) = \mathcal{N}(&\mu_l(\mathbf{z}^{(<l)}) + \Delta\mu_l(\mathbf{z}^{(<l)}, \mathbf{x}), \nonumber\\
    &\sigma_l(\mathbf{z}^{(<l)}) \cdot \Delta\sigma_l(\mathbf{z}^{(<l)}, \mathbf{x})).
    \label{eqn:normal_posterior}
\end{align}

\cutparagraphup
\paragraph{Invariance and equivariance}
We assume that the decoding distribution $p(\mathbf{x}|\mathbf{z}^{(0)}, \mathbf{z}^{(1:L)})$  is equivariant to the permutation of $\mathbf{z}^{(0)}$ and invariant to the permutation of $\mathbf{z}^{(1:L)}$ since such model induces an exchangeable model:
\begin{align}
    \lefteqn{p(\pi(\mathbf{x})) = \int p(\pi(\mathbf{x})|\pi(\mathbf{z}^{(0)}), \mathbf{z}^{(1:L)}) p(\pi(\mathbf{z}^{(0)})) p(\mathbf{z}^{(1:L)}) d\mathbf{z}}\nonumber\\
    &= \int p(\mathbf{x}|\mathbf{z}^{(0)},\mathbf{z}^{(1:L)}) p(\mathbf{z}^{(0)}) p(\mathbf{z}^{(1:L)}) d\mathbf{z} = p(\mathbf{x}).
\end{align}
We further assume that the approximate posterior distributions $q(\mathbf{z}^{(l)}|\mathbf{z}^{(<l)}, \mathbf{x})$ are invariant to the permutation of $\mathbf{x}$.
In the following section, we describe how we implement the encoder and decoder satisfying these criteria.

\section{SetVAE Framework}
\label{sec:architecture}
We present the overall framework of the proposed SetVAE. 
Figure~\ref{fig:overview} illustrates an overview.
SetVAE is based on the bidirectional inference~\cite{sonderby2016ladder}, which is composed of the bottom-up encoder and top-down generator sharing the same dependency structure.
In this framework, the inference network forms the approximate posterior by merging bottom-up information from data with the top-down information from the generative prior.
We construct the encoder using a stack of ISABs in Section~\ref{sec:settransformer}, and treat each of the projected set $\mathbf{h}$ as a deterministic encoding of data.

Our generator is composed of a stack of special layers called Attentive Bottleneck Layer (ABL), which extends the ISAB in Section~\ref{sec:settransformer} with the stochastic interaction with the latent variable.
Specifically, ABL processes a set at each layer of the generator as follows:
\begin{align}
    \text{ABL}_m(\mathbf{x}) &= \text{MAB}(\mathbf{x}, \text{FF}(\mathbf{z}))\in \mathbb{R}^{n\times d}\label{eqn:abp}\\
    \text{with}~~\mathbf{h} &= \text{MAB}(I, \mathbf{x})\in \mathbb{R}^{m\times d}\label{eqn:abl_bottleneck},
\end{align}
where $\text{FF}$ denotes a feed-forward layer, and the latent variable $\mathbf{z}$ is derived from the projection $\mathbf{h}$.
For generation (Figure~\ref{fig:setvae_generation}), we sample $\mathbf{z}$ from the prior in Eq.~\eqref{eqn:normal_prior} by,
\begin{align}
    \mathbf{z} &\sim \mathcal{N}(\mu, \sigma)~\text{where}~\mu, \sigma = \text{FF}(\mathbf{h}).
    \label{eqn:abp_z_prior}
\end{align}
For inference (Figure~\ref{fig:setvae_inference}), we sample $\mathbf{z}$ from the posterior in Eq.~\eqref{eqn:normal_posterior} by,
\begin{align}
&\mathbf{z} \sim \mathcal{N}(\mu+\Delta\mu, \sigma\cdot\Delta\sigma) \nonumber\\
&\text{where}~\Delta\mu, \Delta\sigma = \text{FF}(\mathbf{h}+\mathbf{h}_{\text{enc}}),
    \label{eqn:abp_z_posterior}
\end{align}
where $\mathbf{h}_{\text{enc}}$ is obtained from the corresponding ISAB layer of the bottom-up encoder.
Following \cite{sonderby2016ladder}, we share the parameters between the generative and inference networks.
A detailed illustration of ABL is in the supplementary file.

To generate a set, we first sample the initial elements $\mathbf{z}^{(0)}$ and the latent variable $\mathbf{z}^{(1)}$ from the prior $p(\mathbf{z}^{(0)})$ and $p(\mathbf{z}^{(1)})$, respectively. 
Given these inputs, the generator iteratively samples the subsequent latent variables $\mathbf{z}^{(l)}$ from the prior $p(\mathbf{z}^{(l)}|\mathbf{z}^{(<l)})$ one by one at each layer of ABL, while processing the set conditioned on the sampled latent variable via Eq.~\eqref{eqn:abp}. 
The data $\mathbf{x}$ is then decoded elementwise from the final output.

\subsection{Analysis}
\paragraph{Modeling Exchangeable Likelihood}
The architecture of SetVAE satisfies the invariance and equivariance criteria in Section~\ref{sec:method}.
This is, in part, achieved by producing latent variables from the projected sets of bottom-up ISAB and top-down ABL.
As the projected sets are permutation invariant to input (Section~\ref{sec:preliminary}), the latent variables $\mathbf{z}^{(1:L)}$ provide an invariant representation of the data.
Furthermore, due to permutation equivariance of ISAB, the top-down stack of ABLs produce an output equivariant to the initial set $\mathbf{z}^{(0)}$. This renders the decoding distribution $p(\mathbf{x}|\mathbf{z}^{(0)}, \mathbf{z}^{(1:L)})$ permutation equivariant to $\mathbf{z}^{(0)}$.
Consequently, the decoder of SetVAE induces an exchangeable model.

\begin{figure}[t!]
    \centering
    \begin{subfigure}[b]{0.17\textwidth}
        \centering
        \includegraphics[height=5.3cm]{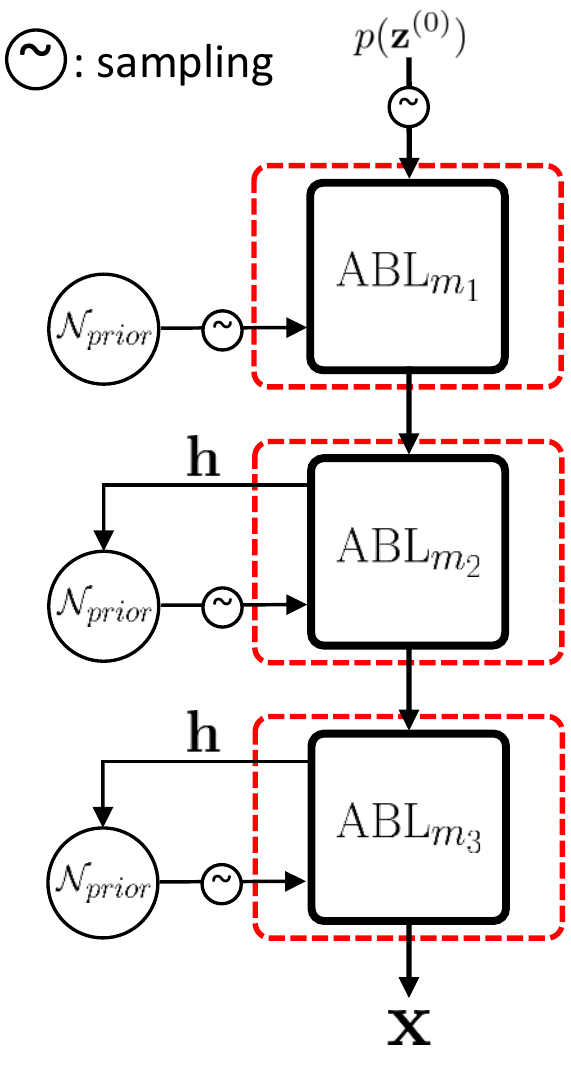}
        \vspace{-0.1cm}
        \caption{Generation}
        \label{fig:setvae_generation}
    \end{subfigure}
    \hfill
    \begin{subfigure}[b]{0.3\textwidth}
        \centering
        \includegraphics[height=5.3cm]{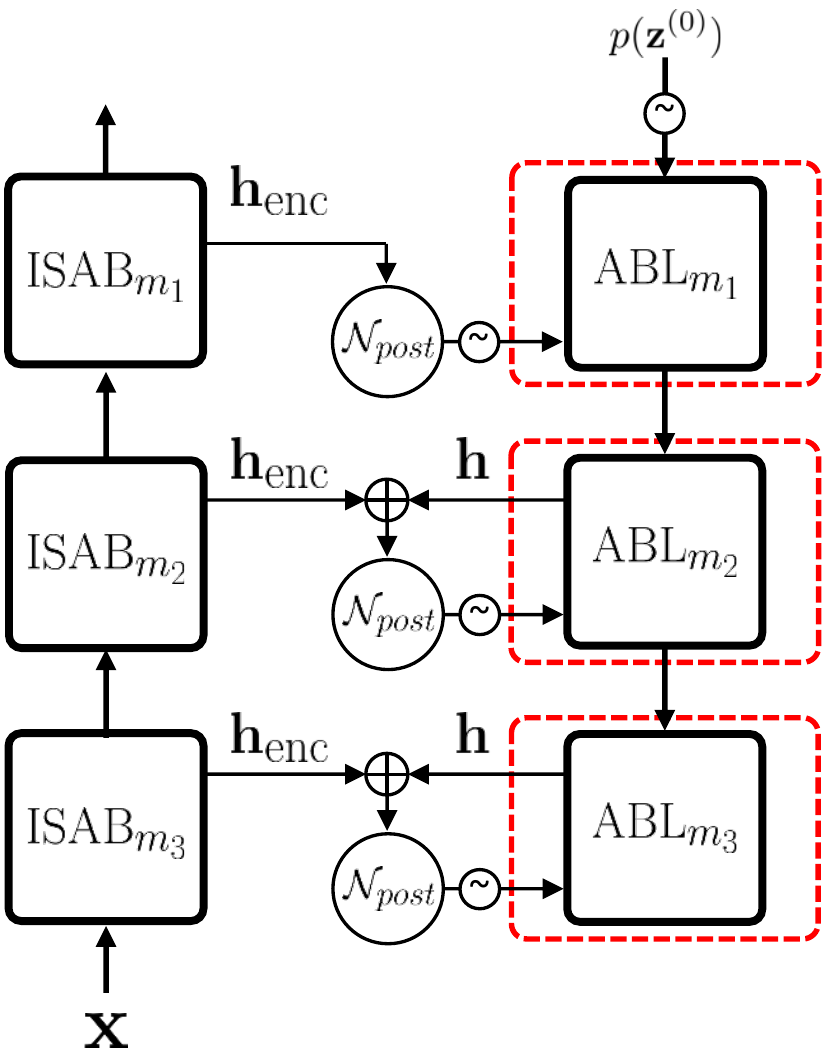}
        \vspace{-0.1cm}
        \caption{Inference}
        \label{fig:setvae_inference}
    \end{subfigure}
    \vspace{-0.6cm}
    \caption{The hierarchical SetVAE.
    $\mathcal{N}_{prior}$ denotes the prior (Eq.~\eqref{eqn:abp_z_prior}) and $\mathcal{N}_{post}$ denotes the posterior (Eq.~\eqref{eqn:abp_z_posterior}).
    }
    \vspace{-0.5cm}
\label{fig:overview}
\end{figure}

\cutparagraphup
\paragraph{Learning Coarse-to-Fine Dependency}
In SetVAE, both the ISAB and ABL project the input set $\mathbf{x}$ of cardinality $n$ to the projected set $\mathbf{h}$ of cardinality $m$ via multi-head attention (Eq.~\eqref{eqn:isab_bottleneck} and \eqref{eqn:abl_bottleneck}).
In the case of $m<n$, this projection functions as a bottleneck to the cardinality.
This allows the model to encode some features of $\mathbf{x}$ into the $\mathbf{h}$ and discover interesting \emph{subset} dependencies across the set elements.
Denoting $m_l$ as the bottleneck cardinality at layer $l$ (Figure~\ref{fig:overview}), we set $m_{l}<m_{l+1}$ to induce the model to discover coarse-to-fine dependency of the set, such as object parts.
Such bottleneck also effectively reduces network size, allowing our model to perform competitive or better than the prior arts with less than 50\% of their parameters.
This coarse-to-fine structure is a unique feature of SetVAE.

\subsection{Implementation Details}
\label{sec:implementation}
\cutsubsectiondown
This section discusses the implementation of SetVAE.
We leave comprehensive details on the supplementary file.

\cutparagraphup
\paragraph{Multi-Modal Prior.} 
Although a unimodal Gaussian is a typical choice for the initial element distribution $p(\mathbf{z}^{(0)}_i)$ \cite{kosiorek2020conditional, yang2019pointflow}, we find that the model converges significantly faster when we employ the multi-modal prior.
We use a mixture of Gaussians (MoG) with $K$ components:
\begin{equation}
    p(\mathbf{z}^{(0)}_i) = \sum_{k=1}^K{\pi_k\mathcal{N}(\mathbf{z}^{(0)}_i;\mu^{(0)}_k, \sigma^{(0)}_k)}.
    \label{eqn:prior_mog}
\end{equation}

\cutparagraphup
\paragraph{Likelihood.}
For the likelihood $p_\theta(\mathbf{x}|\mathbf{z})$, we may consider a Gaussian distribution centered at the reconstruction.
In the case of point sets, we design the likelihood by
\begin{align}
    \mathcal{L}_\text{recon}(\mathbf{x}) &= -\log p_\theta(\mathbf{x}|\mathbf{z})  
    \nonumber\\
    &= \frac{1}{2}d(\mathbf{x}, \hat{\mathbf{x}}) + \mathrm{const},
\end{align}
where $d(\mathbf{x}, \hat{\mathbf{x}})$ is the optimal matching distance defined as
\begin{align}
    d(\mathbf{x}, \hat{\mathbf{x}}) = \min_{\pi} {\sum_i}{\| \mathbf{x}_i - \hat{\mathbf{x}}_{\pi(i)}\|_2^2}.
\end{align}
In other words, we measure the likelihood with the Gaussian at optimally permuted $\mathbf{x}$, and thus maximizing this likelihood is equivalent to minimizing the optimal matching distance between the data and the reconstruction. Unfortunately, directly maximizing this likelihood requires $O(n^3)$ computation due to the matching. Instead, we choose the Chamfer Distance (CD) as a proxy reconstruction loss,
\begin{align}
    \lefteqn{\mathcal{L}_\text{recon}(\mathbf{x}) = \textnormal{CD}(\mathbf{x}, \hat{\mathbf{x}})} \nonumber\\
    &= \sum_{i} \min_j \| \mathbf{x}_i - \hat{\mathbf{x}}_j\|_2^2 + \sum_j \min_i \|\mathbf{x}_i - \hat{\mathbf{x}}_j\|_2^2.
    \label{eqn:reconstruction_cd}
\end{align}

The CD may not admit a direct interpretation as a negative log-likelihood of $p_\theta(\mathbf{x}|\mathbf{z})$, but shares the optimum with the matching distance having a proper interpretation. 
By employing the CD for the reconstruction loss, we learn the VAE with a surrogate for the likelihood $p_\theta(\mathbf{x}|\mathbf{z})$.
CD requires $O(n^2)$ computation time, so is scalable to the moderately large sets.
Note also that the CD should be scaled appropriately to match the likelihood induced by optimal matching distance.
We implicitly account for this by applying weights to KL divergence in our final objective function:
\begin{equation}
    \mathcal{L}_{\textnormal{HSVAE}}(\mathbf{x}) = \mathcal{L}_{\textnormal{recon}}(\mathbf{x}) + \beta\mathcal{L}_{\textnormal{KL}}(\mathbf{x}),
    \label{eqn:beta_vae_loss}
\end{equation}
where $\mathcal{L}_{\textnormal{KL}}(\mathbf{x})$ is the KL divergence in Eq.~\eqref{eqn:elbo_hierarchicalSetVAE}.

%% file: arxiv/related_work.tex
\section{Related Work}
\label{sec:related}
\input{arxiv/table} 

\paragraph{Set generative modeling.} SetVAE is closely related to recent works on permutation-equivariant set prediction \cite{zhang2020deep, kosiorek2020conditional, carion2020endtoend, locatello2020objectcentric, li2020exchangeable}.
Closest to our approach is the autoencoding TSPN \cite{kosiorek2020conditional} that uses a stack of ISABs \cite{lee2019set} to predict a set from randomly initialized elements.
However, TSPN does not allow sampling, as it uses a pooling-based deterministic set encoding (FSPool) \cite{zhang2020fspool} for reconstruction.
SetVAE instead discards FSPool and access projected sets in ISAB directly, which allows an efficient variational inference and a direct extension to hierarchical multi-scale latent.

Our approach differs from previous generative models treating each element \emph{i.i.d.} and processing a random initial set with an elementwise function \cite{edwards2017neural, yang2019pointflow, kim2020softflow}.
Notably, PointFlow \cite{yang2019pointflow} uses a continuous normalizing flow (CNF) to process a 3D Gaussian point cloud into an object.
However, assuming elementwise independence could pose a limit in modeling complex element interactions.
Also, CNF requires the invertibility of the generative model, which could further limit its expressiveness.
SetVAE resolves this problem by adopting permutation equivariant ISAB that models inter-element interactions via attention, and a hierarchical VAE framework with flexible latent dependency.

Contrary to previous works specifically designed for a certain type of set-structured data (\emph{e.g.}, point cloud~\cite{achlioptas2018learning, li2018point, yang2019pointflow}), we emphasize that SetVAE can be trivially applied to arbitrary set-structured data. 
We demonstrate this by applying SetVAE to the generation of a scene layout represented by a set of object bounding boxes.


\cutparagraphup
\paragraph{Hierarchical VAE.}
Our model is built upon the prior works on hierarchical VAEs for images \cite{johnson2018structured}, such as Ladder-VAE \cite{sonderby2016ladder}, IAF-VAE \cite{kingma2017improving}, and NVAE \cite{vahdat2020nvae}.
To model long-range pixel correlations in images, these models organize latent variables at each hierarchy as images while gradually increasing their resolution via upsampling.
However, the requirement for permutation equivariance has prevented applying multi-scale approaches to sets.
ABLs in SetVAE solve this problem by defining latent variables in the projected scales of each hierarchy.
\\
\\

%% file: arxiv/table.tex
\begin{table*}[htbp!]
\caption{Comparison against the state-of-the-art generative models. $\uparrow$: the higher the better.
$\downarrow$: the lower the better. The best scores are highlighted in bold. MMD-CD is scaled by $10^3$, and MMD-EMD by $10^2$.}
\vspace{-0.1in}
\footnotesize
\centering
\begin{adjustbox}{width=0.99\textwidth}
\label{table:pointeval}
\centering
\begin{tabular}{clccccccccc}
\Xhline{2\arrayrulewidth}
\\[-1em]& & \multicolumn{2}{c}{\# Parameters (M)} & \multicolumn{2}{c}{MMD($\downarrow$)} & \multicolumn{2}{c}{COV($\%$,$\uparrow$)} &  \multicolumn{2}{c}{1-NNA($\%$,$\downarrow$)} \\
\\[-1em]\cline{3-4} \cline{5-6} \cline{7-8} \cline{9-10}
\\[-1em]Category & Model & Full & Gen & CD & EMD & CD & EMD & CD & EMD \\
\\[-1em]\Xhline{2\arrayrulewidth}
\\[-1em]\multirow{7}{*}{\shortstack{Airplane}}
    & l-GAN (CD)\cite{yang2019pointflow}    & 1.97 & 1.71 & 0.239 & 4.27 & 43.21 & 21.23 & 86.30 & 97.28\\
    & l-GAN (EMD)\cite{yang2019pointflow}   & 1.97 & 1.71 & 0.269 & 3.29 & \textbf{47.90} & \textbf{50.62} & 87.65 & 85.68\\
    & PC-GAN\cite{yang2019pointflow}        & 9.14 & 1.52 & 0.287 & 3.57 & 36.46 & 40.94 & 94.35 & 92.32\\
    & PointFlow\cite{yang2019pointflow}     & 1.61 & 1.06 & 0.217 & 3.24 & 46.91 & 48.40 & 75.68 & \textbf{75.06}\\
    & SetVAE (Ours) & \textbf{0.75} & \textbf{0.39} & \textbf{0.199} & \textbf{3.07} & 43.45 & 44.93 & \textbf{75.31} & 77.65\\
    \\[-1em]\cline{2-10}
    \\[-1em]& Training set & - & - & 0.226 & 3.08 & 42.72 & 49.14 & 70.62 & 67.53 \\
\\[-1em]\Xhline{2\arrayrulewidth}
\\[-1em]\multirow{7}{*}{\shortstack{Chair}}
    & l-GAN (CD)\cite{yang2019pointflow} & 1.97 & 1.71 & 2.46 & 8.91 & 41.39 & 25.68 & 64.43 & 85.27\\
    & l-GAN (EMD)\cite{yang2019pointflow} & 1.97 & 1.71 & 2.61 & 7.85 & 40.79 & 41.69 & 64.73 & 65.56\\
    & PC-GAN\cite{yang2019pointflow} & 9.14 & 1.52 & 2.75 & 8.20 & 36.50 & 38.98 & 76.03 & 78.37\\
    & PointFlow\cite{yang2019pointflow} & 1.61 & 1.06 & \textbf{2.42} & 7.87 & 46.83 & \textbf{46.98} & 60.88 & \textbf{59.89} \\
    & SetVAE (Ours) & \textbf{0.75} & \textbf{0.39} & 2.55 & \textbf{7.82} & \textbf{46.98} & 45.01 & \textbf{58.76} & 61.48\\
    \\[-1em]\cline{2-10}
    \\[-1em]& Training set & - & - & 1.92 & 7.38 & 57.25 & 55.44 & 59.67 & 58.46 \\
\\[-1em]\Xhline{2\arrayrulewidth}
\\[-1em]\multirow{7}{*}{\shortstack{Car}}
    & l-GAN (CD)\cite{yang2019pointflow} & 1.97 & 1.71 & 1.55 & 6.25 & 38.64 & 18.47 & 63.07 & 88.07\\
    & l-GAN (EMD)\cite{yang2019pointflow} & 1.97 & 1.71 & 1.48 & 5.43 & 39.20 & 39.77 & 69.74 & 68.32\\
    & PC-GAN\cite{yang2019pointflow} & 9.14 & 1.52 & 1.12 & 5.83 & 23.56 & 30.29 & 92.19 & 90.87\\
    & PointFlow\cite{yang2019pointflow} & 1.61 & 1.06 & 0.91 & 5.22 & 44.03 & \textbf{46.59} & 60.65 & \textbf{62.36} \\
    & SetVAE (Ours) & \textbf{0.75} & \textbf{0.39} & \textbf{0.88} & \textbf{5.05} & \textbf{48.58} & 44.60 & \textbf{59.66} & 63.35\\
    \\[-1em]\cline{2-10}
    \\[-1em]& Training set & - & - & 1.03 & 5.33 & 48.30 & 51.42 & 57.39 & 53.27 \\
\\[-1em]\Xhline{2\arrayrulewidth}
\end{tabular}
\end{adjustbox}
\vspace{-0.1in}
\end{table*}

%% file: arxiv/experiment.tex
\cutsectionup
\section{Experiments}
\label{sec:experiment}

\begin{figure}[t!]
    \centering
    \includegraphics[width=0.47\textwidth]{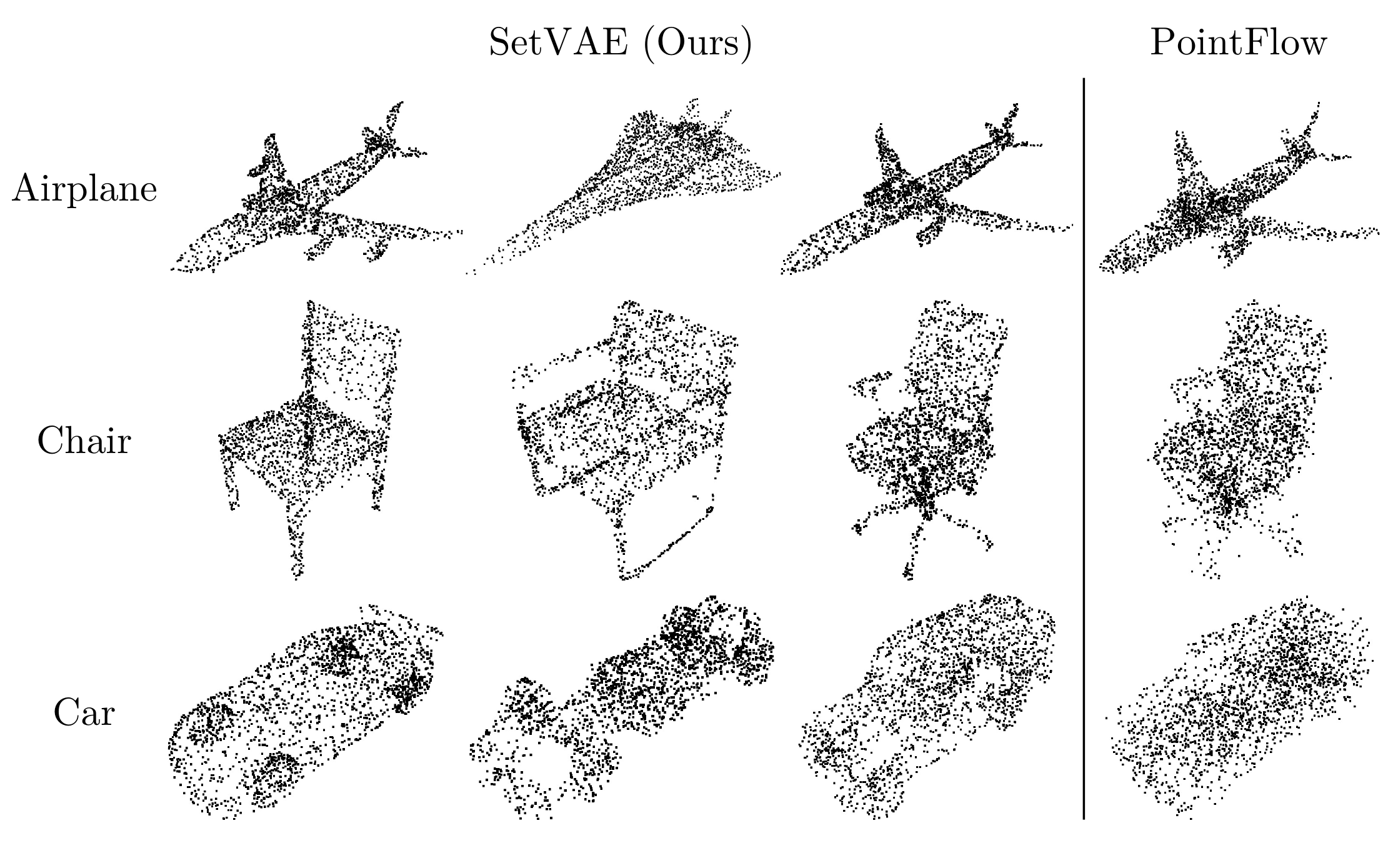}
    \caption{Examples of randomly generated point sets from SetVAE (ours) and PointFlow in ShapeNet. 
    }
    \label{fig:compare_pointflow}
    \vspace{-0.65cm}
\end{figure}

\subsection{Experimental Setup}
\label{sec:quantitative}

\cutparagraphup
\paragraph{Dataset}
We examine SetVAE using ShapeNet~\cite{chang2015shapenet}, Set-MNIST~\cite{zhang2020deep}, and Set-MultiMNIST~\cite{eslami2016attend} datasets.
For ShapeNet, we follow the prior work using 2048 points sampled uniformly from the mesh surface~\cite{yang2019pointflow}.
For Set-MNIST, we binarized the images in MNIST and scaled the coordinates to $[0, 1]$ \cite{kosiorek2020conditional}.
Similarly, we build Set-MultiMNIST using $64\times64$ images of MultiMNIST~\cite{eslami2016attend} with two digits randomly located without overlap.

\cutparagraphup
\paragraph{Evaluation Metrics}
For evaluation in ShapeNet, we compare the standard metrics including Minimum Matching Distance (MMD), Coverage (COV), and 1-Nearest Neighbor Accuracy (1-NNA), where the similarity between point clouds are computed with Chamfer Distance (CD) (Eq.~\eqref{eqn:reconstruction_cd}), and Earth Mover's Distance (EMD) based on optimal matching.
The details are in the supplementary file.

\begin{figure}[!t]
    \centering
    \includegraphics[width=0.9\linewidth]{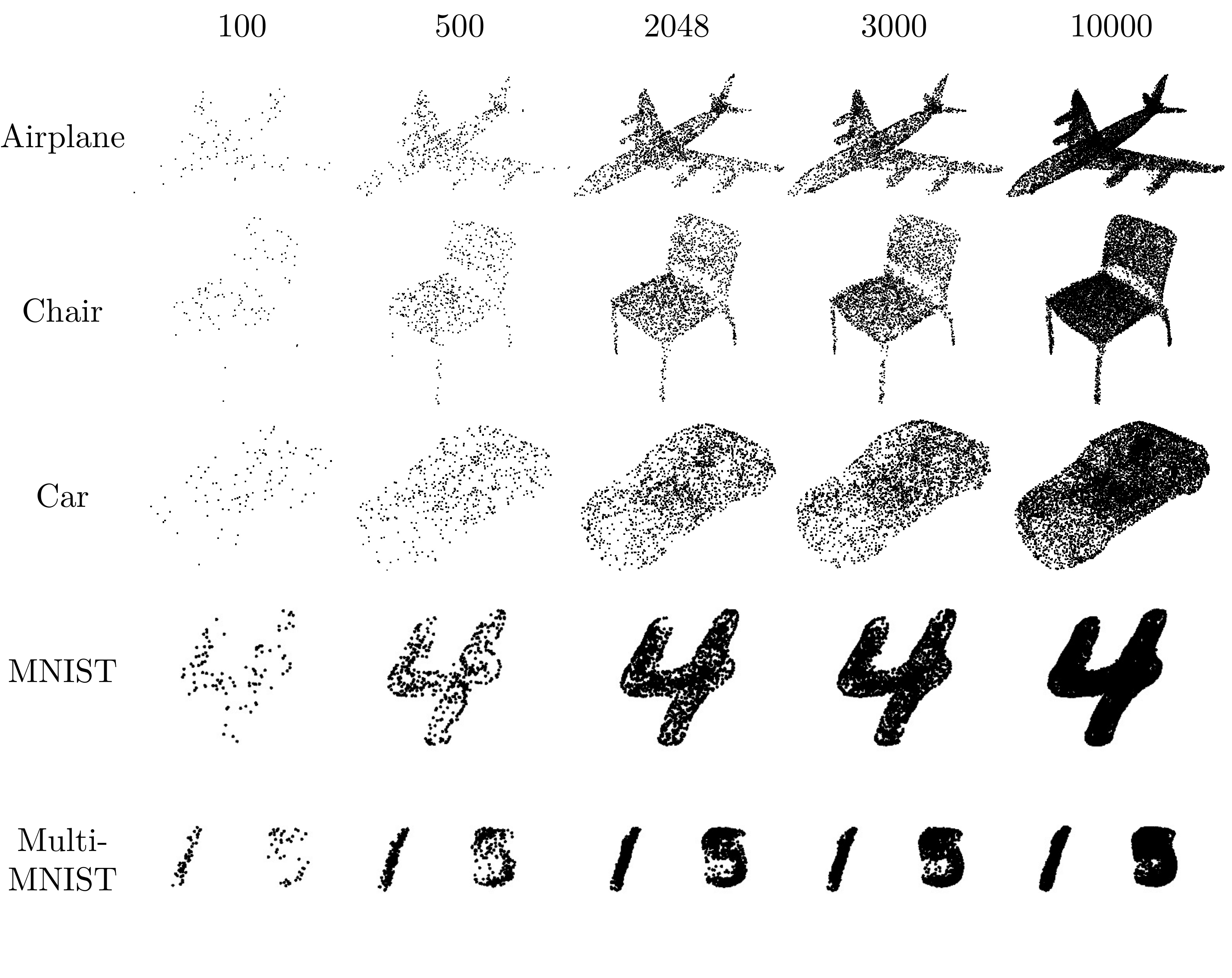}
    \vspace{-0.7cm}
    \caption{
    Samples from SetVAE for different cardinalities. At each row, the hierarchical latent variables are fixed and the initial set is re-sampled with different cardinality.
    }
    \label{fig:cardinality}
    \vspace{-0.5cm}
\end{figure}
\begin{figure}[!t]
    \centering
    \includegraphics[width=0.8\linewidth]{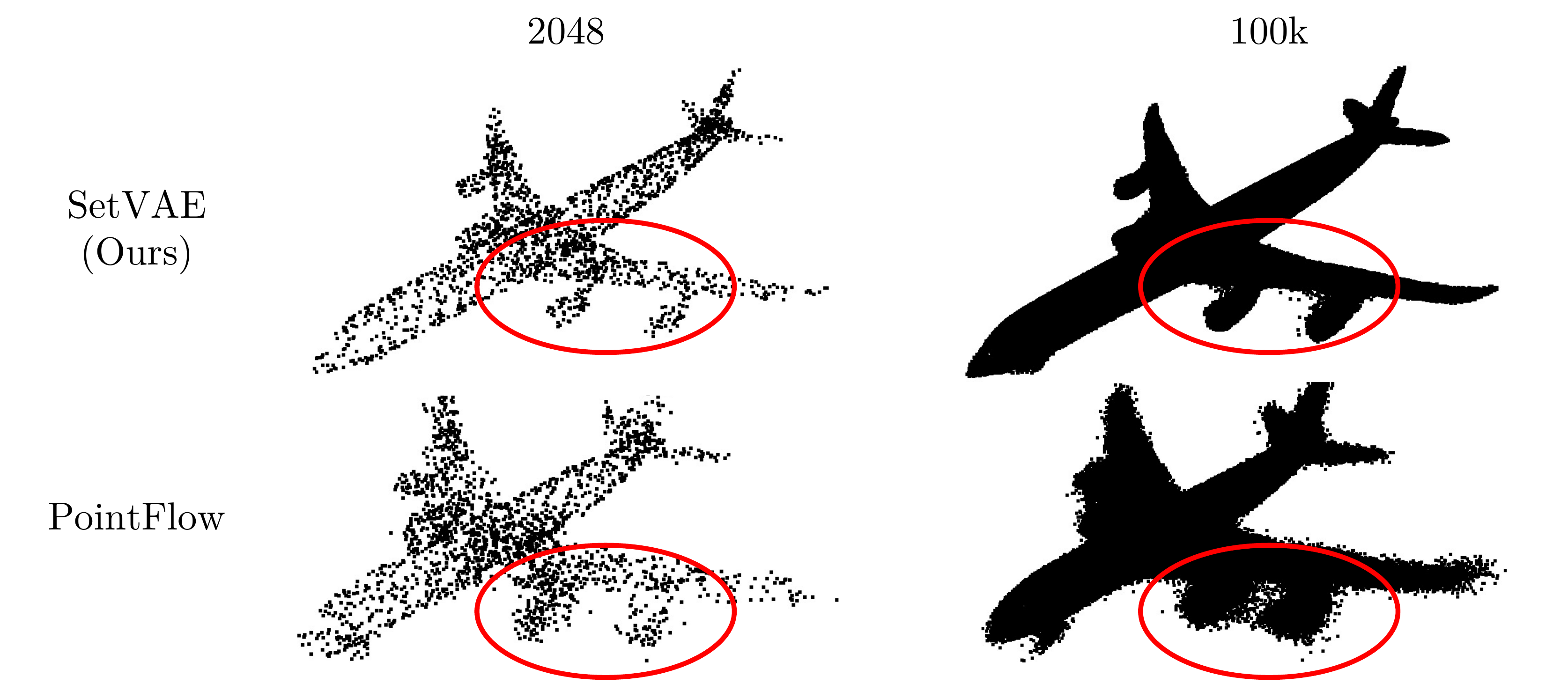}
    \vspace{-0.15cm}
    \caption{
    Samples from SetVAE and PointFlow in a high cardinality setting. Only the initial sets are re-sampled.
    }
    \label{fig:mega-cardinality}
    \vspace{-0.5cm}
\end{figure}

\subsection{Comparison to Other Methods}
We compare SetVAE with the state-of-the-art generative models for point clouds including l-GAN~\cite{achlioptas2018learning}, PC-GAN~\cite{li2018point}, and PointFlow~\cite{yang2019pointflow}.
Following these works, we train our model for each category of airplane, chair, and car.

Table~\ref{table:pointeval} summarizes the evaluation result.
SetVAE achieves better or competitive performance to the prior arts using a \emph{much smaller} number of parameters ($8\%$ to $45\%$ of competitors).
Notably, SetVAE often outperforms PointFlow with a substantial margin in terms of minimum matching distance (MMD) and has better or comparable coverage (COV) and 1-NNA.
Lower MMD indicates that SetVAE generates high-fidelity samples, and high COV and low 1-NNA indicate that SetVAE generates diverse samples covering various modes in data.
Together, the results indicate that SetVAE generates realistic, high-quality point sets.
Notably, we find that SetVAE trained with CD (Eq.~\eqref{eqn:reconstruction_cd}) generalizes well to EMD-based metrics than l-GAN.

We also observe that SetVAE is significantly faster than PointFlow in both training (56$\times$ speedup; 0.20s vs. 11.2s) and testing (68$\times$ speedup; 0.052s vs. 3.52s)\footnote{Measured on a single GTX 1080ti with a batch size of 16.}.
It is because PointFlow requires a costly ODE solver for both training and inference, and has much more parameters.

Figure \ref{fig:compare_pointflow} illustrates qualitative comparisons.
Compared to PointFlow, we observe that SetVAE generates sharper details, especially in small object parts such as wings and engines of an airplane or wheels of a car.
We conjecture that this is because our model generates samples considering inter-element dependency while capturing shapes in various granularities via a hierarchical latent structure.

\subsection{Internal Analysis}
\label{sec:qualitative}

\begin{figure}[!t]
    \centering
    \includegraphics[width=1.0\linewidth]{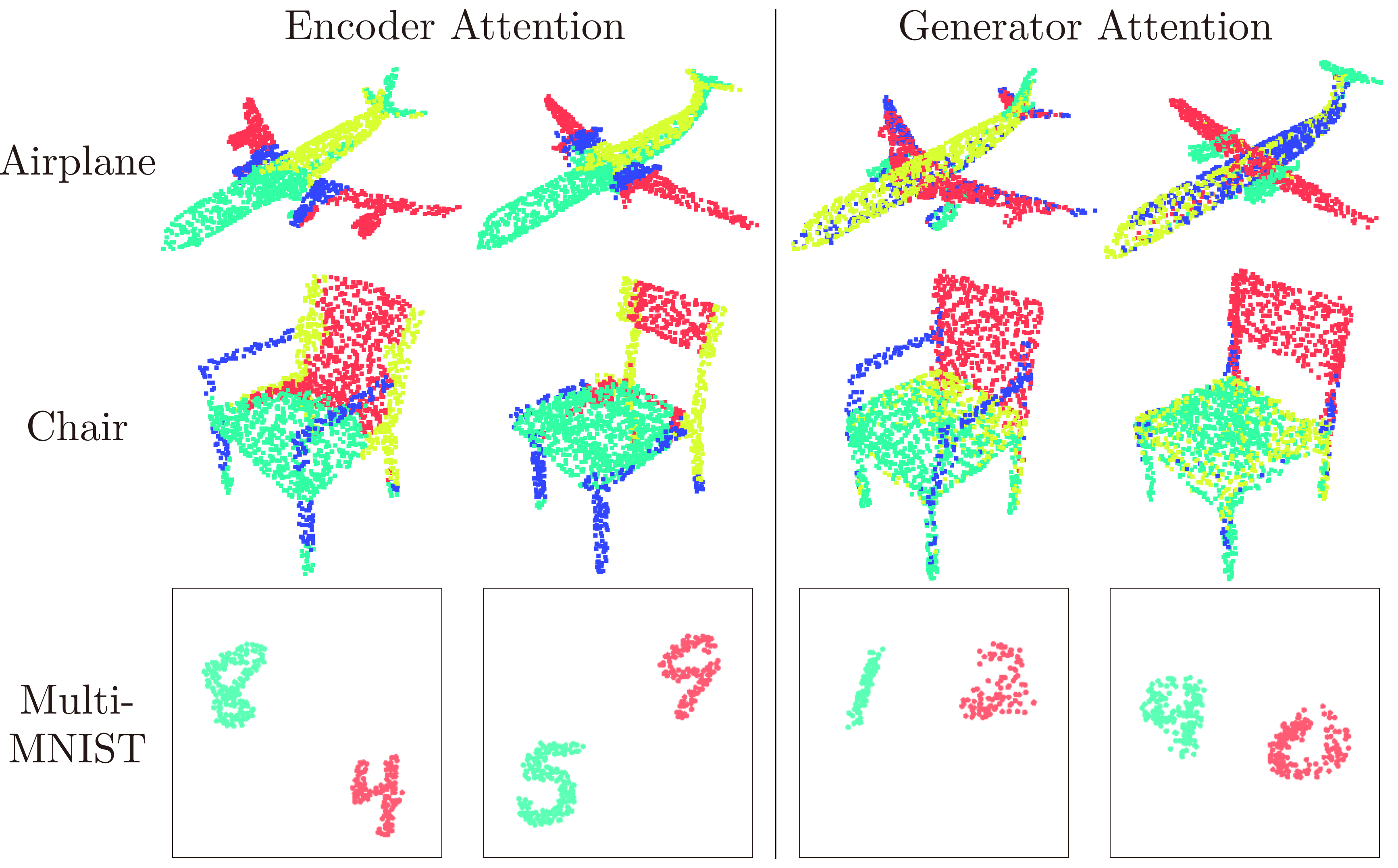}
    \vspace{-0.7cm}
    \caption{
    Attention visualization at a selected layer.
    Each point is color-coded by its assignment based on attention.
    }
    \label{fig:subset}
    \vspace{-0.6cm}
\end{figure}

\paragraph{Cardinality disentanglement}
Ideally, a generative model for sets should be able to disentangle the cardinality of a set from the rest of the generative factors (\emph{e.g.}, structure, style). 
SetVAE partly achieves this by decomposing the latent variables into the initial set $\mathbf{z}^{(0)}$ and the hierarchical latent variables $\mathbf{z}^{(1:L)}$.
To validate this, we generate samples by changing the initial set's cardinality while fixing the rest.
Figure~\ref{fig:cardinality} illustrates the result.
We observe that SetVAE generates samples having consistent global structure with a varying number of elements.
Surprisingly, it even generalizes well to the cardinalities not seen during training.
For instance, the model generalizes to any cardinality between 100 and 10,000, although it is trained with only 2048 points in ShapeNet and less than 250 points in Set-MNIST.
It shows that SetVAE disentangles the cardinality from the factors characterizing an object.

We compare SetVAE to PointFlow in extremely high cardinality setting (100k points) in Figure~\ref{fig:mega-cardinality}.
Although PointFlow innately disentangles cardinality by modeling each element independently, we observe that it tends to generate noisy, blurry boundaries in large cardinality settings.
In contrast, SetVAE retains the sharpness of the structure even for extreme cardinality, presumably because it considers inter-element dependency in the generation process.

\begin{figure}[!t]
    \centering
    \includegraphics[width=0.83\linewidth]{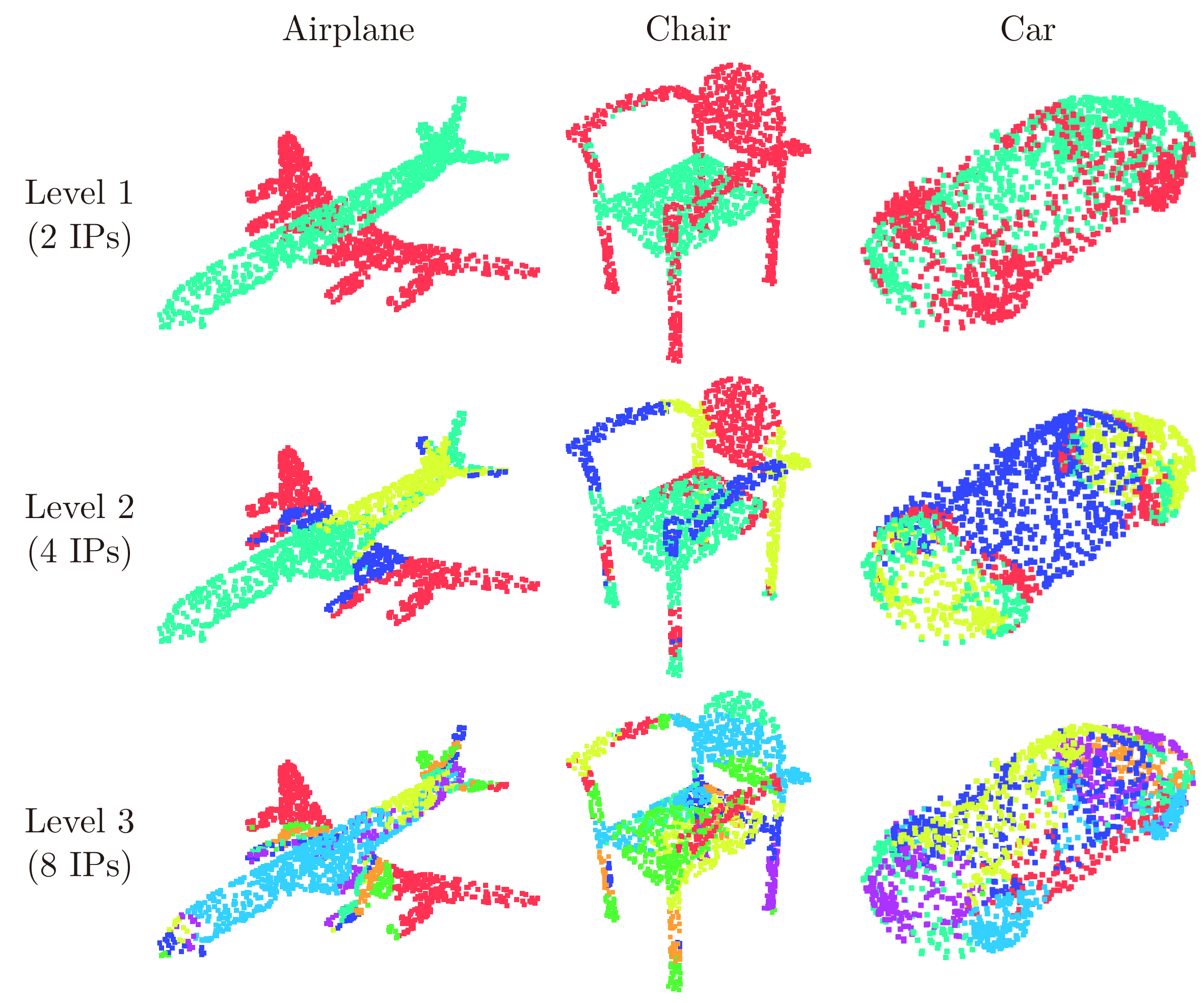}
    \vspace{-0.35cm}
    \caption{Visualization of encoder attention across multiple layers.
    IP notes the number of inducing points at each level.
    See the supplementary file for more results.
    }
    \label{fig:composition}
    \vspace{-0.35cm}
\end{figure}
\begin{figure}[!t]
\centering
    \includegraphics[width=0.52\linewidth]{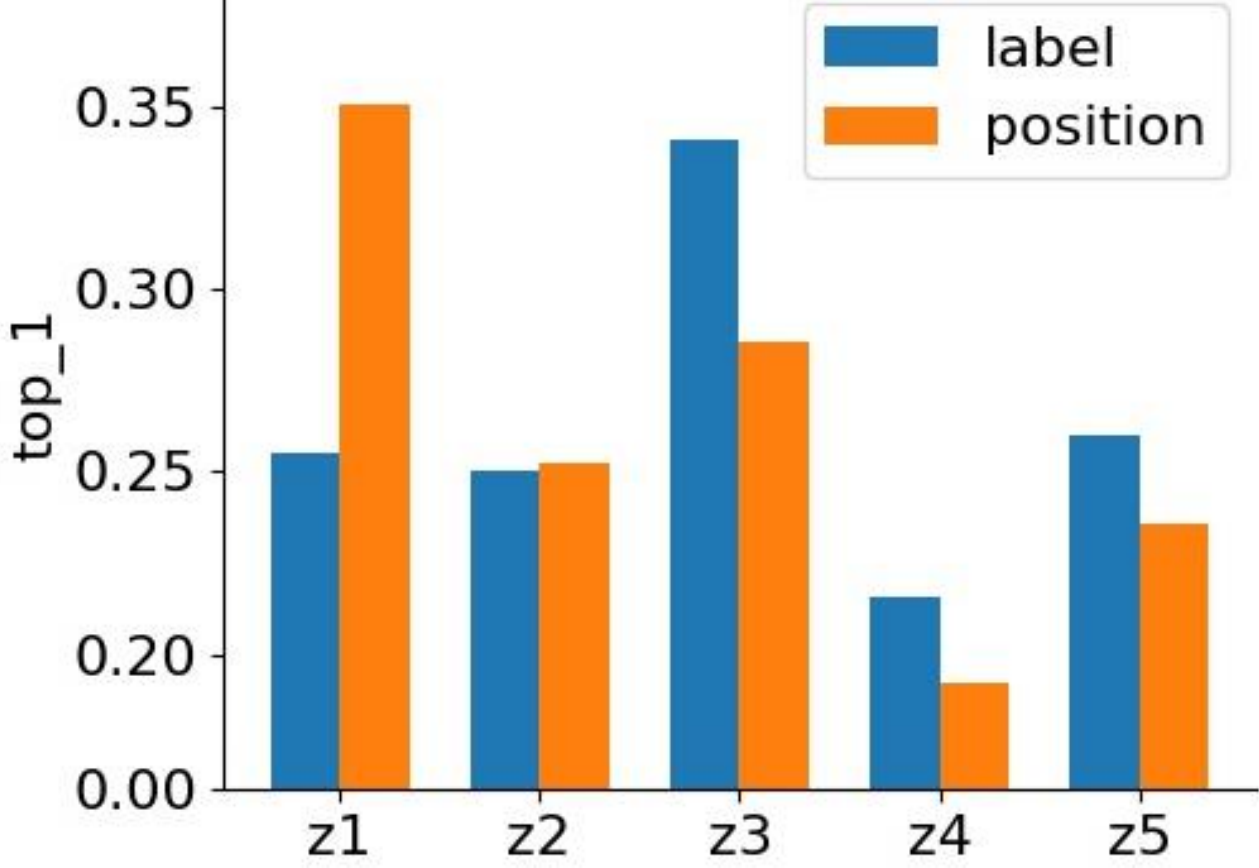}
    \vspace{-0.3cm}
    \caption{
    Layer-wise classification results using the hierarchical latent variables in Set-MultiMNIST dataset.
    }
    \label{fig:lda}
    \vspace{-0.6cm}
\end{figure}

\cutparagraphup
\paragraph{Discovering coarse-to-fine dependency}
SetVAE discovers interesting subset structures via hierarchical bottlenecks in ISAB and ABL.
To demonstrate this, we visualize the encoder attention (Eq.~\eqref{eqn:isab_bottleneck}) and generator attention (Eq.~\eqref{eqn:abl_bottleneck}) in Figure~\ref{fig:subset}, where each point is color-coded based on its hard assignment to one of $m$ inducing points\footnote{For illustrative purposes, we present results from a selected head.}.
We observe that SetVAE attends to semantically interesting parts consistently across samples, such as wings and engines of an airplane, legs and backs of a chair, and even different instances of multiple digits.

Figure~\ref{fig:composition} illustrates the point-wise attention across levels.
We observe that the top-level tends to capture the coarse and symmetric structures, such as wings and wheels, which are further decomposed into much finer granularity in the subsequent levels.
We conjecture that this coarse-to-fine subset dependency helps the model to generate accurate structure in various granularity from global structure to local details.

Interestingly, we find that the hierarchical structure of SetVAE sometimes leads to the disentanglement of generative factors across layers.
To demonstrate this, we train two classifiers in Set-MultiMNIST, one for digit class and the other for their positions.
We train the classifiers using latent variables at each generator layer as an input, and measure the accuracy at each layer.
In Figure~\ref{fig:lda}, the latent variables at the lower layers tend to contribute more in locating the digits, while higher layers contribute to generating shape.

\cutparagraphup
\paragraph{Ablation study}
Table~\ref{table:ablation} summarizes the ablation study of SetVAE (see the supplementary file for qualitative results and evaluation detail).
We consider two baselines: Vanilla SetVAE using a global latent variable $\mathbf{z}^{(1)}$ (Section~\ref{sec:method}), and hierarchical SetVAE with unimodal prior (Section~\ref{sec:implementation}).

The Vanilla SetVAE performs much worse than our full model.
We conjecture that a single latent variable is not expressive enough to encode complex variations in MultiMNIST, such as identity and position of multiple digits.
We also find that multi-modal prior stabilizes the training of attention and guides the model to better local optima.

\begin{table}[t!]
\begin{center}
\caption{Ablation study performed on Set-MultiMNIST dataset using FID scores for {$64\times 64$} rendered images.}
\vspace{-0.2cm}
\footnotesize
\label{table:ablation}
\begin{tabular}{lc}
\Xhline{2\arrayrulewidth}
\\[-1em] Model & FID($\downarrow$) \\
\\[-1em]\Xhline{2\arrayrulewidth}
\\[-1em]
SetVAE (Ours) & \textbf{1047} \\
Non-hierarchical & 1470 \\
Unimodal prior & 1252 \\
\\[-1em]\Xhline{2\arrayrulewidth}
\end{tabular}
\end{center}
\vspace{-0.4cm}
\end{table}

\begin{figure}[!t]
    \centering
    \includegraphics[width=1\linewidth]{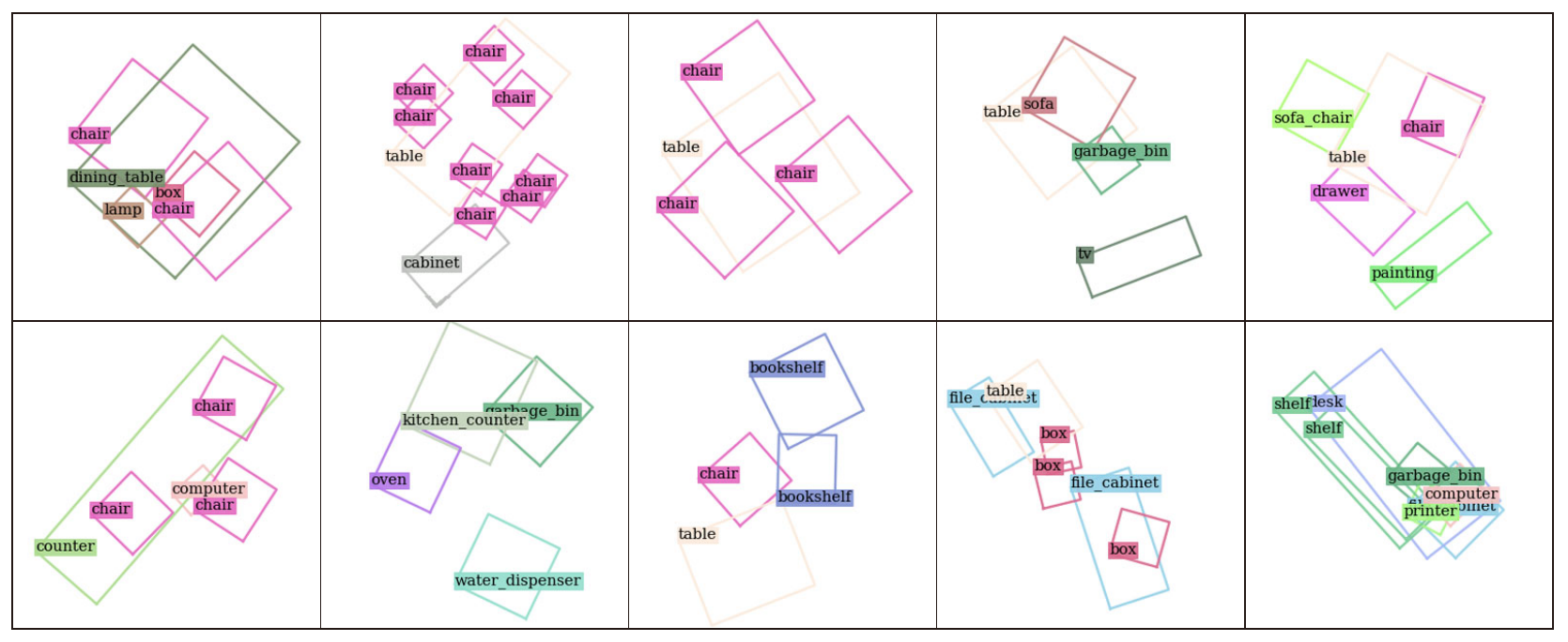}
    \vspace{-0.25in}
    \caption{Generation results from SetVAE trained on SUN-RGBD dataset. \textbf{Zoom-in} for a better view.}
\label{fig:sunrgbd}
\vspace{-0.15in}
\end{figure}

\cutparagraphup
\paragraph{Extension to categorical bounding boxes}
SetVAE provides a solid exchangeability guarantee over sets, thus applicable to any set-structured data.
To demonstrate this, we trained SetVAE on categorical bounding boxes in indoor scenes of the SUN-RGBD dataset~\cite{song2015sun}.
As shown in Figure~\ref{fig:sunrgbd}, SetVAE generates plausible layouts, modeling a complicated distribution of discrete semantic categories and continuous spatial instantiation of objects.

%% file: arxiv/conclusion.tex
\vspace*{-0.1in}
\section{Conclusion}
\label{sec:conclusion}
\vspace*{-0.05in}
We introduced SetVAE, a novel hierarchical VAE for sets of varying cardinality.
Introducing a novel bottleneck equivariant layer that learns subset representations, SetVAE performs hierarchical subset reasoning to encode and generate sets in a coarse-to-fine manner.
As a result, SetVAE generates high-quality, diverse sets with reduced parameters.
We also showed that SetVAE achieves cardinality disentanglement and generalization.

%% file: arxiv/supp.tex
\section{Implementation details}
In this section, we discuss detailed derivations and descriptions of SetVAE presented in Section~\ref{sec:method} and Section~\ref{sec:architecture}.

\subsection{KL Divergence of Initial Set Distribution}
\label{appendix:initial_kl}
This section provides proof that the KL divergence between the approximate posterior and the prior of the initial elements in Eq.~\eqref{eqn:elbo_vanillaSetVAE} and \eqref{eqn:elbo_hierarchicalSetVAE} is a constant.

Following the definition in Eq.~\eqref{eqn:prior_initialset} and Eq.~\eqref{eqn:posterior_vanillaSetVAE}, we decompose the prior as $p(\mathbf{z}^{(0)}) = p(n)p(\mathbf{z}^{(0)}|n)$ and the approximate posterior as $q(\mathbf{z}^{(0)}|\mathbf{x}) = \delta(n)q(\mathbf{z}^{(0)}|n, \mathbf{x})$, where $\delta(n)$ is defined as a delta function centered at $n=|\mathbf{x}|$.
Here, the conditionals are given by
\begin{align}
     p(\mathbf{z}^{(0)}|n) &= \prod_{i=1}^n{p(\mathbf{z}_i^{(0)})}, \\
     q(\mathbf{z}^{(0)}|n, \mathbf{x}) &= \prod_{i=1}^n{q(\mathbf{z}_i^{(0)}|\mathbf{x})}.
\end{align}
As described in the main text, we set the elementwise distributions identical, $p(\mathbf{z}_i^{(0)}) = q(\mathbf{z}_i^{(0)}|\mathbf{x})$.
This renders the conditionals equal,
\begin{equation}
    p(\mathbf{z}^{(0)}|n) = q(\mathbf{z}^{(0)}|n, \mathbf{x}).\label{eqn:initial_set_equal}
\end{equation}
Then, the KL divergence between the approximate posterior and the prior in Eq.~\eqref{eqn:elbo_vanillaSetVAE} is written as
\begin{align}
    \textnormal{KL}&(q(\mathbf{z}^{(0)}|\mathbf{x})||p(\mathbf{z}^{(0)})) \nonumber\\
    &= \textnormal{KL}(\delta(n)q(\mathbf{z}^{(0)}|n, \mathbf{x})||p(n)p(\mathbf{z}^{(0)}|n)) \\
    &= \textnormal{KL}(\delta(n)p(\mathbf{z}^{(0)}|n)||p(n)p(\mathbf{z}^{(0)}|n)), \label{eqn:kl_initial_set}
\end{align}
where the second equality comes from the Eq.~\eqref{eqn:initial_set_equal}.
From the definition of KL divergence, we can rewrite Eq.~\eqref{eqn:kl_initial_set} as
\begin{align}
    \textnormal{KL}&(q(\mathbf{z}^{(0)}|\mathbf{x})||p(\mathbf{z}^{(0)})) \nonumber\\
    &= \mathbb{E}_{\delta(n)p(\mathbf{z}^{(0)}|n)}{\left[\log{\frac{\delta(n)p(\mathbf{z}^{(0)}|n)}{p(n)p(\mathbf{z}^{(0)}|n)}}\right]} \\
    &= \mathbb{E}_{\delta(n)p(\mathbf{z}^{(0)}|n)}{\left[\log{\frac{\delta(n)}{p(n)}}\right]}, \label{eqn:kl_initial_set_reduced}
\end{align}
As the logarithm in Eq.~\eqref{eqn:kl_initial_set_reduced} does not depend on $\mathbf{z}^{(0)}$, we can take it out from the expectation over $p(\mathbf{z}^{(0)}|n)$ as follows:
\begin{align}
    \textnormal{KL}&(q(\mathbf{z}^{(0)}|\mathbf{x})||p(\mathbf{z}^{(0)})) \nonumber\\
    &= \mathbb{E}_{\delta(n)}{\left[\mathbb{E}_{p(\mathbf{z}^{(0)}|n)} {\left[\log{\frac{\delta(n)}{p(n)}}\right]}\right]} \\
    &= \mathbb{E}_{\delta(n)}{\left[\log{\frac{\delta(n)}{p(n)}}\right]}, \label{eqn:kl_initial_set_cardinality}
\end{align}
which can be rewritten as
\begin{align}
    \mathbb{E}_{\delta(n)}{\left[\log{\delta(n)} - \log{p(n)}\right]}.
\end{align}
The expectation over the delta function $\delta(n)$ is simply an evaluation at $n = |\mathbf{x}|$.
As $\delta$ is defined over a discrete random variable $n$, its probability mass at the center $|\mathbf{x}|$ equals 1.
Therefore, $\log{\delta(n)}$ at $n = |\mathbf{x}|$ reduces to $\log1 = 0$, and we obtain
\begin{align}
    \textnormal{KL}&(q(\mathbf{z}^{(0)}|\mathbf{x})||p(\mathbf{z}^{(0)})) = - \log p(|\mathbf{x}|). \label{eqn:kl_constant}
\end{align}
As discussed in the main text, we model $p(n)$ using the empirical distribution of data cardinality.
Thus, $p(|\mathbf{x}|)$ only depends on data distribution, and $- \textnormal{KL}(q(\mathbf{z}^{(0)}|\mathbf{x})||p(\mathbf{z}^{(0)}))$ in Eq.~\eqref{eqn:elbo_vanillaSetVAE} can be omitted from optimization.

\subsection{Implementation of SetVAE}
\label{appendix:abl}

\begin{figure}[!t]
    \centering
    \vspace{0.2in}
    \begin{subfigure}[b]{0.395\linewidth}
        \centering
        \includegraphics[width=0.95\linewidth]{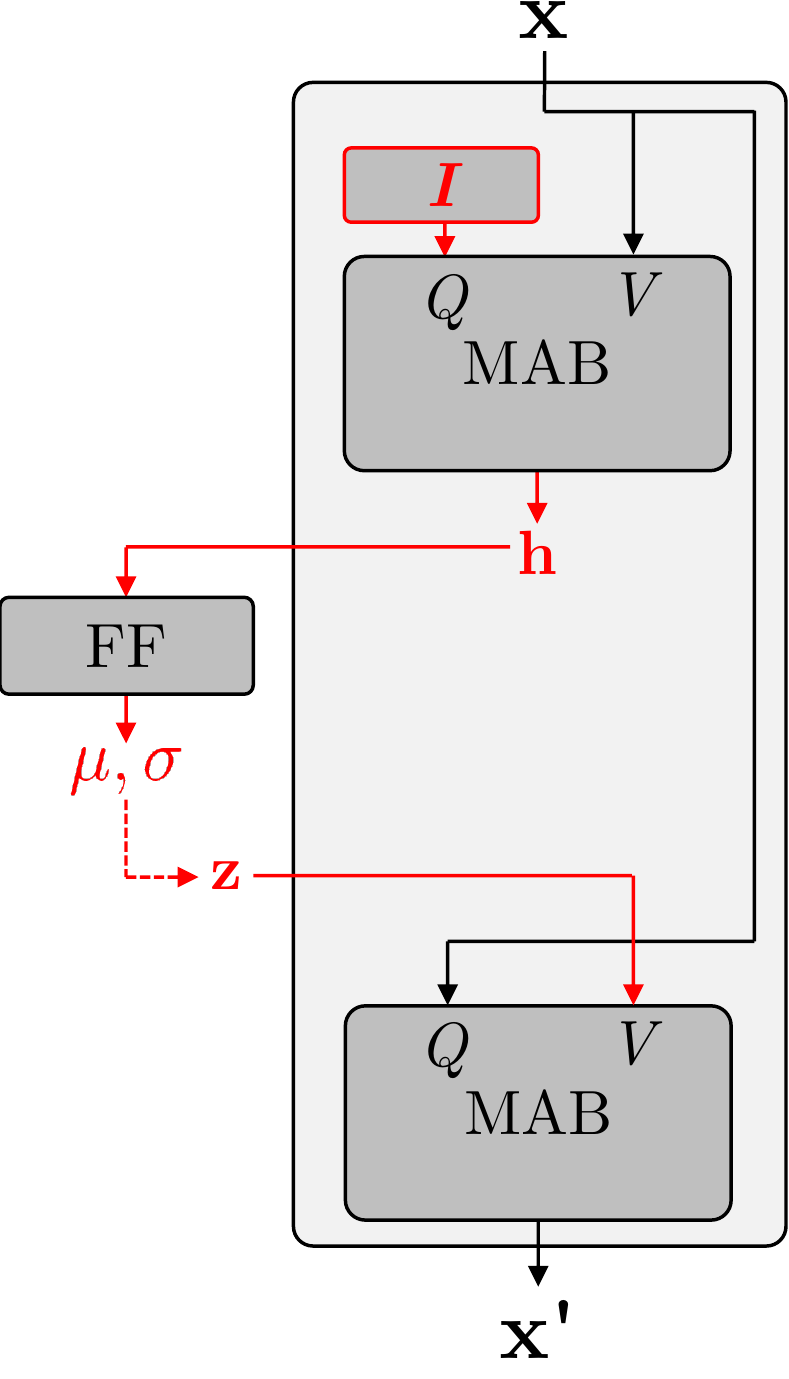}
        \caption{Generation}
        \label{fig:abl_prior}
    \end{subfigure}
    \begin{subfigure}[b]{0.59\linewidth}
        \centering
        \includegraphics[width=0.95\linewidth]{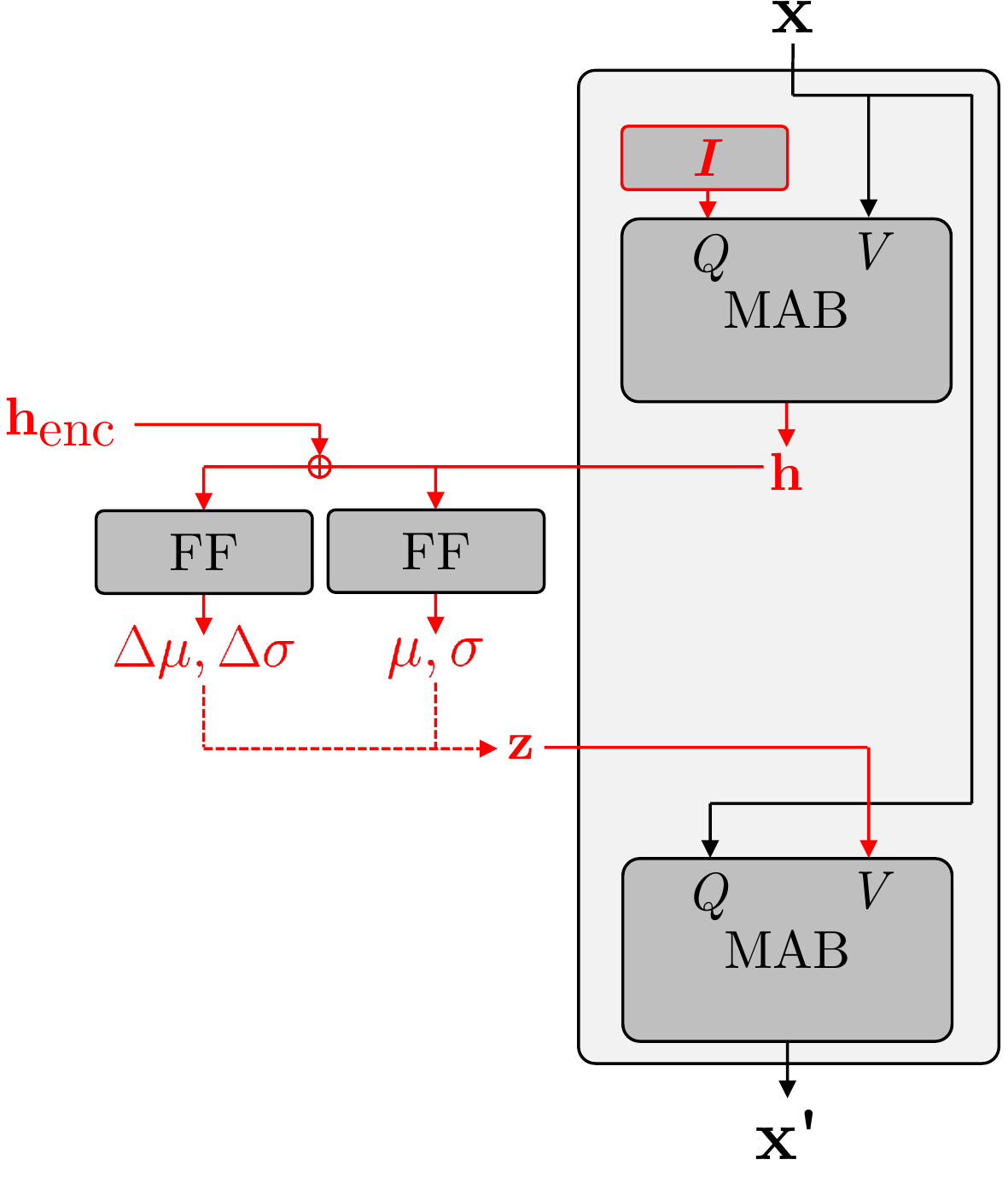}
        \caption{Inference}
        \label{fig:abl_posterior}
    \end{subfigure}
    \caption{The detailed structure of Attentive Bottleneck Layer during sampling (for generation) and inference (for reconstruction).}
\label{fig:abl_structure}
\end{figure}

\paragraph{Attentive Bottleneck Layer.}
In Figure~\ref{fig:abl_structure}, we provide the detailed structure of Attentive Bottleneck Layer (ABL) that composes the top-down generator of SetVAE (Section~\ref{sec:architecture}).
We share the parameters in ABL for generation and inference, which is known to be effective in stabilizing the training of hierarchical VAE~\cite{kingma2017improving, vahdat2020nvae}.

During generation (Figure~\ref{fig:abl_prior}), $\mathbf{z}$ is sampled from a Gaussian prior $\mathcal{N}(\mu, \sigma)$ (Eq.~\eqref{eqn:abp_z_prior}).
To predict $\mu$ and $\sigma$ from $\mathbf{h}$, we use an elementwise fully-connected ($\textnormal{FC}$) layer, of which parameters are shared across elements of $\mathbf{h}$.
During inference, we sample the latent variables from the approximate posterior $\mathcal{N}(\mu+\Delta\mu, \sigma\cdot\Delta\sigma)$, where the correction factors $\Delta\mu, \Delta\sigma$ are predicted from the bottom-up encoding $\mathbf{h}_\textnormal{enc}$ by an additional $\textnormal{FC}$ layer.
Note that the $\textnormal{FC}$ for predicting $\mu, \sigma$ is shared for generation and inference, but the $\textnormal{FC}$ that predicts $\Delta\mu, \Delta\sigma$ is used only for inference.
\\

\paragraph{Slot Attention in ISAB and ABL.}
SetVAE discovers subset representation via projection attention in ISAB (Eq.~\eqref{eqn:isab_bottleneck}) and ABL (Eq.~\eqref{eqn:abl_bottleneck}).
However, with a basic attention mechanism, the projection attention may ignore some parts of input by simply not attending to them.
To prevent this, in both ISAB and ABL, we change the projection attention to Slot Attention \cite{locatello2020objectcentric}.

Specifically, plain projection attention\footnote{For simplicity, we explain with single-head attention instead of $\textnormal{MultiHead}$.} (Eq.~\eqref{eqn:MAB_att}) treats input $\mathbf{x}\in\mathbb{R}^{n\times d}$ as key ($K$) and value ($V$), and uses a set of inducing points $\mathbf{I}\in\mathbb{R}^{m\times d}$ as query ($Q$).
First, it obtains the attention score matrix as follows:
\begin{equation}
    A = \frac{QK^\mathrm{T}}{\sqrt{d}} \in\mathbb{R}^{m\times n}.
\end{equation}
Each row index of $A$ denotes an inducing point, and each column index denotes an input element.
Then, the value set $V$ is aggregated using $A$.
With $\textnormal{Softmax}_{\textnormal{axis}=d}(\cdot)$ denoting softmax normalization along $d$-th axis, the plain attention normalizes each row of $A$, as follows:
\begin{align}
    \textnormal{Att}(Q, K, V) &= WV\in\mathbb{R}^{m\times d}, \\
    \textnormal{ where } W &= \textnormal{Softmax}_{\textnormal{axis}=2}(A)\in\mathbb{R}^{m\times n}.
\end{align}
As a result, an input element can get zero attention if every query suppresses it.
To prevent this, Slot Attention normalizes each column of $A$ and computes weighted mean:
\begin{align}
    \textnormal{SlotAtt}(Q, K, V) &= W'V, \\
    \textnormal{ where } W'_{ij} = \frac{A'_{ij}}{\sum_{l=1}^n{A'_{il}}}&
    \textnormal{ for } A' = \textnormal{Softmax}_{\textnormal{axis}=1}(A).
\end{align}
As attention coefficients across a row sum up to 1 after softmax, slot attention guarantees that an input element is not ignored by every inducing point.

With the adaptation of Slot Attention, we observe that inducing points often attend to distinct subsets of the input to produce $\mathbf{h}$, as illustrated in the Figure~\ref{fig:subset} and Figure~\ref{fig:composition} of the main text.
This is similar to the observation of \cite{locatello2020objectcentric} that the competition across queries encouraged segmented representations (slots) of objects from a multi-object image.
A difference is that unlike in \cite{locatello2020objectcentric} where the queries are noise vectors, we design the query set as a learnable parameter $\mathbf{I}$.
Also, we do not introduce any refinement steps to the projected set $\mathbf{h}$ to avoid the complication of the model.

\section{Experiment Details}
This section discusses the detailed descriptions and additional results of experiments in Section~\ref{sec:experiment} in the main paper.

\subsection{ShapeNet Evaluation Metrics}
\label{appendix:metrics}
We provide descriptions of evaluation metrics used in the ShapeNet experiment (Section~\ref{sec:experiment} in the main paper).
We measure standard metrics including coverage (COV), minimum matching distance (MMD), and 1-nearest neighbor accuracy (1-NNA) \cite{achlioptas2018learning, yang2019pointflow}.
Following recent literature \cite{kim2020softflow}, we omit Jensen-Shannon Divergence (JSD) \cite{achlioptas2018learning} because it does not assess the fidelity of each point cloud. To measure the similarity $D(\mathbf{x}, \mathbf{y})$ between point clouds $\mathbf{x}$ and $\mathbf{y}$, we use Chamfer Distance (CD) (Eq.~\eqref{eqn:reconstruction_cd}) and Earth Mover's Distance (EMD), where the EMD is defined as:
\begin{align}
    \textnormal{EMD}(\mathbf{x}, \mathbf{y}) = \min_{\pi} {\sum_i}{\| \mathbf{x}_i - \mathbf{y}_{\pi(i)}\|_2}.
\end{align}

Let $S_g$ be the set of generated point clouds and $S_r$ be the set of reference point clouds with $|S_r| = |S_g|$.

\emph{Coverage (COV)} measures the percentage of reference point clouds that is a nearest neighbor of at least one generated point cloud, computed as follows:
\begin{equation}
    \textnormal{COV}(S_g, S_r) = \frac{|\{\textnormal{argmin}_{\mathbf{y}\in S_r}D(\mathbf{x}, \mathbf{y})|\mathbf{x}\in S_g\}|}{|S_r|}.
\end{equation}

\emph{Minimum Matching Distance (MMD)} measures the average distance from each reference point cloud to its nearest neighbor in the generated point clouds:
\begin{equation}
    \textnormal{MMD}(S_g, S_r) = \frac{1}{|S_r|}\sum_{\mathbf{y}\in S_r}{\min_{\mathbf{x}\in S_g}{D(\mathbf{x}, \mathbf{y})}}.
\end{equation}

\emph{1-Nearest Neighbor Accuracy (1-NNA)} assesses whether two distributions are identical.
Let $S_{-\mathbf{x}}=S_r\cup S_g-\{\mathbf{x}\}$ and $N_{\mathbf{x}}$ be the nearest neighbor of $\mathbf{x}$ in $S_{-\mathbf{x}}$. With $\mathbf{1}(\cdot)$ an indicator function:
\begin{align}
    \textnormal{1-NNA}&(S_g, S_r) \nonumber\\
    &= \frac{\sum_{\mathbf{x}\in S_g}\mathbf{1}(N_{\mathbf{x}}\in S_g) + \sum_{\mathbf{y}\in S_r}\mathbf{1}(N_{\mathbf{y}}\in S_r)}{|S_g| + |S_r|}.
\end{align}

\subsection{Hierarchical Disentanglement}
This section describes an evaluation protocol used in Figure~\ref{fig:lda} in the main paper.
To investigate the latent representations learned at each level, we employed Linear Discriminant Analysis (LDA) as simple layer-wise classifiers.
The classifiers take the latent variable at each layer $\mathbf{z}^{l},~\forall l\in[1,L]$ as an input, and predict the identity and position of two digits (in $4\times4$ quantized grid) respectively in Set-MultiMNIST dataset.
To this end, we first train the SetVAE in the training set of Set-MultiMNIST.
Then we train the LDA classifiers using the validation dataset, where the input latent variables are sampled from the posterior distribution of SetVAE (Eq.~\eqref{eqn:posterior}). 
We report the training accuracy of the classifiers at each layer in Figure~\ref{fig:lda}.

\subsection{Ablation study}
In this section, we provide details of the ablation study presented in Table~\ref{table:ablation} of the main text.

\paragraph{Baseline}
As baselines, we use a SetVAE with unimodal Gaussian prior over the initial set elements, and a non-hierarchical, Vanilla SetVAE presented in Section~\ref{sec:method}.

To implement a SetVAE with unimodal prior, we only change the initial element distribution $p(\mathbf{z}_i^{(0)})$ from MoG (Eq.~\eqref{eqn:prior_mog}) to a multivariate Gaussian with a diagonal covariance matrix $\mathcal{N}(\mu^{(0)}, \sigma^{(0)})$ with learnable $\mu^{(0)}$ and $\sigma^{(0)}$.
This approach is adopted in several previous works in permutation-equivariant set generation \cite{yang2019pointflow, kosiorek2020conditional, locatello2020objectcentric}.

\begin{figure}[!t]
    \centering
    \begin{subfigure}[b]{0.395\linewidth}
        \centering
        \includegraphics[width=0.9\linewidth]{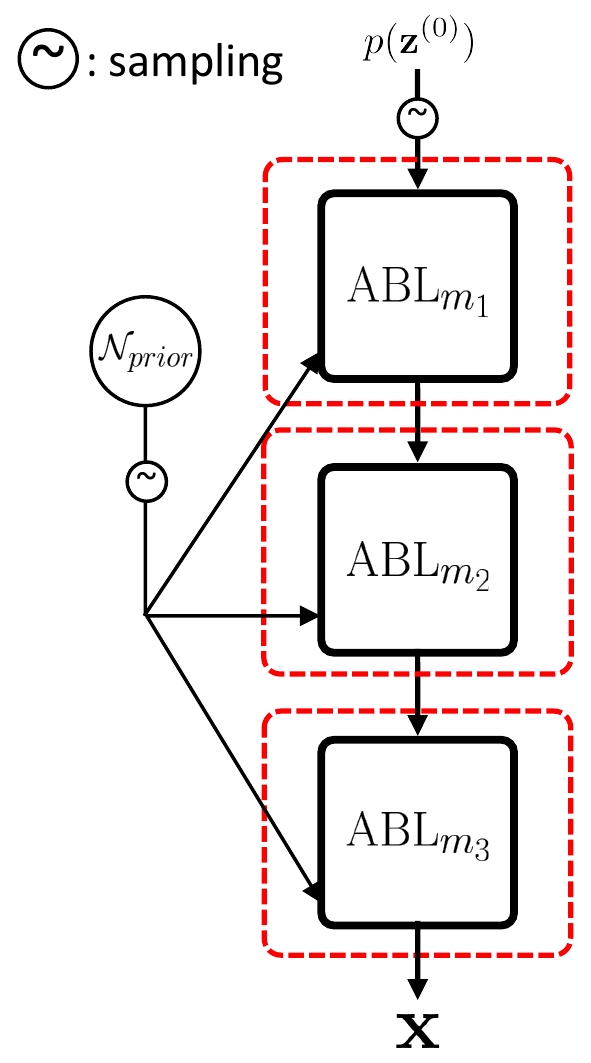}
        \caption{Generation}
        \label{fig:vanilla_prior}
    \end{subfigure}
    \begin{subfigure}[b]{0.59\linewidth}
        \centering
        \includegraphics[width=0.9\linewidth]{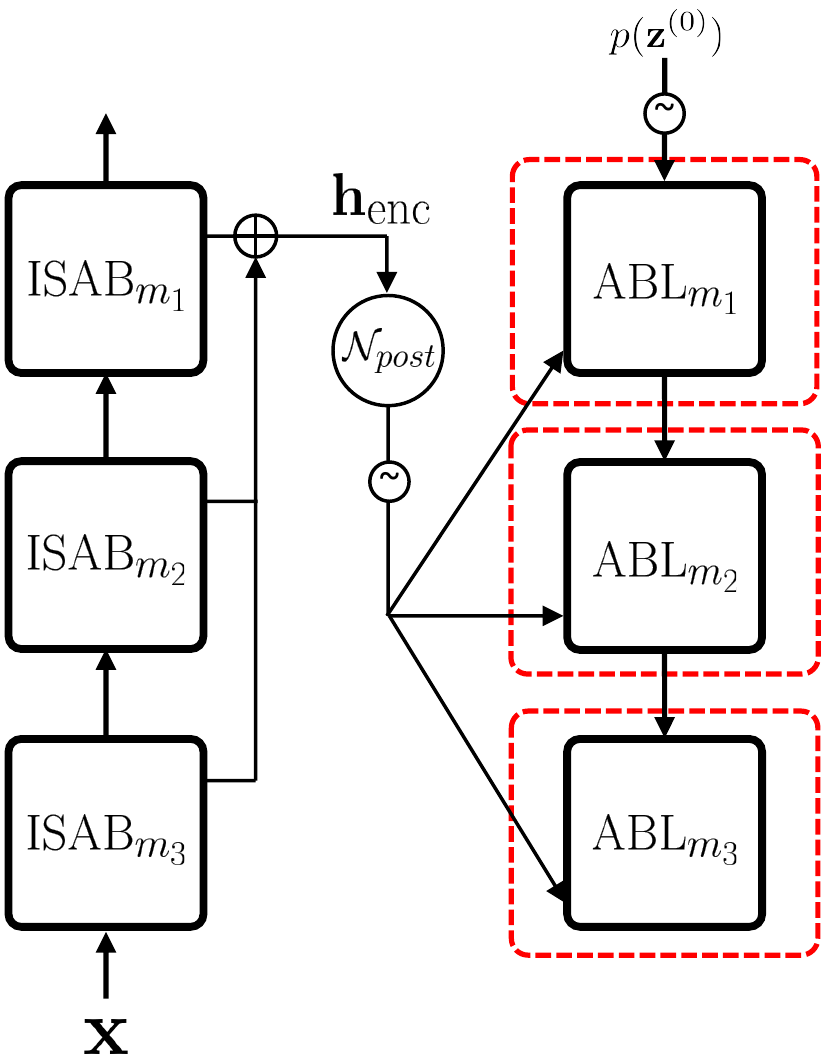}
        \caption{Inference}
        \label{fig:vanilla_posterior}
    \end{subfigure}
    \caption{Structure of Vanilla SetVAE without hierarchical priors and subset reasoning in generator.}
\label{fig:vanilla_setvae}
\end{figure}
To implement a Vanilla SetVAE, we employ a bottom-up encoder same to our full model and make the following changes to the top-down generator.
As illustrated in Figure~\ref{fig:vanilla_setvae}, we remove the subset relation in the generator by fixing the latent cardinality to 1 and employing a global prior $\mathcal{N}(\mu_1, \sigma_1)$ with $\mu_1, \sigma_1 \in \mathbb{R}^{1\times d}$ for all ABL.
To compute permutation-invariant $\mathbf{h}_{\textnormal{enc}} \in \mathbb{R}^{1\times d}$,
we aggregate every elements of $\mathbf{h}$ from all levels of encoder network by average pooling.
During inference, $\mathbf{h}_{\textnormal{enc}}$ is provided to every ABL in the top-down generator.

\paragraph{Evaluation metric}
For the ablation study of SetVAE on the Set-MultiMNIST dataset, we measure the generation quality in image space by rendering each set instance to $64\times 64$ binary image based on the occurrence of a point in a pixel bin.
To measure the generation performance, we compute Frechet Inception Distance (FID) score \cite{heusel2017gans} using the output of the penultimate layer of a VGG11 network trained from scratch for MultiMNIST image classification into 100 labels (00-99).
Given the channel-wise mean $\mu_g$, $\mu_r$ and covariance matrix $\Sigma_g$, $\Sigma_r$ of generated and reference set of images respectively, we compute FID as follows:
\begin{equation}
    d^2 = {\| \mu_g - \mu_r \|}^2 + \textnormal{Tr} (\Sigma_g + \Sigma_r - 2 \sqrt{\Sigma_g \Sigma_r}).
\end{equation}
To train the VGG11 network, we replace the first conv layer to take single-channel inputs, and use the same MultiMNIST train set as in SetVAE.
We use the SGD optimizer with Nesterov momentum, with learning rate 0.01, momentum 0.9, and L2 regularization weight 5e-3 to prevent overfitting.
We train the network for 10 epochs using batch size 128 so that the training top-1 accuracy exceeds 95\%.

\section{More Qualitative Results}

\begin{figure}[!t]
    \centering
    \includegraphics[width=0.8\linewidth]{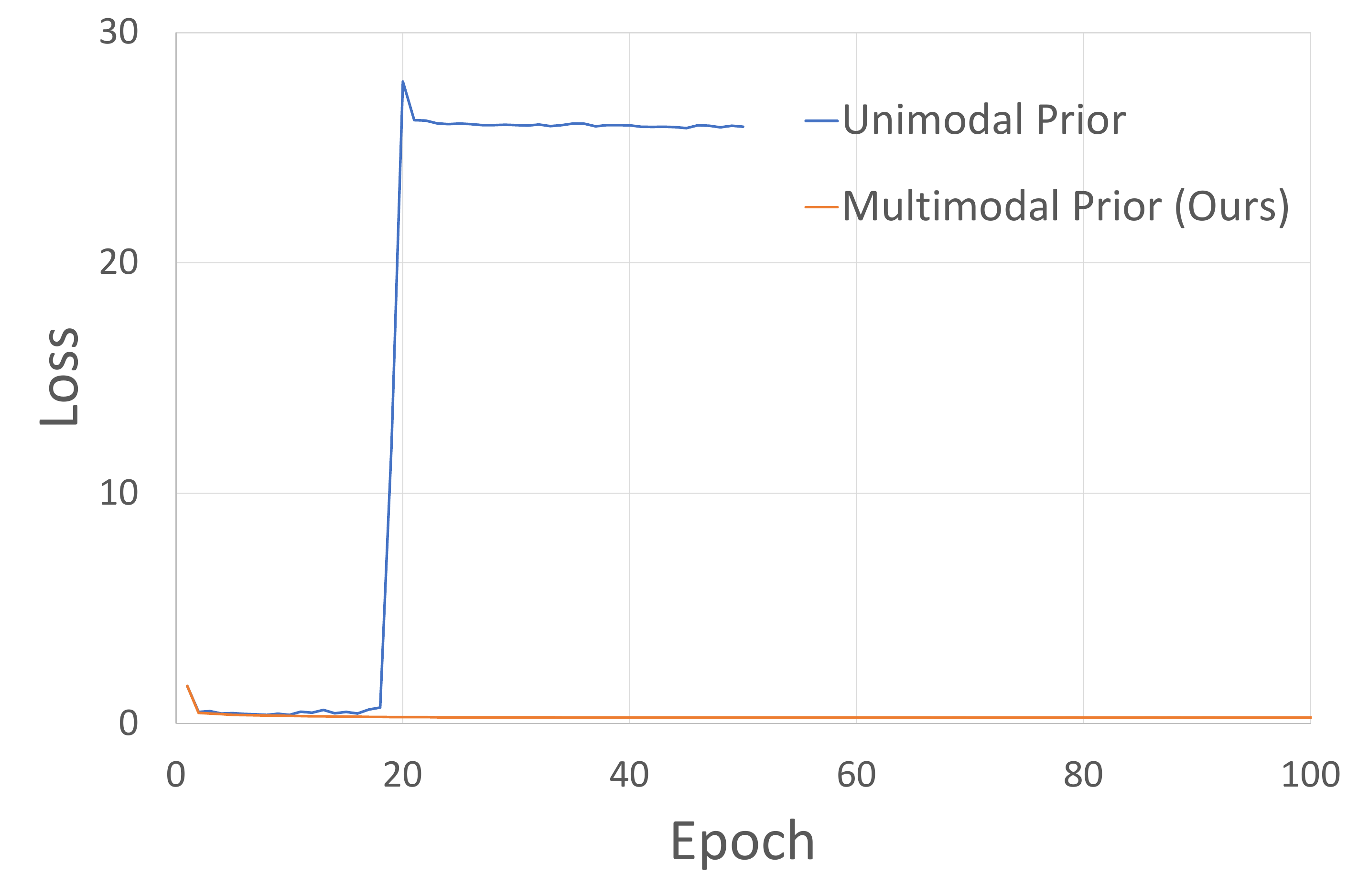}
    \caption{Training loss curves from SetVAE with multimodal and unimodal initial set trained on Set-MultiMNIST dataset.}
    \label{fig:ablation-curve}
    \centering
    \includegraphics[width=0.47\textwidth]{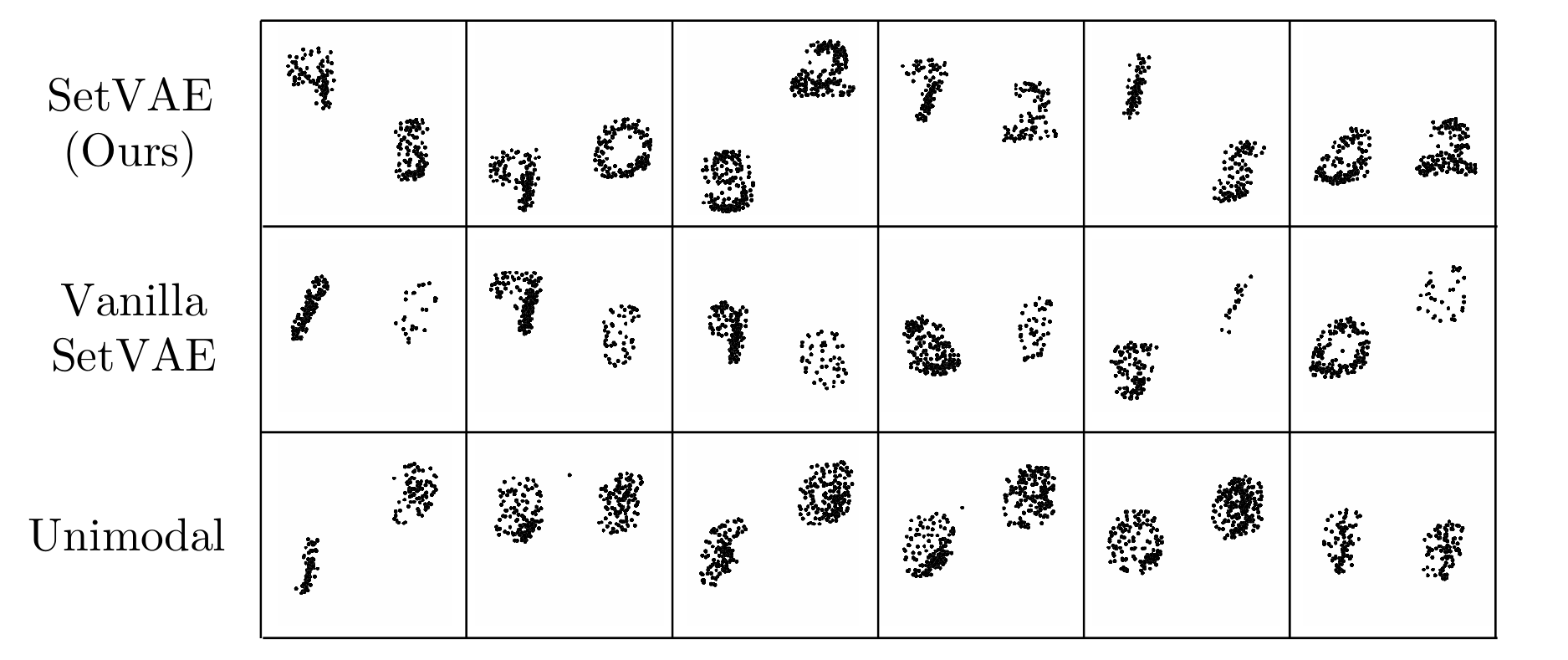}
    \caption{Samples from SetVAE and its ablated version trained on Set-MultiMNIST dataset.}
    \label{fig:ablation}
    \vspace{-0.2in}
\end{figure}

\paragraph{Ablation study}
This section provides additional results of the ablation study, which corresponds to the Table~\ref{table:ablation} of the main paper.
We compare the SetVAE with two baselines: SetVAE with a unimodal prior and the one using a single global latent variable (\emph{i.e.}, Vanilla SetVAE).

Figure~\ref{fig:ablation-curve} shows the training loss curves of SetVAE and the unimodal prior baseline on the Set-MultiMNIST dataset.
We observe that training of the unimodal baseline is unstable compared to SetVAE which uses a 4-component MoG.
We conjecture that a flexible initial set distribution provides a cue for the generator to learn stable subset representations.

In Figure~\ref{fig:ablation}, we visualize samples from SetVAE and the two baselines.
As the training of unimodal SetVAE was unstable, we provide the results from a checkpoint before the training loss diverges (third row of Figure~\ref{fig:ablation}).
The Vanilla SetVAE without hierarchy (second row of Figure~\ref{fig:ablation}) focuses on modeling the left digit only and fails to assign a balanced number of points.
This failure implies that multi-level subset reasoning in the generative process is essential in faithfully modeling complex set data such as Set-MultiMNIST.

\paragraph{Role of mixture initial set}
\begin{figure}[!t]
    \centering
    \includegraphics[width=0.95\linewidth]{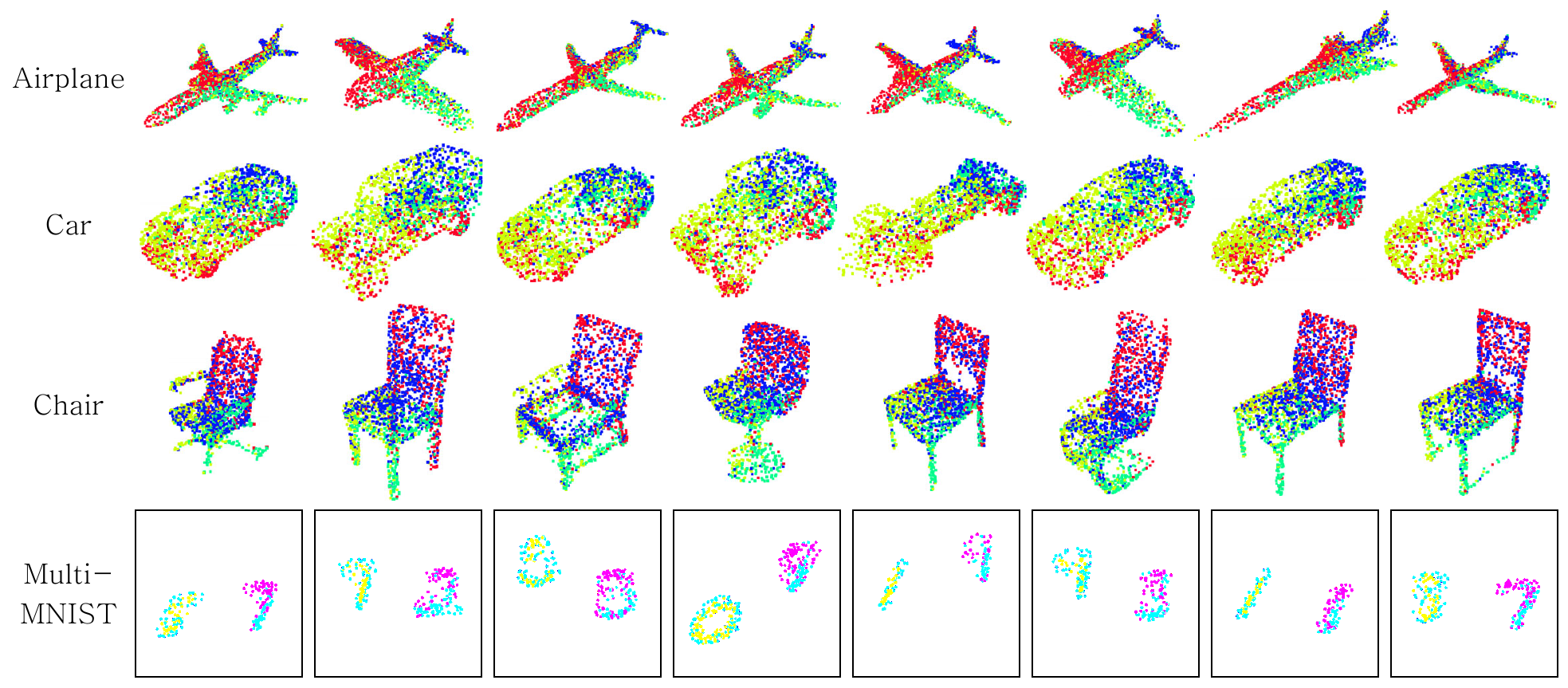}
    \caption{Color-coded mixture assignments on output sets.}
\label{fig:mog_attention}
\end{figure}
Although the multi-modal prior is not a typical choice, we emphasize that it marginally adds complexity to the model since it only introduces the additional learnable mixture parameters ($\pi_k^{(0)},\mu_k^{(0)},\sigma_k^{(0)}$).
Despite the simplicity, we observe that mixture prior is effective in stabilizing training, especially when there are clearly separating modes in data such as in the Set-MultiMNIST dataset (Fig.~\ref{fig:mog_attention}).
Still, the choice of the initial prior is flexible and orthogonal to our contribution.

\begin{figure*}[!ht]
    \centering
    \includegraphics[width=1\textwidth]{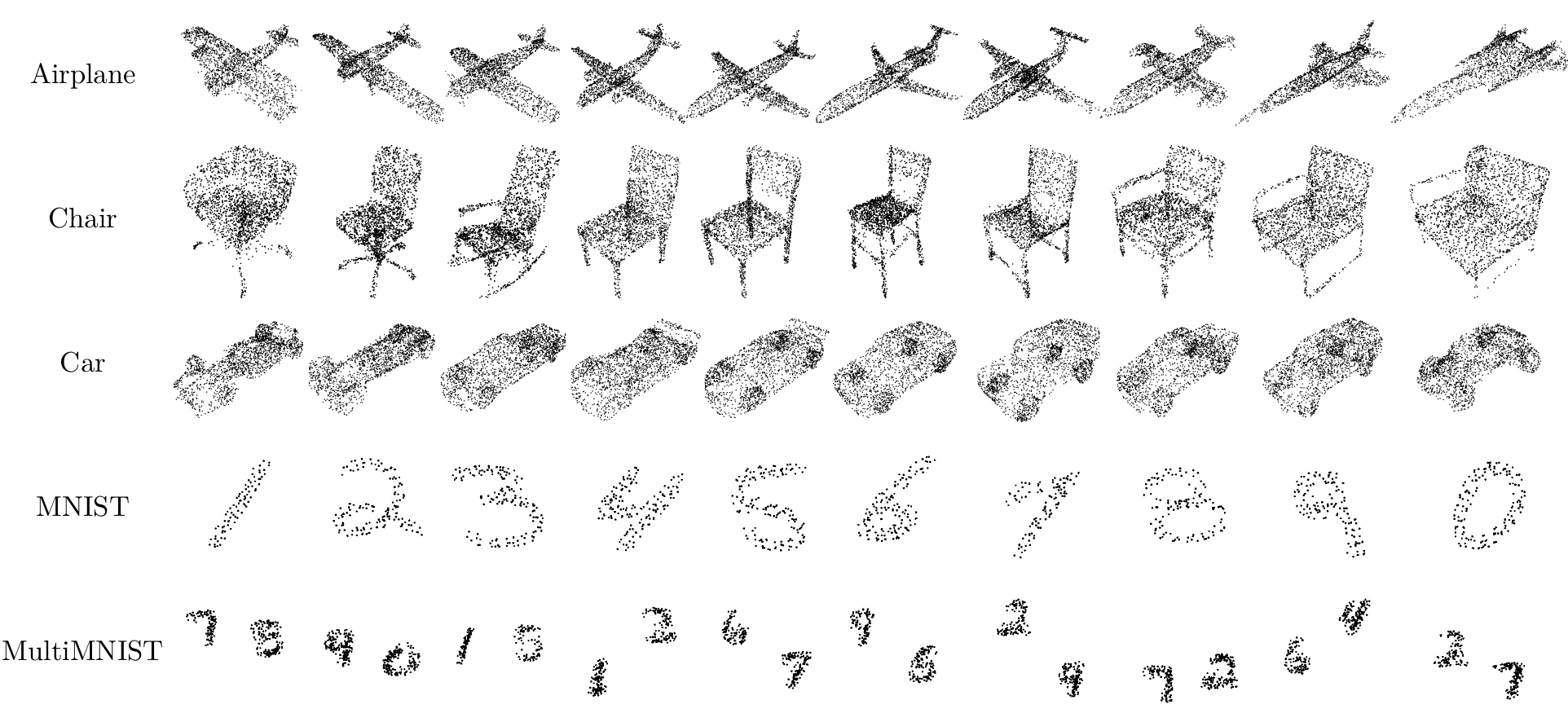}
    \caption{Additional examples of generated point clouds from SetVAE.}
    \label{fig:more_samples}
\end{figure*}

\begin{figure*}[!ht]
    \centering
    \includegraphics[width=1\textwidth]{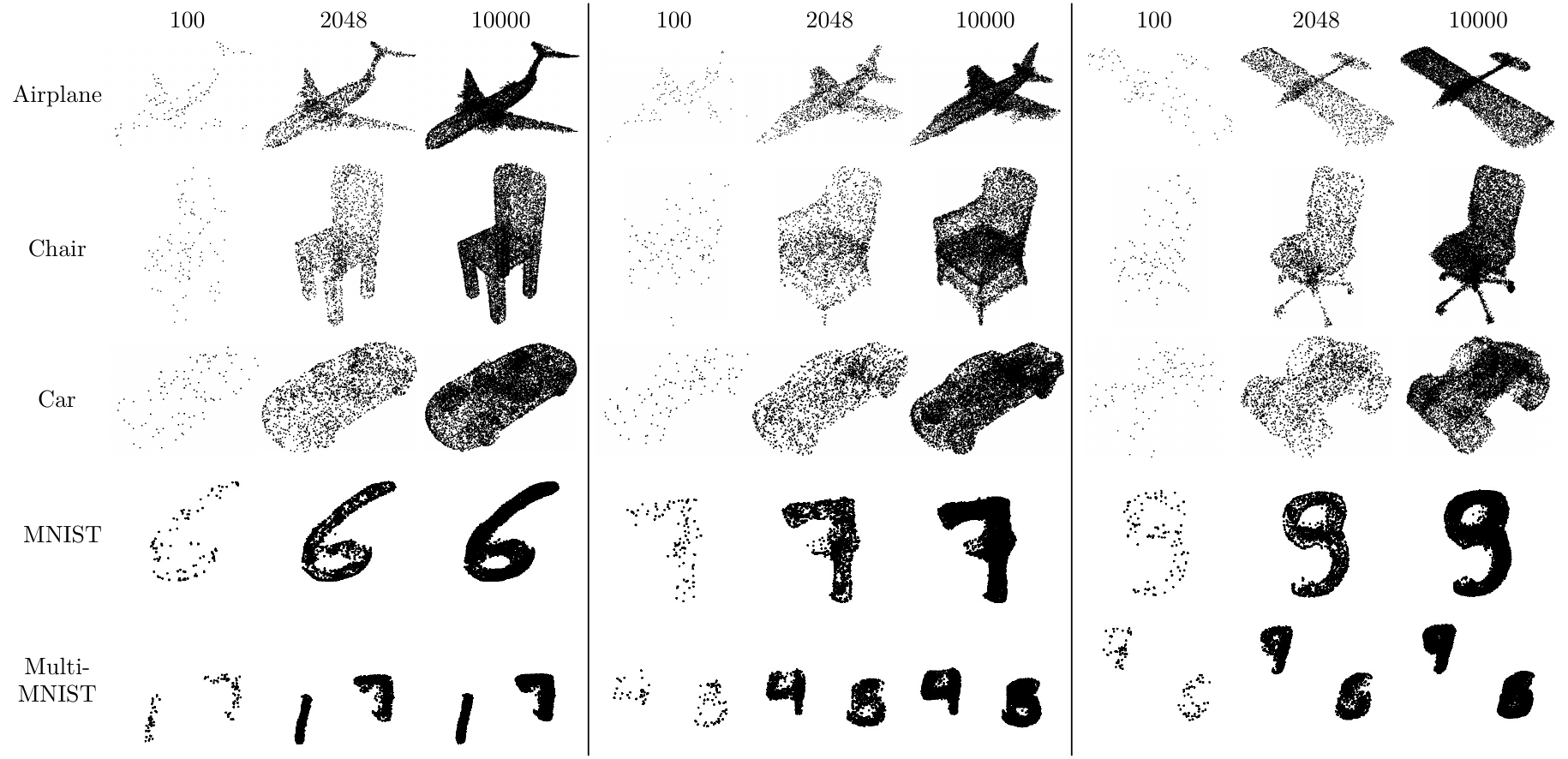}
    \caption{Additional examples demonstrating cardinality generalization of SetVAE.}
    \label{fig:more_disentanglement}
\end{figure*}

\paragraph{ShapeNet results}
Figure~\ref{fig:more_samples} presents the generated samples from SetVAE on ShapeNet, Set-MNIST and Set-MultiMNIST datasets, extending the results in Figure~\ref{fig:compare_pointflow} of the main text.
As illustrated in the figure, SetVAE generates point sets with high diversity while capturing thin and sharp details (\eg engines of an airplane and legs of a chair, \etc).
\paragraph{Cardinality disentanglement}
Figure~\ref{fig:more_disentanglement} presents the additional results of Figure~\ref{fig:cardinality} in the main paper, which illustrates samples generated by increasing the cardinality of the initial set $\mathbf{z}^{(0)}$ while fixing the hierarchical latent variables $\mathbf{z}^{(1:L)}$.
As illustrated in the figure, SetVAE is able to disentangle the cardinality of a set from the rest of its generative factors, and is able to generalize to unseen cardinality while preserving the disentanglement.

\begin{figure*}[!ht]
    \centering
    \includegraphics[width=1\textwidth]{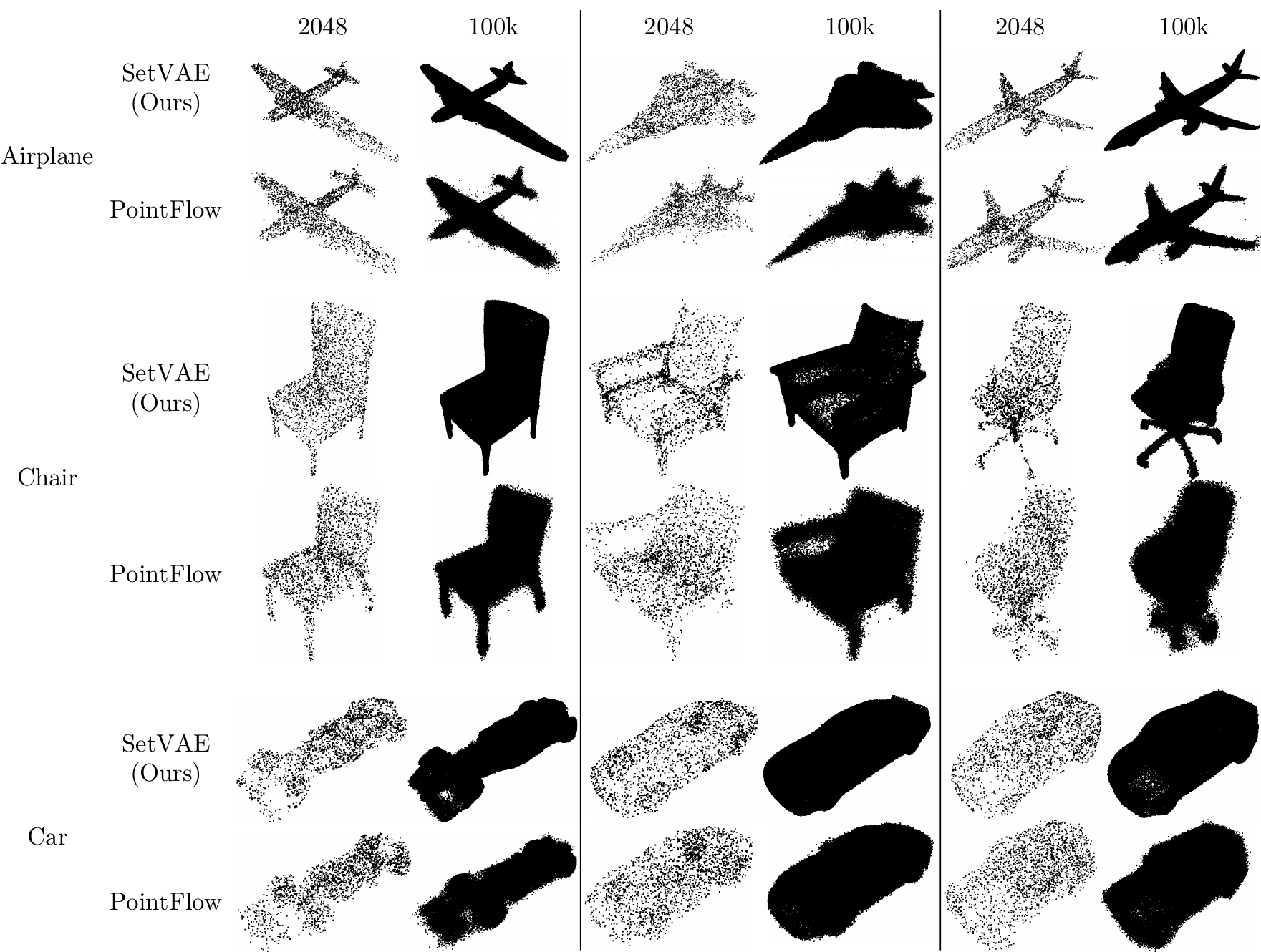}
    \caption{More examples in high-cardinality setting, compared with PointFlow.}
    \label{fig:more_mega_cardinality}
\end{figure*}

Notably, SetVAE can retain the disentanglement and generalize even to a high cardinality (100k) as well.
Figure~\ref{fig:more_mega_cardinality} presents the comparison to PointFlow with varying cardinality, which extends the results of the Figure~\ref{fig:mega-cardinality} in the main paper.
Unlike PointFlow that exhibits degradation and blurring of fine details, SetVAE retains the fine structure of the generated set even for extreme cardinality.

\begin{figure*}[!ht]
    \centering
    \includegraphics[width=1\textwidth]{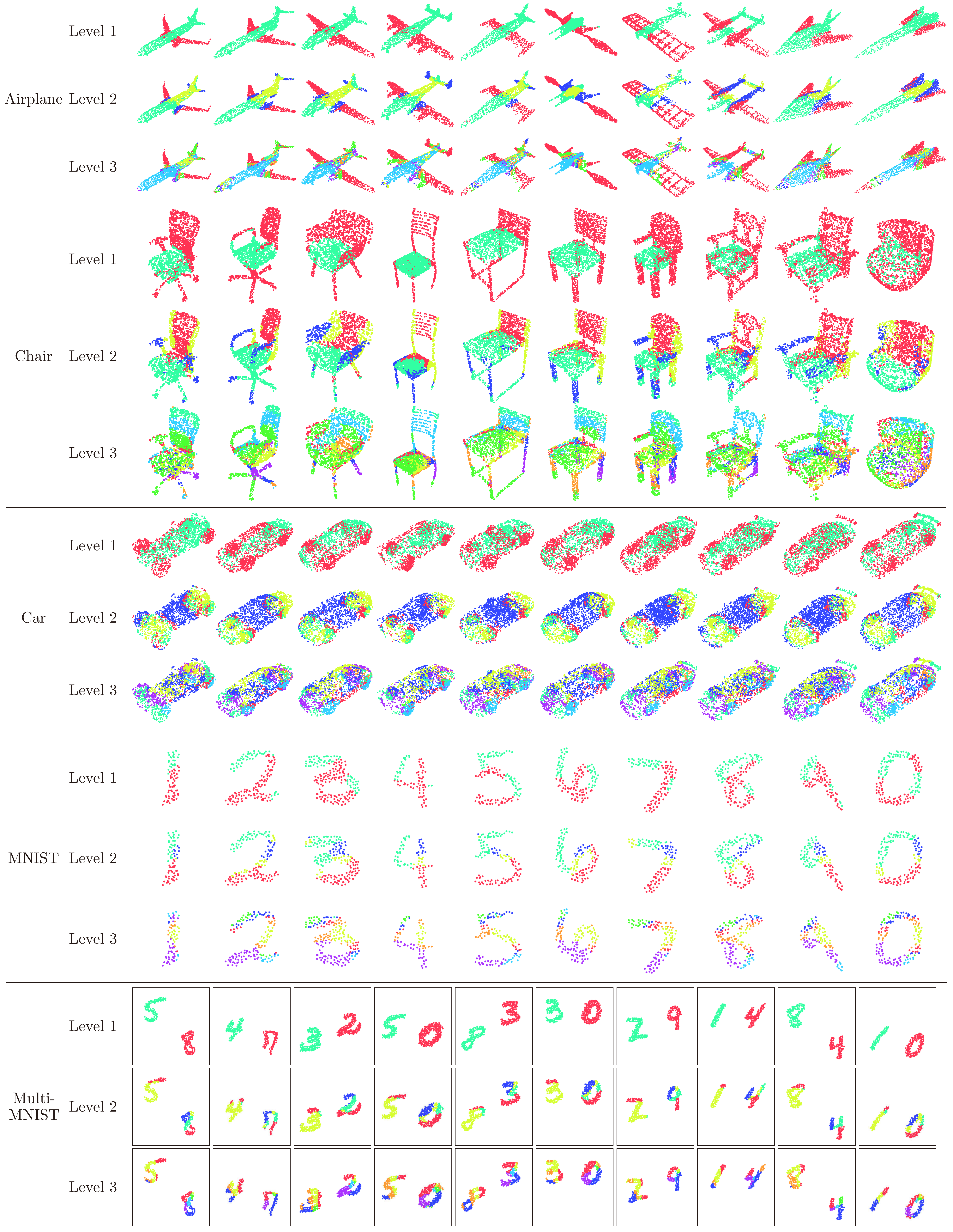}
    \caption{More examples of color-coded encoder attention.}
    \label{fig:more_encoder_attention}
\end{figure*}

\begin{figure*}[!ht]
    \centering
    \includegraphics[width=1\textwidth]{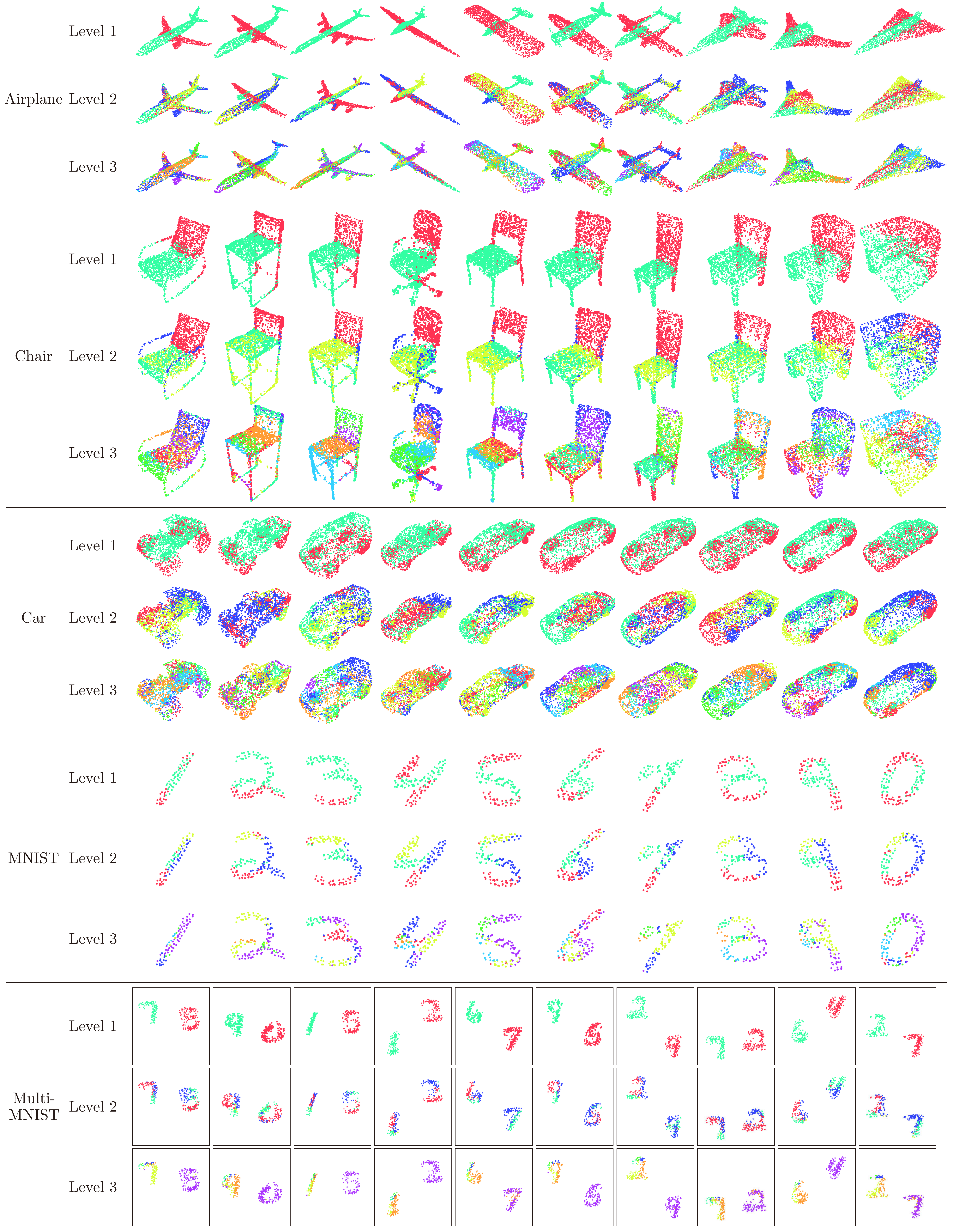}
    \caption{More examples of color-coded generator attention.}
    \label{fig:more_generator_attention}
\end{figure*}

\paragraph{Coarse-to-fine dependency}
In Figure~\ref{fig:more_encoder_attention} and Figure~\ref{fig:more_generator_attention}, we provide additional visualization of encoder and generator attention, extending the Figure~\ref{fig:subset} and Figure~\ref{fig:composition} in the main text.
We observe that SetVAE learns to attend to a subset of points consistently across examples.
Notably, these subsets often have a bilaterally symmetric structure or correspond to semantic parts.
For example, in the top level of the encoder (rows marked level 1 in Figure~\ref{fig:more_encoder_attention}), the subsets include wings of an airplane, legs \& back of a chair, or wheels \& rear wing of a car (colored red).

Furthermore, SetVAE extends the subset modeling to multiple levels with a top-down increase in latent cardinality.
This allows SetVAE to encode or generate the structure of a set in various granularities, ranging from global structure to fine details.
Each column in Figure~\ref{fig:more_encoder_attention} and Figure~\ref{fig:more_generator_attention} illustrates the relations.
For example, in level 3 of Figure~\ref{fig:more_encoder_attention}, the bottom-up encoder partitions an airplane into fine-grained parts such as an engine, a tip of the tail wing, \etc.
Then, going bottom-up to level 1, the encoder composes them to fuselage and symmetric pair of wings.
As for the top-down generator in Figure~\ref{fig:more_generator_attention}, it starts in level 1 by composing an airplane via the coarsely defined body and wings.
Going top-down to level 3, the generator descends into fine-grained subsets like an engine and tail wing.

\section{Architecture and Hyperparameters}
\label{appendix:training}
Table~\ref{table:architecture} provides the network architecture and hyperparameters of SetVAE.
In the table, $\textnormal{FC}(d, f)$ denotes a fully-connected layer with output dimension $d$ and nonlinearity $f$.
$\textnormal{ISAB}_m(d, h)$ denotes an $\textnormal{ISAB}_m$ with $m$ inducing points, hidden dimension $d$, and $h$ heads (in Section~\ref{sec:settransformer}).
$\textnormal{MoG}_K(d)$ denotes a mixture of Gaussian (in Eq.~\eqref{eqn:prior_mog}) with $K$ components and dimension $d$.
$\textnormal{ABL}_m(d, d_z, h)$ denotes an $\textnormal{ABL}_m$ with $m$ inducing points, hidden dimension $d$, latent dimension $d_z$, and $h$ heads (in Section~\ref{sec:architecture}).
All $\textnormal{MAB}$s used in $\textnormal{ISAB}$ and $\textnormal{ABL}$ uses fully-connected layer with bias as $\textnormal{FF}$ layer.

In Table~\ref{table:hyperparameters}, we provide detailed training hyperparameters.
For all experiments, we used Adam optimizer with first and second momentum parameters $0.9$ and $0.999$, respectively and decayed the learning rate linearly to zero after $50\%$ of the training schedule.
Following \cite{vahdat2020nvae}, we linearly annealed $\beta$ from 0 to 1 during the first 2000 epochs for ShapeNet datasets, 40 epochs for Set-MNIST, and 50 epochs for Set-MultiMNIST dataset.

\clearpage


\begin{figure*}[!t]
\begin{minipage}{\linewidth}
\captionof{table}{Detailed network architectures used in our experiments.}
\vspace{-0.2cm}
\centering
    \begin{adjustbox}{width=0.95\textwidth}
        \label{table:architecture}
        \begin{tabular}{cc|cc|cc}
        \Xhline{2\arrayrulewidth}
        \\[-1em] \multicolumn{2}{c}{\textbf{ShapeNet}} & \multicolumn{2}{c}{\textbf{Set-MNIST}} & \multicolumn{2}{c}{\textbf{Set-MultiMNIST}}\\
        \\[-1em]\Xhline{2\arrayrulewidth}
        \\[-1em] \textbf{Encoder} & \textbf{Generator} & \textbf{Encoder} & \textbf{Generator} & \textbf{Encoder} & \textbf{Generator} \\
        \\[-1em]\Xhline{2\arrayrulewidth}
        \\[-1em]
        Input: $\textnormal{FC}(64, -)$ & Initial set: $\textnormal{MoG}_4(32)$ & Input: $\textnormal{FC}(64, -)$ & Initial set: $\textnormal{MoG}_4(32)$ & Input: $\textnormal{FC}(64, -)$ & Initial set: $\textnormal{MoG}_{16}(64)$ \\
        $\textnormal{ISAB}_{32}(64, 4)$ & $\textnormal{ABL}_{1}(64, 16, 4)$ & $\textnormal{ISAB}_{32}(64, 4)$ & $\textnormal{ABL}_{2}(64, 16, 4)$ & $\textnormal{ISAB}_{32}(64, 4)$ & $\textnormal{ABL}_{2}(64, 16, 4)$ \\
        $\textnormal{ISAB}_{16}(64, 4)$ & $\textnormal{ABL}_{1}(64, 16, 4)$ & $\textnormal{ISAB}_{16}(64, 4)$ & $\textnormal{ABL}_{4}(64, 16, 4)$ & $\textnormal{ISAB}_{16}(64, 4)$ & $\textnormal{ABL}_{4}(64, 16, 4)$ \\
        $\textnormal{ISAB}_{8}(64, 4)$ & $\textnormal{ABL}_{2}(64, 16, 4)$ & $\textnormal{ISAB}_{8}(64, 4)$ & $\textnormal{ABL}_{8}(64, 16, 4)$ & $\textnormal{ISAB}_{8}(64, 4)$ & $\textnormal{ABL}_{8}(64, 16, 4)$ \\
        $\textnormal{ISAB}_{4}(64, 4)$ & $\textnormal{ABL}_{4}(64, 16, 4)$ & $\textnormal{ISAB}_{4}(64, 4)$ & $\textnormal{ABL}_{16}(64, 16, 4)$ & $\textnormal{ISAB}_{4}(64, 4)$ & $\textnormal{ABL}_{16}(64, 16, 4)$ \\
        $\textnormal{ISAB}_{2}(64, 4)$ & $\textnormal{ABL}_{8}(64, 16, 4)$ & $\textnormal{ISAB}_{2}(64, 4)$ & $\textnormal{ABL}_{32}(64, 16, 4)$ & $\textnormal{ISAB}_{2}(64, 4)$ & $\textnormal{ABL}_{32}(64, 16, 4)$ \\
        $\textnormal{ISAB}_{1}(64, 4)$ & $\textnormal{ABL}_{16}(64, 16, 4)$ & & Output: $\textnormal{FC}(2, \textnormal{tanh})$ & & Output: $\textnormal{FC}(2, \textnormal{tanh})$ \\
        $\textnormal{ISAB}_{1}(64, 4)$ & $\textnormal{ABL}_{32}(64, 16, 4)$ & & $(\textnormal{Output}+1)/2$ & & $(\textnormal{Output}+1)/2$ \\
        & Output: $\textnormal{FC}(3, -)$ & & & & \\
        & & & & & \\
        \\[-1em]\Xhline{2\arrayrulewidth}
        \end{tabular}
    \end{adjustbox}
    \vspace{1cm}
    \captionof{table}{Detailed training hyperparameters used in our experiments.}
    \vspace{0.1cm}
    \begin{adjustbox}{width=0.95\textwidth}
        \label{table:hyperparameters}
        \footnotesize
        \begin{tabular}{cccc}
        \Xhline{2\arrayrulewidth}
        \\[-1em] & \textbf{ShapeNet} & \textbf{Set-MNIST} & \textbf{Set-MultiMNIST}\\
        \\[-1em] \Xhline{2\arrayrulewidth}
        \\[-1em]
        Minibatch size & 128 & 64 & 64 \\
        Training epochs & 8000 & 200 & 200 \\
        Learning rate & \multicolumn{3}{c}{1e-3, linear decay to zero after half of training} \\
        $\beta$ (Eq.~\eqref{eqn:beta_vae_loss}) & 1.0, annealed (-2000epoch) & 0.01, annealed (-50epoch) & 0.01, annealed (-40epoch) \\
        \\[-1em]\Xhline{2\arrayrulewidth}
        \end{tabular}
    \end{adjustbox}
\end{minipage}
\end{figure*}

%% file: arxiv.bbl
\begin{thebibliography}{10}\itemsep=-1pt

\bibitem{achlioptas2018learning}
Panos Achlioptas, Olga Diamanti, Ioannis Mitliagkas, and Leonidas~J. Guibas.
\newblock Learning representations and generative models for 3d point clouds.
\newblock In {\em ICLR}, 2018.

\bibitem{carion2020endtoend}
Nicolas Carion, Francisco Massa, Gabriel Synnaeve, Nicolas Usunier, Alexander
  Kirillov, and Sergey Zagoruyko.
\newblock End-to-end object detection with transformers.
\newblock In {\em ECCV}, 2020.

\bibitem{chang2015shapenet}
Angel~X. Chang, Thomas Funkhouser, Leonidas Guibas, Pat Hanrahan, Qixing Huang,
  Zimo Li, Silvio Savarese, Manolis Savva, Shuran Song, Hao Su, Jianxiong Xiao,
  Li Yi, and Fisher Yu.
\newblock {ShapeNet: An Information-Rich 3D Model Repository}.
\newblock Technical report, Stanford University --- Princeton University ---
  Toyota Technological Institute at Chicago, 2015.

\bibitem{edwards2017neural}
Harrison Edwards and Amos~J. Storkey.
\newblock Towards a neural statistician.
\newblock In {\em ICLR}, 2017.

\bibitem{eslami2016attend}
S.~M.~Ali Eslami, Nicolas Heess, Theophane Weber, Yuval Tassa, David
  Szepesvari, Koray Kavukcuoglu, and Geoffrey~E. Hinton.
\newblock Attend, infer, repeat: Fast scene understanding with generative
  models.
\newblock In {\em NeurIPS}, 2016.

\bibitem{finn2017modelagnostic}
Chelsea Finn, Pieter Abbeel, and Sergey Levine.
\newblock Model-agnostic meta-learning for fast adaptation of deep networks.
\newblock In {\em ICML}, 2017.

\bibitem{goodfellow2014generative}
Ian Goodfellow, Jean Pouget-Abadie, Mehdi Mirza, Bing Xu, David Warde-Farley,
  Sherjil Ozair, Aaron Courville, and Yoshua Bengio.
\newblock Generative adversarial nets.
\newblock In {\em NeurIPS}, 2014.

\bibitem{greff2019multi}
Klaus Greff, Rapha{\"{e}}l~Lopez Kaufman, Rishabh Kabra, Nick Watters,
  Christopher Burgess, Daniel Zoran, Loic Matthey, Matthew Botvinick, and
  Alexander Lerchner.
\newblock Multi-object representation learning with iterative variational
  inference.
\newblock In {\em ICML}, 2019.

\bibitem{heusel2017gans}
Martin Heusel, Hubert Ramsauer, Thomas Unterthiner, Bernhard Nessler, and Sepp
  Hochreiter.
\newblock Gans trained by a two time-scale update rule converge to a local nash
  equilibrium.
\newblock In {\em NeurIPS}, 2017.

\bibitem{hong2018inferring}
Seunghoon Hong, Dingdong Yang, Jongwook Choi, and Honglak Lee.
\newblock Inferring semantic layout for hierarchical text-to-image synthesis.
\newblock In {\em CVPR}, 2018.

\bibitem{johnson2018structured}
Matthew~J. Johnson, David Duvenaud, Alexander~B. Wiltschko, Sandeep~R. Datta,
  and Ryan~P. Adams.
\newblock Structured vaes: Composing probabilistic graphical models and
  variational autoencoders.
\newblock {\em CoRR}, 2018.

\bibitem{karras2019stylebased}
Tero Karras, Samuli Laine, and Timo Aila.
\newblock A style-based generator architecture for generative adversarial
  networks.
\newblock In {\em CVPR}, 2019.

\bibitem{kim2020softflow}
Hyeongju Kim, Hyeonseung Lee, Woo~Hyun Kang, Joun~Yeop Lee, and Nam~Soo Kim.
\newblock Softflow: Probabilistic framework for normalizing flow on manifolds.
\newblock {\em CoRR}, 2020.

\bibitem{kingma2017improving}
Diederik~P. Kingma, Tim Salimans, and Max Welling.
\newblock Improving variational inference with inverse autoregressive flow.
\newblock In {\em NeurIPS}, 2016.

\bibitem{kingma2014autoencoding}
Diederik~P. Kingma and Max Welling.
\newblock Auto-encoding variational bayes.
\newblock In {\em ICLR}, 2014.

\bibitem{kosiorek2020conditional}
Adam~R. Kosiorek, Hyunjik Kim, and Danilo~J. Rezende.
\newblock Conditional set generation with transformers.
\newblock {\em CoRR}, 2020.

\bibitem{lee2019set}
Juho Lee, Yoonho Lee, Jungtaek Kim, Adam~R. Kosiorek, Seungjin Choi, and
  Yee~Whye Teh.
\newblock Set transformer: {A} framework for attention-based
  permutation-invariant neural networks.
\newblock In {\em ICML}, 2019.

\bibitem{li2018point}
Chun{-}Liang Li, Manzil Zaheer, Yang Zhang, Barnab{\'{a}}s P{\'{o}}czos, and
  Ruslan Salakhutdinov.
\newblock Point cloud {GAN}.
\newblock In {\em ICLR}, 2019.

\bibitem{li2019grains}
Manyi Li, Akshay~Gadi Patil, Kai Xu, Siddhartha Chaudhuri, Owais Khan, Ariel
  Shamir, Changhe Tu, Baoquan Chen, Daniel Cohen{-}Or, and Hao~(Richard) Zhang.
\newblock {GRAINS:} generative recursive autoencoders for indoor scenes.
\newblock In {\em {ACM} Trans. Graph.}, 2019.

\bibitem{li2020exchangeable}
Yang Li, Haidong Yi, Christopher~M. Bender, Siyuan Shan, and Junier~B. Oliva.
\newblock Exchangeable neural {ODE} for set modeling.
\newblock {\em NeurIPS}, 2020.

\bibitem{locatello2020objectcentric}
Francesco Locatello, Dirk Weissenborn, Thomas Unterthiner, Aravindh Mahendran,
  Georg Heigold, Jakob Uszkoreit, Alexey Dosovitskiy, and Thomas Kipf.
\newblock Object-centric learning with slot attention.
\newblock In {\em NeurIPS}, 2020.

\bibitem{ritchie2019fast}
Daniel Ritchie, Kai Wang, and Yu{-}An Lin.
\newblock Fast and flexible indoor scene synthesis via deep convolutional
  generative models.
\newblock In {\em CVPR}, 2019.

\bibitem{sonderby2016ladder}
Casper~Kaae S{\o}nderby, Tapani Raiko, Lars Maal{\o}e, S{\o}ren~Kaae
  S{\o}nderby, and Ole Winther.
\newblock Ladder variational autoencoders.
\newblock In {\em NeurIPS}, 2016.

\bibitem{song2015sun}
Shuran Song, Samuel~P. Lichtenberg, and Jianxiong Xiao.
\newblock {SUN} {RGB-D:} {A} {RGB-D} scene understanding benchmark suite.
\newblock In {\em {IEEE} Conference on Computer Vision and Pattern Recognition,
  {CVPR} 2015, Boston, MA, USA, June 7-12, 2015}, 2015.

\bibitem{stelzner2020generative}
Karl Stelzner, Kristian Kersting, and Adam R.~Kosiorek 2.
\newblock Generative adversarial set transformers, 2020.

\bibitem{vahdat2020nvae}
Arash Vahdat and Jan Kautz.
\newblock {NVAE:} {A} deep hierarchical variational autoencoder.
\newblock In {\em NeurIPS}, 2020.

\bibitem{oord2016conditional}
A{\"{a}}ron van~den Oord, Nal Kalchbrenner, Lasse Espeholt, Koray Kavukcuoglu,
  Oriol Vinyals, and Alex Graves.
\newblock Conditional image generation with pixelcnn decoders.
\newblock In {\em NeurIPS}, 2016.

\bibitem{oord2016pixel}
A{\"{a}}ron van~den Oord, Nal Kalchbrenner, and Koray Kavukcuoglu.
\newblock Pixel recurrent neural networks.
\newblock In {\em ICML}, 2016.

\bibitem{vaswani2017attention}
Ashish Vaswani, Noam Shazeer, Niki Parmar, Jakob Uszkoreit, Llion Jones,
  Aidan~N. Gomez, Lukasz Kaiser, and Illia Polosukhin.
\newblock Attention is all you need.
\newblock In {\em NeurIPS}, 2017.

\bibitem{yang2019pointflow}
Guandao Yang, Xun Huang, Zekun Hao, Ming{-}Yu Liu, Serge~J. Belongie, and
  Bharath Hariharan.
\newblock Pointflow: 3d point cloud generation with continuous normalizing
  flows.
\newblock In {\em ICCV}, 2019.

\bibitem{yang2020energybased}
Mengjiao Yang, Bo Dai, Hanjun Dai, and Dale Schuurmans.
\newblock Energy-based processes for exchangeable data.
\newblock In {\em ICML}, 2020.

\bibitem{zaheer2017deep}
Manzil Zaheer, Satwik Kottur, Siamak Ravanbakhsh, Barnab{\'{a}}s P{\'{o}}czos,
  Ruslan Salakhutdinov, and Alexander~J. Smola.
\newblock Deep sets.
\newblock In {\em NeurIPS}, 2017.

\bibitem{zhang2020deep}
Yan Zhang, Jonathon~S. Hare, and Adam Pr{\"{u}}gel{-}Bennett.
\newblock Deep set prediction networks.
\newblock In {\em NeurIPS}, 2019.

\bibitem{zhang2020fspool}
Yan Zhang, Jonathon~S. Hare, and Adam Pr{\"{u}}gel{-}Bennett.
\newblock Fspool: Learning set representations with featurewise sort pooling.
\newblock In {\em ICLR}, 2020.

\end{thebibliography}
